\documentclass{article}
\usepackage{abstract}
\usepackage{authblk}
\usepackage{arxiv}
\usepackage{microtype}
\usepackage{graphicx}
\usepackage{booktabs} % for professional tables
\usepackage{comment}
\usepackage{hyperref}
\usepackage{natbib}
\RequirePackage{algorithm}
\RequirePackage{algorithmic}

\usepackage{amsmath}
\usepackage{amssymb}
\usepackage{mathtools}
\usepackage{amsthm}

\usepackage[capitalize,noabbrev]{cleveref}

\theoremstyle{plain}
\newtheorem{theorem}{Theorem}[section]

\theoremstyle{definition}

\theoremstyle{remark}

\usepackage[textsize=tiny]{todonotes}

\usepackage{subcaption}
\usepackage{xcolor}
\usepackage{tabularx} 
\newcolumntype{Y}{>{\centering\arraybackslash}X}
\usepackage{multirow}
\usepackage[export]{adjustbox}
\newcommand{\argmin}{\operatornamewithlimits{argmin}}

\newcommand{\fromto}{\longrightarrow}
\renewcommand{\vec}[1]{\boldsymbol{#1}}
\definecolor{credal_color}{RGB}{57, 172, 115}
\definecolor{gt_color}{RGB}{255, 153, 0}
\title{ Conformalized Credal Set Predictors}
\author[1,2]{Alireza Javanmardi}
\author[3]{David Stutz }
\author[1,2]{Eyke H\"ullermeier }
\affil[1]{Institute of Informatics, LMU Munich, Germany}
\affil[2]{Munich Center for Machine Learning (MCML), Germany}
\affil[3]{Max Planck Institute for Informatics, Saarland Informatics Campus, Germany}
\affil[ ]{\href{mailto:alireza.javanmardi@ifi.lmu.de}{\texttt{alireza.javanmardi@ifi.lmu.de}},~~ \href{mailto:david.stutz@mpi-inf.mpg.de}{\texttt{david.stutz@mpi-inf.mpg.de}},~~ \href{mailto:eyke@ifi.lmu.de}{\texttt{eyke@ifi.lmu.de}}}
\date{}
\renewcommand{\shorttitle}{Conformalized Credal Set Predictors}

\begin{document}
\maketitle

\begin{abstract}
Credal sets are sets of probability distributions that are considered as candidates for an imprecisely known ground-truth distribution. In machine learning, they have recently attracted attention as an appealing formalism for uncertainty representation, in particular due to their ability to represent both the aleatoric and epistemic uncertainty in a prediction. However, the design of methods for learning credal set predictors remains a challenging problem. In this paper, we make use of conformal prediction for this purpose. More specifically, we propose a method for predicting credal sets in the classification task, given training data labeled by probability distributions. Since our method inherits the coverage guarantees of conformal prediction, our conformal credal sets are guaranteed to be valid with high probability (without any assumptions on model or distribution). We demonstrate the applicability of our method to natural language inference, a highly ambiguous natural language task where it is common to obtain multiple annotations per example. 

%Conformal Prediction (CP) is one of the most promising ways of providing predictive uncertainty in Machine learning. Despite its widespread applicability and solid theoretical foundation, CP falls short in distinguishing between two critical forms of uncertainty: aleatoric and epistemic. Within the context of the classification problem, we aim to address this gap by extending CP to generate a credal set—essentially, a collection of probability distributions—for a given instance. Under the assumption that each data instance is associated with a ground truth probability distribution across classes, and given the availability of such data, the proposed conformal credal set adheres to the coverage property, ensuring that the ground truth probability distribution is encompassed with a specified probability. Furthermore, using these credal sets facilitates a clearer distinction between irreducible and reducible components of uncertainty.  

\end{abstract}
\keywords{Conformal Prediction \and Uncertainty Representation \and Credal Sets}

\section{Introduction}
Representing and quantifying uncertainty is becoming increasingly important in machine learning (ML), particularly as ML models are employed in safety-critical application domains such as medicine or autonomous driving. In such domains, a distinction between so-called \textit{aleatoric uncertainty} and \textit{epistemic uncertainty} is often useful \citep{hora1996aleatory}. Broadly speaking, aleatoric uncertainty is due to the inherent randomness of the data-generating process, whereas epistemic uncertainty stems from the learner's lack of knowledge about the best predictive model. Thus, while the former is irreducible, the latter can in principle be reduced through additional information, e.g., by gathering additional data to learn from. 
%Addressing this distinction remains an interesting challenge in the machine learning community.

%hullermeier2021aleatoric
%Generally speaking, a learner encounters two types of uncertainty when predicting a label for a given query instance: 

\begin{figure}[t]
\begin{subfigure}{0.25\columnwidth}
    \includegraphics[width=\textwidth]{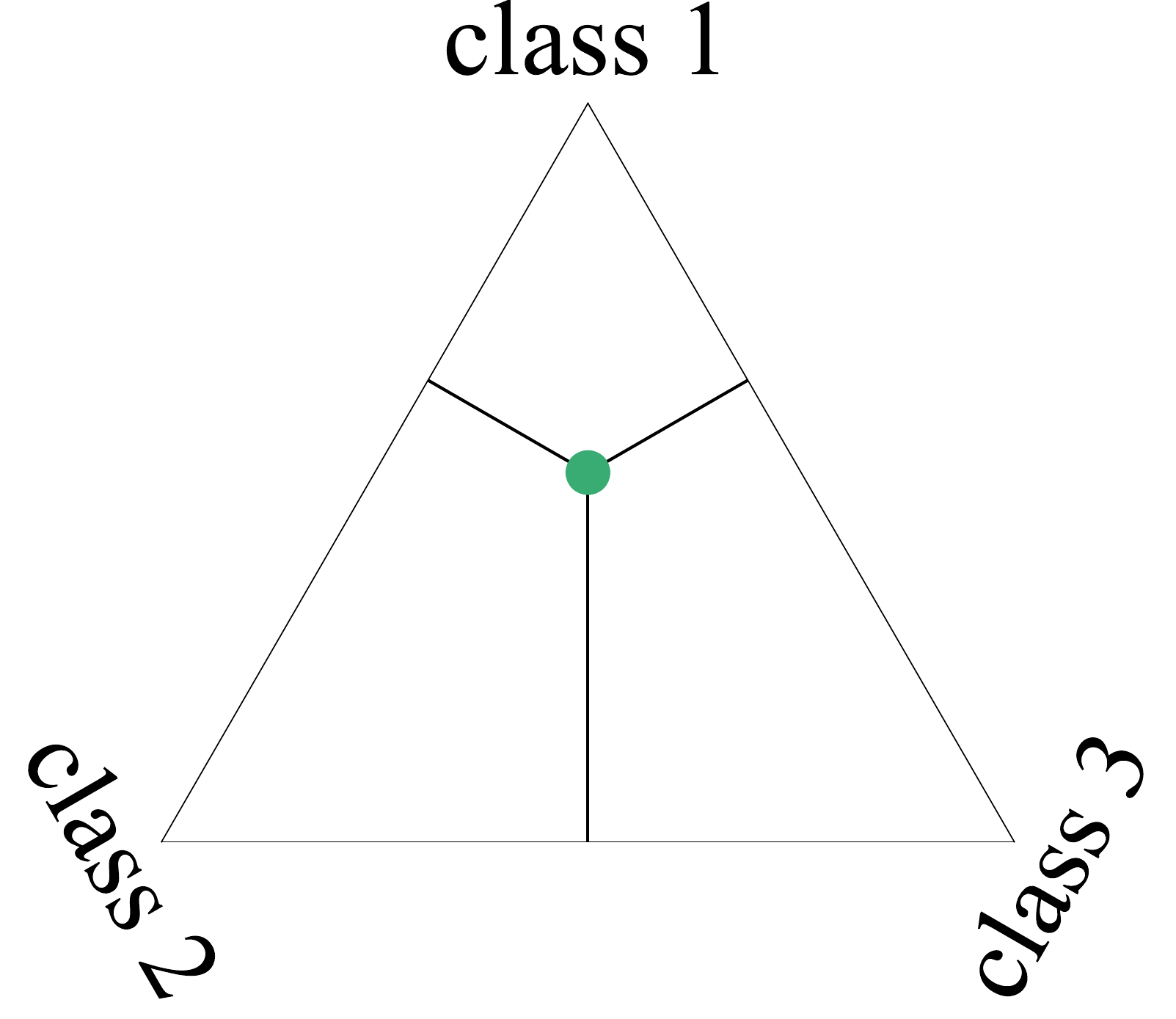}
\end{subfigure}
\hfill
\begin{subfigure}{0.25\columnwidth}
    \includegraphics[width=\textwidth]{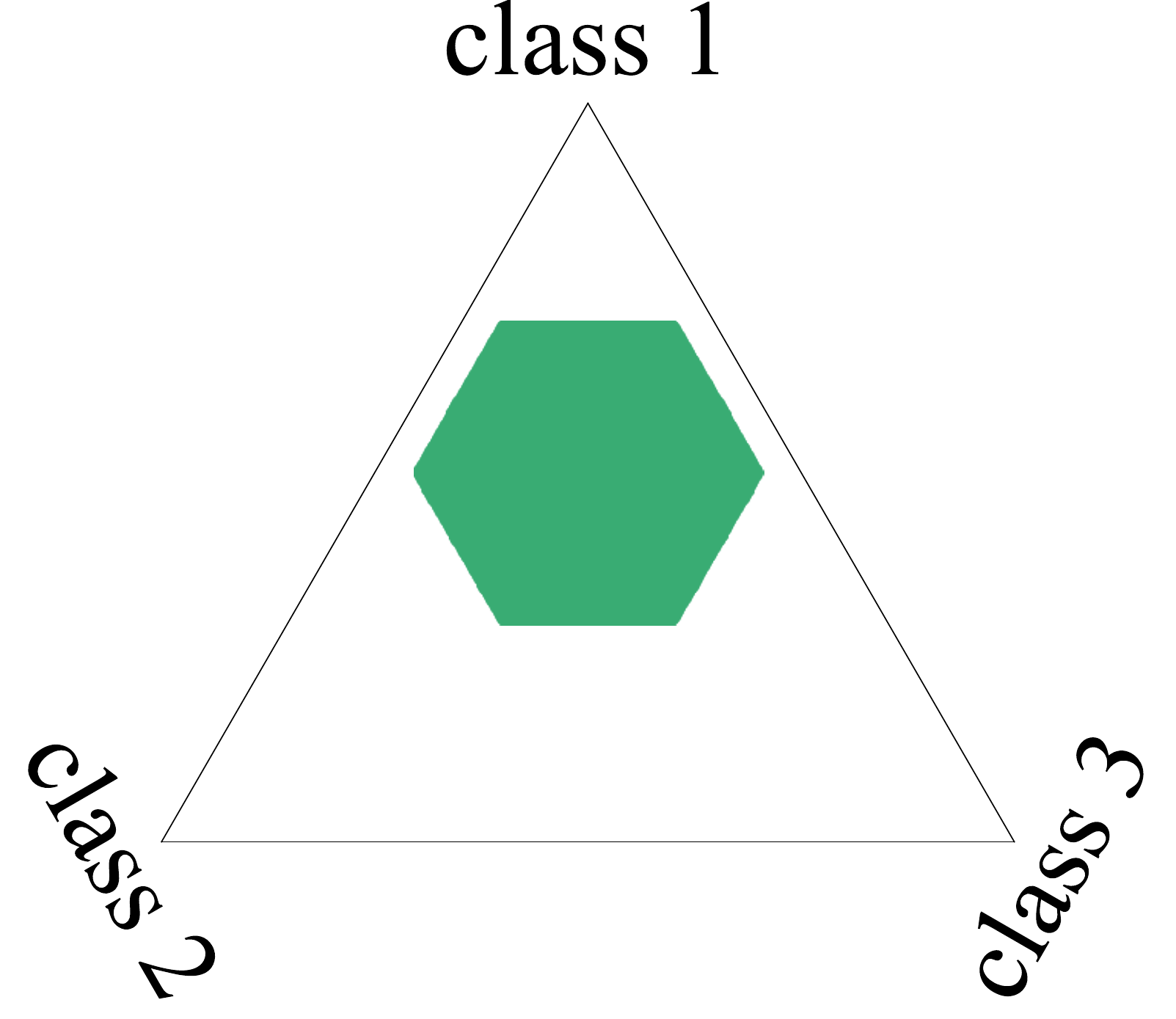}
\end{subfigure}
\hfill
\begin{subfigure}{0.25\columnwidth}
    \includegraphics[width=\textwidth]{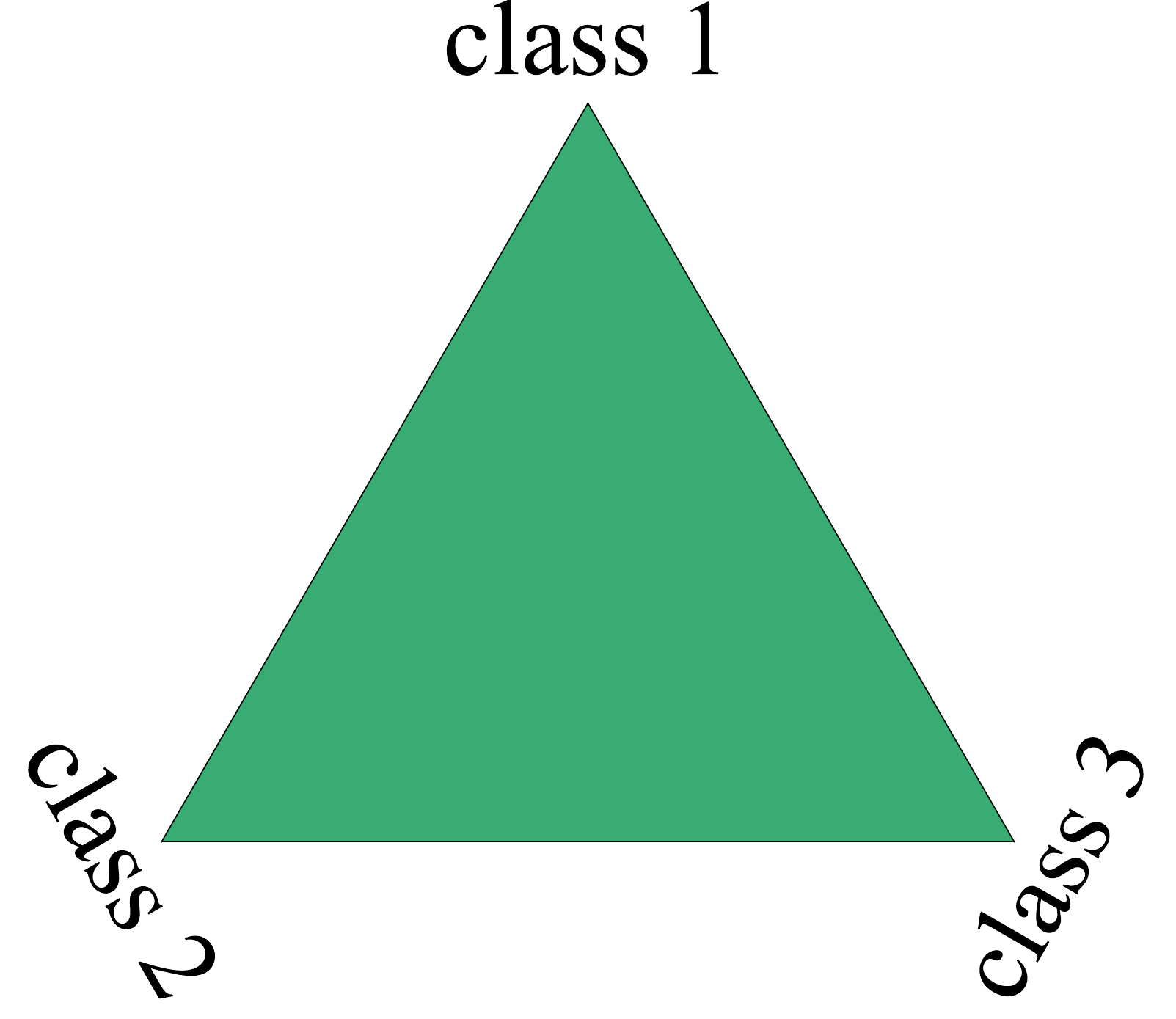}
\end{subfigure}
    \caption{For the three-class classification setting, the space of probability distributions can be illustrated by a two-dimensional simplex: each point in the simplex corresponds to a probability distribution so that credal sets can be depicted as regions.  The left case corresponds to the special case of a singleton (credal) set, i.e., a precise probability distribution, signifying aleatoric but no epistemic uncertainty. The case in the middle represents partial knowledge with a certain degree of (epistemic) uncertainty about the true distribution, and the right one corresponds to the case of complete ignorance, where nothing is known about the distribution.}
    \label{fig:credal}
\end{figure}

Representation of aleatoric and epistemic uncertainty requires formalism more expressive than standard probability distributions \citep{hullermeier2021aleatoric}. One such formalism which prevails in the recent ML literature is second-order probability distributions. Essentially, in a classification setting, these are distributions over distributions over classes. Models producing second-order distributions as predictions can be learned in a classical Bayesian way \citep{kendall2017uncertainties,depeweg2018decomposition} or using more recent approaches such as evidential deep learning \citep{sensoy2018evidential}. Yet, approaches of that kind are not unproblematic and have been subject to criticism \citep{bengs22pitfals,Bengs2023}. 
%Specifically, commonly used second-order loss functions have been shown to result in unfaithful estimation of epistemic uncertainty.
Another formalism suitable for representing both types of uncertainty is the concept of a \emph{credal set}, which is well-established in the field of imprecise probability theory \citep{walley1991statistical} and meanwhile also attracted attention in ML \citep{shaker2020aleatoric,hullermeier2022quantification}. Credal sets are (convex) sets of probability distributions that can be considered as candidates for an imprecisely known ground-truth distribution. Figure \ref{fig:credal} shows examples of credal sets in a three-class scenario, where the space of distributions can be visualized by the two-dimensional probability simplex. Broadly speaking, the larger the credal set, the higher the epistemic uncertainty, and the more ``in the middle'' the set is located, i.e., the closer it is to the uniform distribution, the higher the aleatoric uncertainty. 

\begin{figure*}[t!]
\begin{subfigure}{0.29\textwidth}
% \vspace{-100em}
\resizebox{\textwidth}{!}{
\adjustbox{valign=c}{
{\footnotesize 
\begin{tabularx}{\textwidth}{p{0.25\linewidth}X}
    \toprule
Premise & \textit{a child is on the ground crying.}\\ 
Hypothesis & \textit{A child on the ground crying as others look on.}\\ 
True dist. & $[0.06, 0.94, 0.00 ]$ \\ 
    \bottomrule
  \end{tabularx}
  }
  }
  }
\end{subfigure}
\begin{subfigure}{0.2\textwidth}
    \includegraphics[width=\textwidth, valign=c]{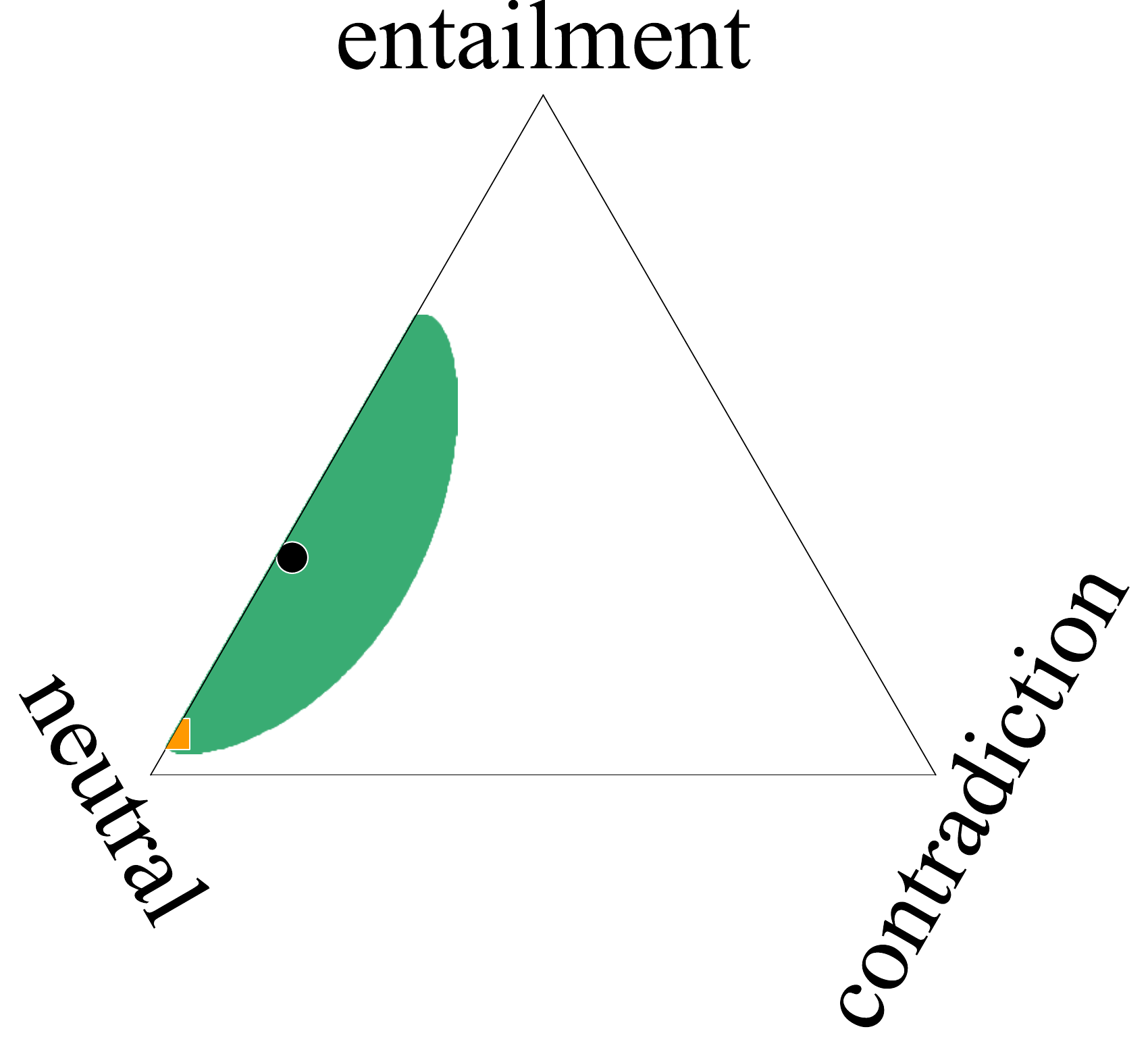}
\end{subfigure}
\hfill
\begin{subfigure}{0.29\textwidth}
% \vspace{-100em}
\resizebox{\textwidth}{!}{
\adjustbox{valign=c}{
{\footnotesize 
\begin{tabularx}{\textwidth}{p{0.25\linewidth}X}
    \toprule
Premise & \textit{uh-huh and is it true i mean is it um} \\ 
Hypothesis & \textit{It is absolutely correct.} \\ 
True dist. & $[0.31, 0.52, 0.17]$ \\ 
    \bottomrule
  \end{tabularx}
  }
  }
  }
\end{subfigure}
\begin{subfigure}{0.2\textwidth}
    \includegraphics[width=\textwidth, valign=c]{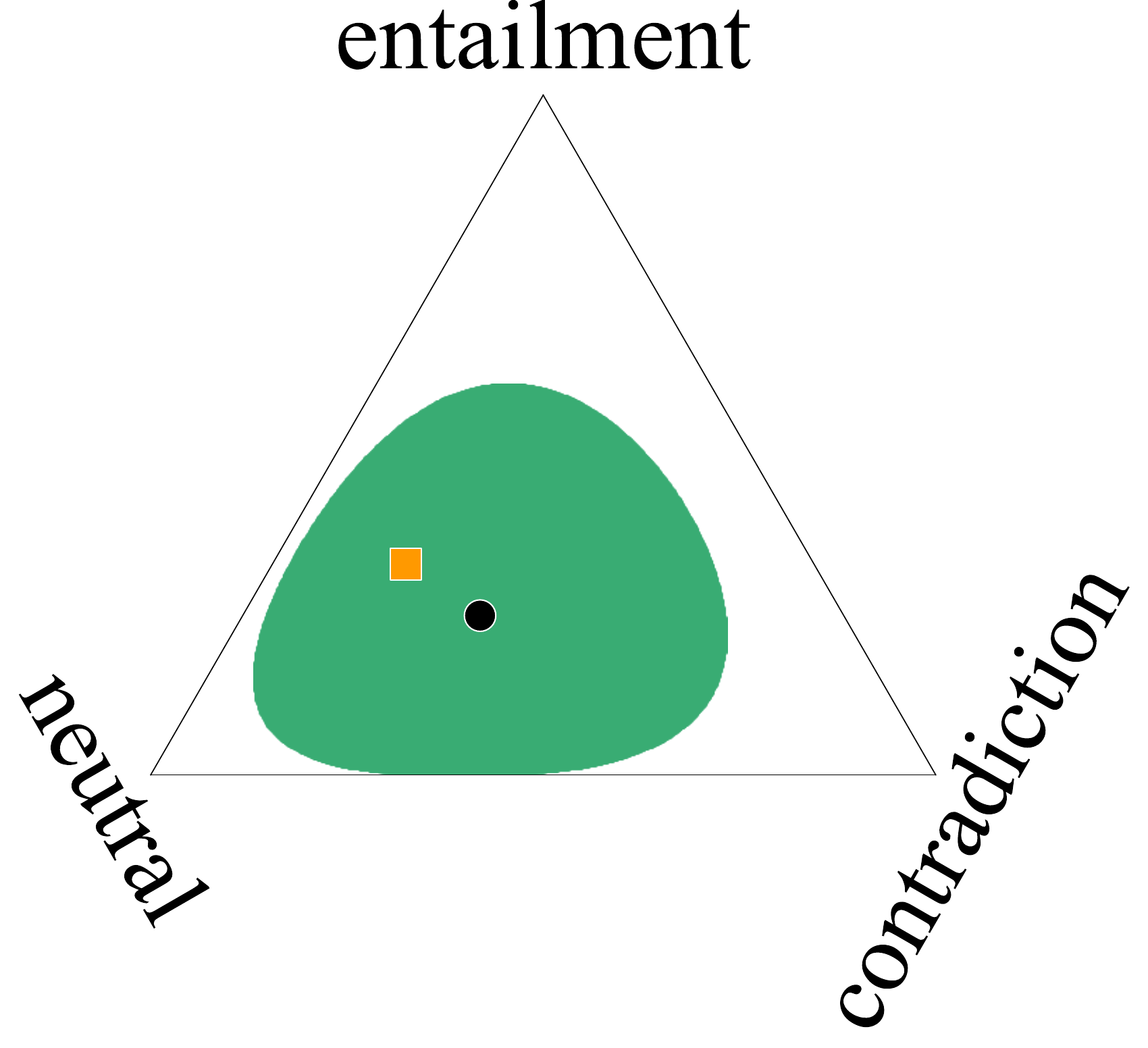}
\end{subfigure}
    \caption{An illustration of our proposed conformalized credal sets on two instances from the ChaosNLI dataset \citep{nie2020what}. {\color{credal_color}Green} regions indicate credal sets, while the true and the predicted distributions are marked with {\color{gt_color}orange} squares and black circles, respectively. }
    \label{fig:chaos:intro}
\end{figure*}
Learning to predict second-order representations, such as credal sets or second-order distributions, from standard ``zero-order'' supervised data\,---\,training instances together with observed class labels\,---\,is a difficult endeavor. For the case of second-order probabilities, it is even provably impossible to predict uncertainty in an ``unbiased'' way, i.e., without imposing strong prior assumptions on the epistemic uncertainty \citep{bengs22pitfals,Bengs2023}. 
In this paper, we assume ``first-order'' training data, i.e., instances associated with probability distributions over the class labels. In other words, instances are labeled probabilistically instead of being assigned a deterministic class label. Obviously, this type of data facilitates second-order learning. Not less importantly, it is becoming increasingly available in practice, for example, in the form of aggregations over multiple annotations per data instance, and hence increasingly relevant in applications \citep{uma2022scaling,stutz2023conformal} 

Our method leverages the framework of conformal prediction (CP), a non-parametric approach for set-valued prediction rooted in classical frequentist statistics \citep{vovk2022algorithmic}. Based on relatively mild assumptions, CP is able to provide theoretical guarantees in the form of marginal coverage: Predicted sets are guaranteed to cover the true target with high probability. Since our method inherits these coverage guarantees, our conformal credal sets are guaranteed to be valid with high probability (without any assumptions on model or distribution).

%Traditionally, CP assumes each data instance to be associated with an unambiguous, crisp ground truth label. 

%This assumption has been relaxed for the classification problem in the recent work \citep{javanmardi2023conformal, stutz2023conformal}. In particular, \citet{stutz2023conformal} assume access to an approximation of the ground truth conditional probability distribution (over label space) obtained from aggregating multiple annotations per data instance. These distributions inherently capture the aleatoric uncertainty, assuming sufficient annotators (such that annotation uncertainty can be ignored \citep{uma2022scaling, stutz2023evaluating}). However, the proposed Monte Carlo conformal prediction still seeks to produce confidence sets. While these sets are shown to capture the observed aleatoric uncertainty better, they still cannot distinguish between aleatoric and epistemic uncertainty.

Our main contribution is the proposal of a novel, conformal method to construct credal set predictors from first-order training data:
\begin{itemize}
% \item We assume access to first-order data in the form of examples with associated ground truth probability distributions or approximations thereof (e.g., relative frequencies originating from multiple annotators).
\item We propose a CP based method to construct conformal credal sets. To this end, we make use of two types of nonconformity functions based on distance resp.\ likelihood, and leveraging first-order resp.\ second-order probability predictors.

\item On ChaosNLI \citep{nie2020what}, a very ambiguous natural language inference task with multiple annotations per example, we show that our conformal credal sets are indeed valid, i.e., include the true ground truth distribution with high probability (see Figure \ref{fig:chaos:intro} for an illustration). We also compare the efficiency of predictions (size of the predicted sets) for different nonconformity functions.
\item We complement this study with controlled experiments on synthetic data, specifically investigating the performance of credal set prediction in the presence of label noise.  
\end{itemize}

\section{Related Work}

 %   \textbf{Credal sets.} {\color{red} maybe just mentioning some credal machine learning approaches while saying that they are different from what we are doing here.  }
%\citep{walley1996inferences, zaffalon2002naive, corani2008learning, abellan2003building, corani2012bayesian}

\textbf{Credal sets} are widely used as models for representing uncertainty, notably within the domain of imprecise probabilities \citep{walley1991statistical}. As already mentioned, they can represent both types of uncertainty, aleatoric and epistemic. In the context of data analysis and statistical inference, credal sets are often used as robust models of prior information, namely for modeling imprecise information about the prior in Bayesian inference \citep{walley1996inferences}.

In machine learning, credal sets have been used for generalizing some of the standard methods, including naive Bayes \citep{zaffalon2002naive,corani2008learning}, Bayesian networks \citep{corani2012bayesian}, and decision trees \citep{abellan2003building}. Typically, these approaches generalize simple frequentist inference to robust Bayesian inference, making use of an imprecise version of the Dirichlet model (a conjugate prior for the multinomial distribution). Compared to our approach, these methods are learning on standard (zero-order) training data. Moreover, despite representing uncertainty in predictions, they do not provide any formal guarantees. 

%broadening their applicability and usefulness in machine learning applications. \textit{Credal classifiers} \citep{zaffalon2002naive,corani2008learning}, based on the Imprecise Dirichlet Model (IDM) , \textit{credal networks} \citep{corani2012bayesian} (a generalization of Bayesian networks), and \textit{credal decision trees} \citep{abellan2003building} (an imprecise adaptation of decision trees) are eminent examples that leverage credal sets, showcasing their practical use in various ML applications. In this paper we focus on credal classification, in particular on the Naive Credal Bayes classifier (NCC) \citep{zaffalon2002naive,corani2008learning} and Imprecise Decision-Trees (ImpTree) \citep{abellan2003building}.

\textbf{Conformal prediction} \citep{vovk2022algorithmic}, briefly introduced in Section \ref{sec:cp} below, has recently gained attention for various applications in machine learning, especially for classification tasks \citep{sadinle2019least,romano2020classification,angelopoulos2020uncertainty,stutz2021learning,fisch2022conformal}. These methods mostly focus on split conformal prediction using a held-out calibration set \citep{papadopoulos2002inductive}, overcoming computational limitations of earlier transductive or bagging approaches \citep{vovk2022algorithmic,vovk2015cross,steinberger2016leave,barber2021predictive,linusson2020efficient}. While tackling classification tasks, our method for constructing conformal credal sets has more similarity with conformal regression \citep{romano2019conformalized, sesia2021conformal}, particularly in multivariate settings \citep{dietterich2022conformal}, as we essentially conformalize the simplex space of categorical distributions. Our conformity scores differ, however, in that they are specific for distributions rather than considering general multivariate spaces. This work also relates to work on appropriate measures of inefficiency \citep{vovk2017criteria} as measuring the inefficiency of our conformal credal sets is non-trivial. Most closely related to our work is the recent work by \citet{stutz2023conformal}, who consider conformal prediction in settings with high aleatoric uncertainty. However, we explicitly target the construction of conformal credal sets, while \citet{stutz2023conformal} mainly focus on constructing confidence sets of classes.

\textbf{First-order data.} In settings with high aleatoric uncertainty, labeling each example with a single, unique class is clearly insufficient. In practice, this is typically captured by high disagreement among annotators -- a problem particularly common in natural language tasks \citep{reidsma2008exploiting,aroyo2014three,aroyo2015truth,schaekermann2016resolvable,dumitrache2019crowdsourced,pavlick2019inherent,rottger2022two,abercrombie2023consistency}. Handling this disagreement has received considerable attention lately \citep{uma2021learning} as it offers to go beyond this zero-order information. For example, recent work on evaluation with disagreeing annotators \citep{stutz2023evaluating} argues the use of these annotations to get approximate first-order information for evaluation. This approach is becoming more and more viable with crowdsourcing tools \citep{kovashka2016crowdsourcing,sorokin2008utility,snow2008cheap} being an integral component of the benchmark, making multiple annotations per data instance more accessible. We follow a similar approach in our construction of conformal credal sets.

\section{Background}

\subsection{Supervised Learning and Predictive Uncertainty}
\label{sec:back1}

We consider the setting of (polychotomous) classification with label space $\mathcal{Y} = \{1,\ldots,K \} $ and an instance space $\mathcal{X}$. As usual, we assume an underlying data-generating process in the form of a probability distribution $P$ on $\mathcal{X} \times \mathcal{Y}$, so that observations $(X,Y)$ are i.i.d.\ samples from $P$. We denote by $\vec{\lambda}^{\vec{x}} \in \Delta^K$ the conditional probability distribution $P(\cdot \, | \, X = \vec{x})$, which we also consider as an element of the $(K-1)$-simplex 
$$
\Delta^K :=  \left\{ \vec{\lambda} = (\lambda_1, \ldots, \lambda_K)^\top \, | \, \lambda_k \geq 0, \| \vec{\lambda} \|_1 = 1 \right \} \subset \mathbb{R}^K \, .
$$
Thus, for each class label $k \in \mathcal{Y}$, the probability to observe $Y=k$ as an outcome for $\vec{x} \in \mathcal{X}$ is given by $\lambda^{\vec{x}}_k$.

Since the dependency between instances $X$ and outcomes $Y$ is non-deterministic, the prediction of $Y$ given $X = \vec{x}$ is necessarily afflicted with uncertainty, even if the ground-truth distribution $\vec{\lambda}^{\vec{x}}$ is known. As already said, this uncertainty is commonly referred to as aleatoric \citep{hullermeier2021aleatoric}. Intuitively, the closer $\vec{\lambda}^{\vec{x}}$ to the uniform distribution $\vec{p}_{\text{uni}} = (1/K, \ldots, 1/K)^\top$, the higher the uncertainty, and the closer it is to a degenerate (Dirac) distribution assigning all probability mass to a single class (a corner point in $\Delta^K$), the lower the uncertainty. Various measures have been proposed to quantify this uncertainty in numerical terms, with Shannon entropy as the arguably best-known representative \citep{depeweg2018decomposition}. 

Instead of assuming $\vec{\lambda}^{\vec{x}}$ to be known, suppose now that only a prediction $\hat{\vec{\lambda}}^{\vec{x}}$ of this distribution is available. Epistemic uncertainty refers to the uncertainty about how well the latter approximates the former, and hence to the additional uncertainty in the prediction of outcome $Y$ that is caused by the discrepancy between $\hat{\vec{\lambda}}^{\vec{x}}$ and the ground-truth $\vec{\lambda}^{\vec{x}}$. We seek to capture this discrepancy by means of credal sets 
$$
Q  \in \mathcal{Q}_K   \subset \Delta^K  \, ,
$$
with the idea that $Q \ni \vec{\lambda}^{\vec{x}} $ holds with high probability. Typically, credal sets are assumed to be convex, and further restrictions might be imposed on $\mathcal{Q}_K$ for practical and computational reasons, for example, a restriction to convex polygons (with a finite number of extreme points).

\subsection{Conformal Prediction}\label{sec:cp}

Conformal prediction provides a general framework for producing set-valued predictions with a certain guarantee of validity.
In a supervised setting, consider data points of the form $Z=(X,U) \in \mathcal{X} \times \mathcal{U}$, and the task is to predict $U$ given $X = \vec{x}$.
We assume the space $Z$ to be equipped with a nonconformity measure $f: \, \mathcal{Z} \fromto \mathbb{R}$ that quantifies the ``strangeness'' of $\vec{z}$, i.e., the higher $f(\vec{z})$, the less normal or expected the data point. Let $\mathcal{D}_\text{calib} \subset \mathcal{Z}$ be a (randomly generated) set of data points, called \emph{calibration data}, and $Z$ another data point that remains unobserved. Under the assumption of exchangeability, i.e., that the calibration data and the query point $Z$ have been generated by an exchangeable process, we want to construct a so-called confidence set $C \subseteq \mathcal{U}$ that guarantees coverage:
\begin{align}\label{eq:validity:cp}
    \mathbb{P}(U \in C) \geq 1-\alpha. 
\end{align}
By a simple combinatorial argument \citep{vovk2022algorithmic}, the confidence set $C$ can be constructed as
\begin{align}\label{eq:validity:cpu}
    C(\vec{x}):= \big\{ \,  u \in \mathcal{U} \, | \, f(\vec{x},u) < q(\mathcal{E}, \alpha') \, \big\} \, ,
\end{align}
where $\alpha' = |\mathcal{E}|^{-1}\lceil(1+|\mathcal{E}|)(1-\alpha)\rceil$, and $q(\mathcal{E};\alpha')$ denotes the $\alpha'$-quantile of $\mathcal{E}$.

Importantly, the guarantee \eqref{eq:validity:cpu} holds regardless of the nonconformity function $f(\cdot)$, which, however, has an influence on the \emph{efficiency} of the prediction: The more appropriate the function, the smaller the prediction set $C$ tends to be. Normally, $f(\cdot)$ is not predefined but constructed in a data-driven way using training data $\mathcal{D}_\text{train}$. For example, a common approach is to train a predictor $\pi: \, \mathcal{X} \fromto \mathcal{U}$ and then define $f(\vec{x}, u)$ in terms of $d( u , \pi(\vec{x}))$, where $d(\cdot, \cdot)$ is an appropriate distance function on $\mathcal{U}$. Replacing the point-prediction $\pi(\vec{x}) \in \mathcal{U}$ by the prediction set $C(\vec{x}) \subset \mathcal{U}$ can then be seen as ``conformalizing'' the predictor $\pi$: Using the calibration data, CP estimates a high-probability upper bound on the distance between point-predictions and actual outcomes, and corrects the former correspondingly.  
\section{Conformal Credal Set Prediction}\label{sec:credal}
Recall the setting and notation from Section \ref{sec:back1}. 
Our goal is to learn a credal set predictor $h:\, \mathcal{X} \fromto \mathcal{Q}_K$, that is, a model that makes predictions in the form of credal sets, thereby representing both aleatoric and epistemic uncertainty. To this end, we assume probabilistic training data of the form 
\begin{equation}\label{eq:data}
\mathcal{D} = \left\{ \big(\vec{x}_1, \vec{\lambda}^{\vec{x}_1} \big) , \ldots , \big(\vec{x}_N, \vec{\lambda}^{\vec{x}_N} \big) \right\} \subset \mathcal{X} \times \Delta^K \, .
\end{equation}
The model $h$ should be able to predict the (probabilistic) outcomes for new query instances in a reliable way. 
%More specifically, suppose that $(\vec{x}_{\text{new}}, \tilde{\vec{\lambda}}_{\text{new}})$ is a new data point following the same distribution as the training data $\mathcal{D}$. Then, the prediction $Q = h(\vec{x}_{\text{new}})$ should be such that $Q \ni \tilde{\vec{\lambda}}_{\text{new}}$ with high probability. 
More specifically, suppose that $\vec{x}_{\text{new}}$ is a new query instance (following the same distribution as the training data) for which a prediction is sought. The credal prediction $Q = h(\vec{x}_{\text{new}})$ should then be valid in the sense that $Q \ni \vec{\lambda}^{\vec{x}_\text{new}}$ with high probability. 
At the same time, the prediction should be informative in the sense that the (epistemic) uncertainty reflected by $Q$ is as small as possible. Again, various measures for quantifying the latter can be found in the literature \citep{klir1999uncertainty,sale2023volume}. 

We aim to construct the credal set predictor $h$ by means of (inductive) conformal prediction. Following the conformalization recipe outlined in Section \ref{sec:cp}, we partition $\mathcal{D}$ into $\mathcal{D}_\text{train}$ and $\mathcal{D}_\text{calib}$, using the former for model training and the latter for calibration. Regarding the training step, we explore two learning strategies, connected with two ways of defining a nonconformity function, which is pivotal in the calibration step.

%defining nonconformity functions is , and the choice depends on the type of predictor used for learning.  

The first approach is based on training a standard (\textit{first-order}) probability predictor, i.e.,  a probabilistic classifier $g: \, \mathcal{X} \fromto \Delta^K$ that maps instances to the (first-order) probability distribution on $\mathcal{Y}$. This can be achieved, for example, by minimizing the cross-entropy loss between the ground truth and the predicted distributions, i.e.,
\begin{align*}
 g = \argmin_{\Bar{g} \in \mathcal{H}}   \sum_{(\vec{x}_i, \vec{\lambda}^{\vec{x}_i}) \in \mathcal{D}_\text{train}}  - \sum_{k=1}^{K} \lambda^{\vec{x}_i}_{k} \log(\Bar{g}(\vec{x}_i)_k), 
\end{align*}
where $\mathcal{H}$ is a hypothesis space. Given a predictor $g(\cdot)$ of this kind, nonconformity is naturally defined in terms of distance:
\begin{align}\label{eq:nonconformity:first}
    f_1(\vec{x}, \vec{\lambda}^{\vec{x}}):=d(\vec{\lambda}^{\vec{x}}, g(\vec{x})) \, ,   
\end{align}
where $d(\cdot, \cdot)$ is a suitable distance function on $\Delta^K$, such as total variation, Wasserstein distance, etc.

An alternative approach is motivated by recent work on (epistemic) uncertainty representation via \textit{second-order} probability distributions.  
A second-order learner $G: \, \mathcal{X} \fromto \mathbb{P}(\Delta^K)$ maps each input $\vec{x}$ to a distribution over $\Delta^K$. Given the training data, meaningful learning in this context can be accomplished, for instance, by parameterizing the second-order distributions using Dirichlet distributions. Specifically, one can assume that each $\vec{x}$ is associated with a Dirichlet distribution characterized by the parameter vector $\vec{\theta}^{\vec{x}} \in \mathbb{R}^K_+$ with the probability density function 
\begin{align}
    P(\vec{\lambda} \, | \, \vec{\theta}^{\vec{x}}) = \frac{1}{B({\vec{\theta}^{\vec{x}}})} \prod_{k=1}^{K} \lambda_k^{\theta^{\vec{x}}_k - 1} \, ,
\end{align}
where $B(\cdot)$ is the multivariate beta function. This way, $\vec{\lambda}^{\vec{x}}$ can be thought of as a sample from that distribution, i.e., $\vec{\lambda}^{\vec{x}} \sim \text{Dir}(\vec{\theta}^{\vec{x}})$. Our model then essentially yields a prediction $\hat{\vec{\theta}}^{\vec{x}}$ of the true parameter $\vec{\theta}^{\vec{x}}$ for every $\vec{x}$, and its optimization involves minimizing the negative log-likelihood loss 
\begin{align*}
          \sum_{(\vec{x}_i, \vec{\lambda}^{\vec{x}_i}) \in \mathcal{D}_\text{train}}\bigg(\log (B(\hat{\vec{\theta}}^{\vec{x}_i} )) - \sum_{k=1}^{K} (\hat{\theta}^{\vec{x}_i}_k - 1)\log(\lambda^{\vec{x}_i}_{k})\bigg).
\end{align*}
Given a second-order predictor $\hat{\vec{\theta}}^{\vec{x}}$, nonconformity can be defined as a decreasing function of likelihood, e.g., as 1 minus relative likelihood:
%(proportion of the predicted likelihood of the true distribution over the maximum predicted likelihood), i.e., 
\begin{align}\label{eq:nonconformity:second}
    f_2(\vec{x}, \vec{\lambda}^{\vec{x}}) = 1 - \frac{P(\vec{\lambda}^{\vec{x}} \, | \, \hat{\theta}^{\vec{x}})}{\max_{\vec{\lambda} \in \Delta^K} P(\vec{\lambda} \, | \, \hat{\theta}^{\vec{x}})}.
\end{align}
%This choice is justified by the intuitive notion that the ideal scenario is when the mode of the predicted second-order distribution aligns with the ground-truth first-order distribution. 

Using the nonconformity function $f_i(\cdot)$ ($i \in \{ 1, 2 \}$), we obtain the set of nonconformity scores by 
\begin{align}\label{eq:nonconformity:both}
    \mathcal{E}_i:= \Big\{ f_i(\vec{x}_j, \vec{\lambda}^{\vec{x}_j})  \, | \, (\vec{x}_j, \vec{\lambda}^{\vec{x}_j}) \in \mathcal{D}_\text{calib} \Big\}.   
\end{align}
Accordingly, the credal set can be defined as 
\begin{align}\label{eq:credal_set:both}
    h_i(\vec{x}_{\text{new}}) := \big\{ \,  \vec{\lambda} \in \Delta^K \, | \, f_i(\vec{x}_{\text{new}}, \vec{\lambda}) < q(\mathcal{E}_i, \alpha') \, \big\} \, .
\end{align}

\begin{algorithm}[t!]
   \caption{Conformal Credal Set Prediction}
   \label{alg:general}
\begin{algorithmic}
   \STATE {\bfseries Input:} 
   \STATE 
   \begin{itemize}
       \item[] Data $\mathcal{D}$; error rate $\alpha$; query instance $\vec{x}_{\text{new}}$.
   \end{itemize}
   \STATE {\bfseries Process:}
   \STATE
   \begin{itemize}
       \item[] Partition $\mathcal{D}$ into $\mathcal{D}_\text{train}$ and $\mathcal{D}_\text{calib}$.
       \item[] Train a first-order ($i=1$) or a second-order ($i=2$) predictor using $\mathcal{D}_\text{train}$.
       \item[] Choose a nonconformity function $f_i$ as in \eqref{eq:nonconformity:first} or \eqref{eq:nonconformity:second} that suits the trained predictor to obtain the set of scores $\mathcal{E}_i$. 
       \item[] Set $\alpha' = |\mathcal{E}_i|^{-1}\lceil(1+|\mathcal{E}_i|)(1-\alpha)\rceil$. 
   \end{itemize}
   \STATE {\bfseries Output:}
   \begin{itemize}
       \item[] $h_i(\vec{x}_{\text{new}}) = \big\{ \,  \vec{\lambda} \in \Delta^K \, | \, f_i(\vec{x}_{\text{new}}, \vec{\lambda}) < q(\mathcal{E}_i, \alpha') \, \big\} \, .$
   \end{itemize}
   % \begin{itemize}
   %     \item[] $h_i(\vec{x}_{\text{new}}) = \big\{ \,  \vec{\lambda} \in \Delta^K \, | \, f_i(\vec{x}_j, \vec{\lambda}^{\vec{x}_j}) < q(\mathcal{E}_i, \alpha') \, \big\} \, .$
   % \end{itemize}
\end{algorithmic}
\end{algorithm}
Algorithm \ref{alg:general} outlines a summary of the proposed methods. In the following theorem, we state the validity of the predicted set, that is, the restatement of the conformal coverage guarantee \citep{vovk2022algorithmic} adjusted to our setting. 
\begin{theorem}
        Let $\mathcal{P}$ denote the joint probability distribution on $(X, \Lambda) \in \mathcal{X} \times \Delta^K$. If data points in $\mathcal{D}_\text{calib}$ and $(\vec{x}_\text{new},\vec{\lambda}^{\vec{x}_\text{new}})$ are drawn i.i.d. (exchangeably) from $\mathcal{P}$, then the conformal credal sets in \eqref{eq:credal_set:both} 
        %and \eqref{eq:credal_set:second} 
        are valid, i.e., 
        \begin{align*}
            \mathbb{P} \big( \vec{\lambda}^{\vec{x}_\text{new}} \in h_i(\vec{x}_\text{new}) \big)\geq 1- \alpha \, , 
            ~ \text{ for } i \in \{ 1, 2 \}.
        \end{align*}
\end{theorem}
\subsection{Noisy Observations}\label{sec:noisy}

So far, we (implicitly) assumed that ground-truth probability distributions $\vec{\lambda}^{\vec{x}_i}$ will be provided as training (and calibration) data. Needless to say, this assumption will rarely hold true in practice. Instead, observations will rather be noisy versions $\tilde{\vec{\lambda}}^{\vec{x}_i}$ of the true probabilities, i.e., the data will be of the form 
%Apart from synthetically created datasets, obtaining datasets in the form of \eqref{eq:data} is exceedingly challenging, if not practically impossible, in the real world. Instead, we can have a dataset of the form
\begin{equation}\label{eq:noisy_data}
\mathcal{D} = \left\{ \big(\vec{x}_1, \tilde{\vec{\lambda}}^{\vec{x}_1} \big) , \ldots , \big(\vec{x}_N, \tilde{\vec{\lambda}}^{\vec{x}_N} \big) \right\} \subset \mathcal{X} \times \Delta^K \, .
\end{equation}
%where $\tilde{\vec{\lambda}^{\vec{x}_i}}$ denotes the noisy version of the true distribution ${\vec{\lambda}^{\vec{x}_i}}$. 
Notably, such datasets emerge in scenarios where each data instance $\vec{x}$ is annotated by multiple human experts, which recently have attracted a lot of attention in the context of machine learning and also conformal prediction \citep{stutz2023conformal, javanmardi2023conformal}. In this context, $\tilde{\vec{\lambda}}^{\vec{x}}$ denotes the distribution derived from aggregating annotator disagreements concerning the label of instance ${\vec{x}}$.
%With the dataset $\mathcal{O}$ in hand, constructing a credal set predictor $h$ is feasible using the guidelines outlined in the preceding section. However, the conformal guarantee needs to be reformulated to 
Of course, conformal prediction can still be applied to noisy data of that kind, but the coverage guarantee will then only hold for the noisy labeling:
\begin{align}\label{eq:covn}
    \mathbb{P} \left( \tilde{\vec{\lambda}}^{\vec{x}_\text{new}} \in h(\vec{x}_\text{new}) \right)\geq 1-\tilde{\alpha}.
\end{align}
%This implies that the credal set ensures coverage of the noisy version of the true distribution $\vec{\lambda}^{\vec{x}_\text{new}}$ for a given test instance $\vec{x}_\text{new}$. 
%One possible argument is that, asymptotically, the noisy distributions should converge to their noise-free counterparts as the number of annotators increases. Consequently, the credal sets regain validity by effectively covering the ground truth distributions. 
%Another approach is to establish more conservative credal sets when an upper bound on the distance between noisy and true distributions is provided, accommodating for the introduced noise.
%{\color{red} Here, we can talk about the fact that given noisy data, the learner (for instance, first-order predictor) tries to capture the dependency between X and Y such that g(x) is close to $\lambda^x$. Now, if we use noisy data in calibration, the nonconformity would probably be higher than the case we would have used the ground truth. Therefore, our credal sets would be more conservative in a sense. }

Practically, one may expect that the guarantees will hold for the ground-truth as well, simply because calibration on noisy instead of clean data will tend to make prediction regions larger and hence more conservative. Moreover, since nonconformity is derived from a predictive model $g(\cdot)$ that seeks to recover ground-truth probabilities, the latter should conform at least as well as noisy distributions.  
Of course, this intuition is not a formal guarantee. In order to provide such a guarantee for the ground-truth probabilities, one obviously needs to make some assumptions. Concretely, let us make the following \emph{bounded noise} assumption for the labeling process: The labeling noise is (stochastically) bounded in the sense that, given the nonconformity function $f$ and a (small) probability $\delta > 0$, there exists a tolerance $\epsilon  > 0$ such that
     \begin{equation}\label{eq:close}
     \mathbb{P}\left(|f(\vec{x}, \vec{\lambda}^{\vec{x}})-f(\vec{x}, \tilde{\vec{\lambda}}^{\vec{x}} ) | < \epsilon \right) \geq 1 - \delta
     \end{equation} 
     all $\vec{x} \in \mathcal{X}$.  

 \begin{theorem}\label{theorem:noisy}
Let $\alpha > 0$ be any miscoverage rate, and suppose the bounded noise assumption holds. 
Let $q = q(\mathcal{E}, \tilde{\alpha})$ be the critical threshold on the noisy calibration data $\mathcal{D}_{\text{calib}}$ for miscoverage rate 
$$
\tilde{\alpha} = \frac{\alpha - \delta}{1-\delta} \, .
$$
Then, for any new query $\vec{x}_\text{new} \in \mathcal{X}$,    
     \begin{align*}
    \mathbb{P} \big( f( \vec{x}_\text{new}, {\vec{\lambda}}^{\vec{x}_\text{new}} ) <  q + \epsilon  \big) \geq 1-\alpha \, .
    \end{align*}
 \end{theorem}
The proof is deferred to \ref{sec:appendix:proof}. As a consequence of this result, a conformal predictor learned on the noisy data with modified miscoverage rate $\tilde{\alpha}$ can be turned into a valid predictor (with miscoverage rate $\alpha$) for the ground-truth data by increasing the learned rejection threshold by $\epsilon$, provided the bounded noise property (\ref{eq:close}) can be ascertained. Thus, if we denote the corresponding credal set predictor by $h_\epsilon$, we can guarantee that
\begin{align}\label{eq:covt}
    \mathbb{P} \big( \vec{\lambda}^{\vec{x}_\text{new}} \in h_\epsilon(\vec{x}_\text{new}) \big)\geq 1- \alpha \, .
\end{align}

\section{Experiments}
In this section, we evaluate the performance of our proposed methods using both synthetic and real datasets. In vanilla conformal prediction, the performance of a method is usually assessed based on the average prediction set size, aka \textit{efficiency}, and the average \textit{coverage} on the test set. It is more appealing to have the promised coverage with smaller sets. In our scenario, the analytical calculation of credal sets is not feasible. Therefore, for the sake of illustration as well as other analyses, such as efficiency calculation, we resort to approximations. We discretize the simplex with a resolution of $0.005$, yielding $M=19969$ distributions. This enables the straightforward construction of credal sets, as defined in \eqref{eq:credal_set:both}. The efficiency is gauged by considering the fraction of all $M$ distributions that lie within the predicted credal sets. All implementations and experiments can be found in the technical supplement of this work.\footnote{{The link to the code: \url{https://github.com/alireza-javanmardi/conformal-credal-sets}}}   

\subsection{Learning Model}
\begin{table}[t!]
\centering
\renewcommand{\arraystretch}{1.1}
    \caption{Summary of the nonconformity functions used in experiments.}\label{tab_setup}
    \vspace{0.5em}
    \adjustbox{max width=\columnwidth}{%
    \begin{tabular}{lcc}
    \toprule
    \textbf{Name} & \textbf{Formulation} & \textbf{Predictor}  \\
    \midrule
    \textbf{TV} & $\frac{1}{2} \sum_{k=1}^K |\lambda_{k}^{\vec{x}} - g(\vec{x})_k|$ & first-order \\
    \textbf{WS} & - & first-order \\
    \textbf{KL} & $\sum_{k=1}^K \lambda_{k}^{\vec{x}} \log(\frac{\lambda_{k}^{\vec{x}}}{g(\vec{x})_k})$ & first-order \\
    \textbf{Inner} & $1 - \sum_{k=1}^K \lambda_{k}^{\vec{x}}g(\vec{x})_k$ & first-order \\
    \textbf{SO} & as in \eqref{eq:nonconformity:second} & second-order \\
    % \rowcolor{tablegray}\emph{SOUM} & Synthetic & -- & 20, 40 & $[0, 1]$  \\
    % \emph{LM} & DistilBert & Token Removal & 14 & $[-1, 1]$  \\
    % \rowcolor{tablegray}\emph{CH} & Neural Net & Mean & 8 & $\mathbb{R}$ \\
    % \emph{BR} & XGBoost & Mean/Mode & 12 & $\mathbb{R}$ \\
    % \rowcolor{tablegray}\emph{ViT} & ViT-32-384 & Token Removal & 16 & $[0, 1]$ \\
    % \emph{CNN} & ResNet18 & Superpixel & 14 & $[0, 1]$ \\
    % \rowcolor{tablegray}\emph{AC} & RF & Mean/Mode & 14 & $[0, 1]$ \\
    \bottomrule
    \end{tabular}}
    \label{tab:nonconformity}
\end{table} 
In our experiments, we employ a deep neural network as the learner. Specifically, the model consists of three hidden layers with $256$, $64$, and $16$ units, utilizing ReLU as the activation function. Prior to the output layer, a dropout layer with a rate of $0.3$ is incorporated. The same model architecture serves both first- and second-order predictors, differing only in the activation functions of the output layers. For the first-order predictor, softmax is used, while for the second-order predictor, ReLU is employed. Learning is facilitated using the Adam optimizer with a learning rate of $10^{-4}$, utilizing cross-entropy as the loss function for the first-order predictor and negative log-likelihood for the second-order predictor.
\subsection{Nonconformity Functions} 
CP should work regardless of the choice of nonconformity score function, while this choice can affect the efficiency and geometry of the prediction set. For the sake of comparison, we examine different nonconformity functions in our experiments. When utilizing a first-order predictor, besides total variation (\textbf{TV}) and the First Wasserstein (\textbf{WS}) distance, we also investigate the Kullback–Leibler (\textbf{KL}) divergence and 1 minus the inner product (\textbf{Inner}) as nonconformity functions. For the second-order predictor, we consider 1 minus the relative likelihood (\textbf{SO}) as defined in \eqref{eq:nonconformity:second}. Table \ref{tab:nonconformity} offers a summary of all five nonconformity functions employed in our experiments.
% Nevertheless, for the sake of comparison, we consider the following nonconformity functions throughout our experiments: 
% \begin{itemize}
%     \item Total Variation distance (\textbf{TV}): 
%     $$
%     E_\text{TV}(x_j, \lambda_j) = \frac{1}{2} \sum_{k=1}^K |\lambda_{j,k} - f(x_j)_k|.
%     $$
%     \item Kullback–Leibler divergence (\textbf{KL}): aka relative entropy 
%     $$
%     E_\text{KL}(x_j, \lambda_j) = \sum_{k=1}^K \lambda_{j,k} \log(\frac{\lambda_{j,k}}{f(x_j)_k}).
%     $$
%     \item First Wasserstein distance (\textbf{WS}):
%     \item Inner product (\textbf{Inner}):
%     $$
%     E_\text{Inner}(x_j, \lambda_j) = \sum_{k=1}^K \lambda_{j,k}f(x_j)_k.
%     $$
% \end{itemize}
\subsection{Real Data}
We focus on the ChaosNLI dataset \citep{nie2020what}, an English Natural Language Inference (NLI) dataset that captures the inherent variability in human judgments of textual entailment. Here, the classes are \textit{entailment}, \textit{neutral}, and \textit{contradiction} for each premise-hypothesis pair. Instances in this dataset are selected from the development sets of SNLI \citep{bowman2015large}, MNLI \citep{williams2018broad}, and AbductiveNLI \citep{bhagavatula2019abductive}, for which the majority vote was less than three among the five human annotators. These instances were then given to $100$ independent humans for annotation, given strict annotation guidelines. 
% We apply our CP algorithm to the predictions made by \citet{baan2022stop}.

\begin{figure*}[t!]
    \centering
    \begin{tabularx}{\textwidth}{m{0.3\textwidth}YYYYY}
       \toprule
        Premise-Hypothesis pair & TV & KL & WS & Inner & SO\\
        \midrule
         {\color{blue}Premise:} \textit{A man in a bar drinks from a pitcher while a man in a green hat looks on and a woman in a black shirt drink from a glass.}
         
         {\color{blue}Hypothesis:} \textit{The women is drinking water.}
         & \includegraphics[width=0.12\textwidth, valign=c]{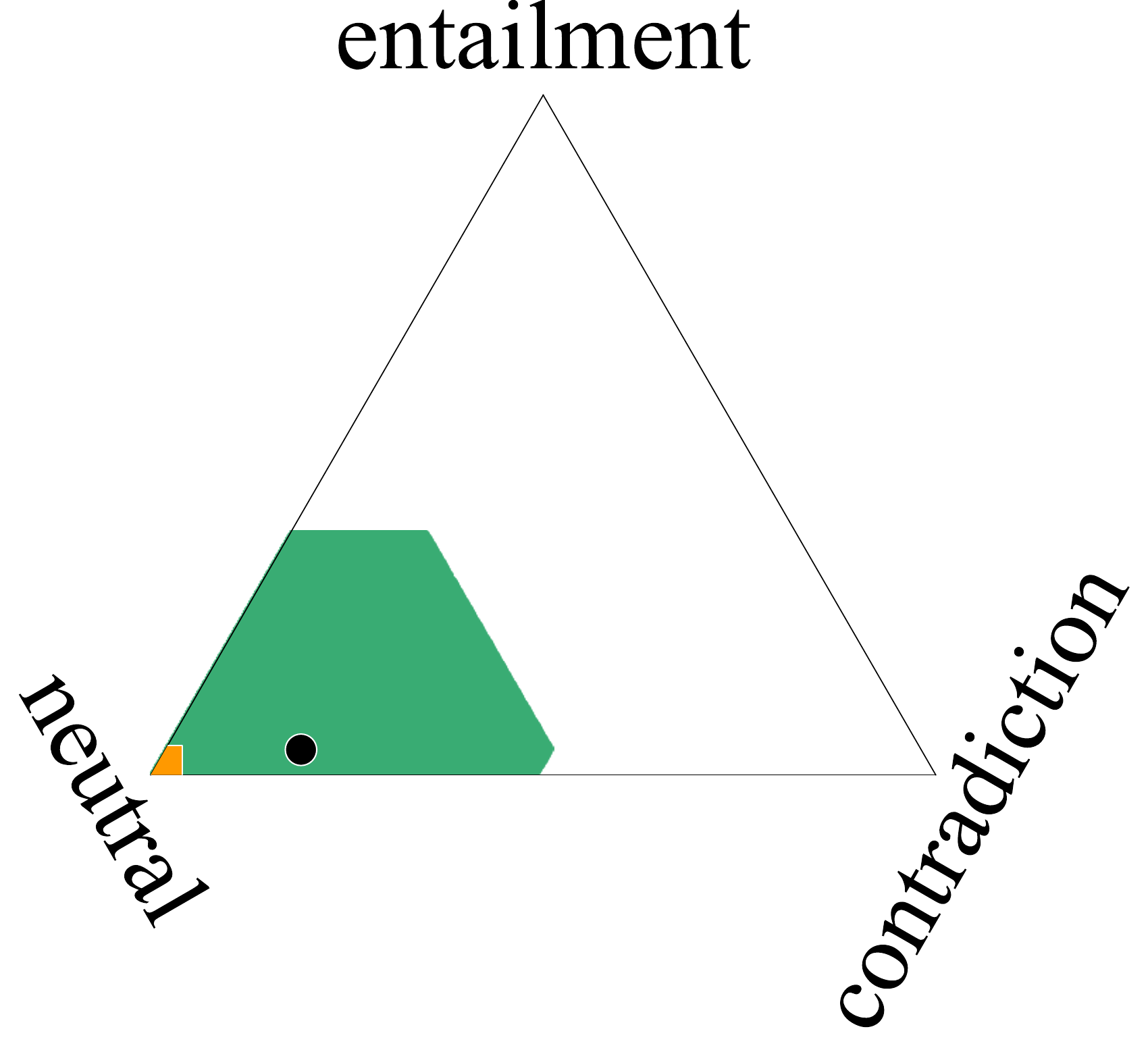} & 
        \includegraphics[width=0.12\textwidth, valign=c]{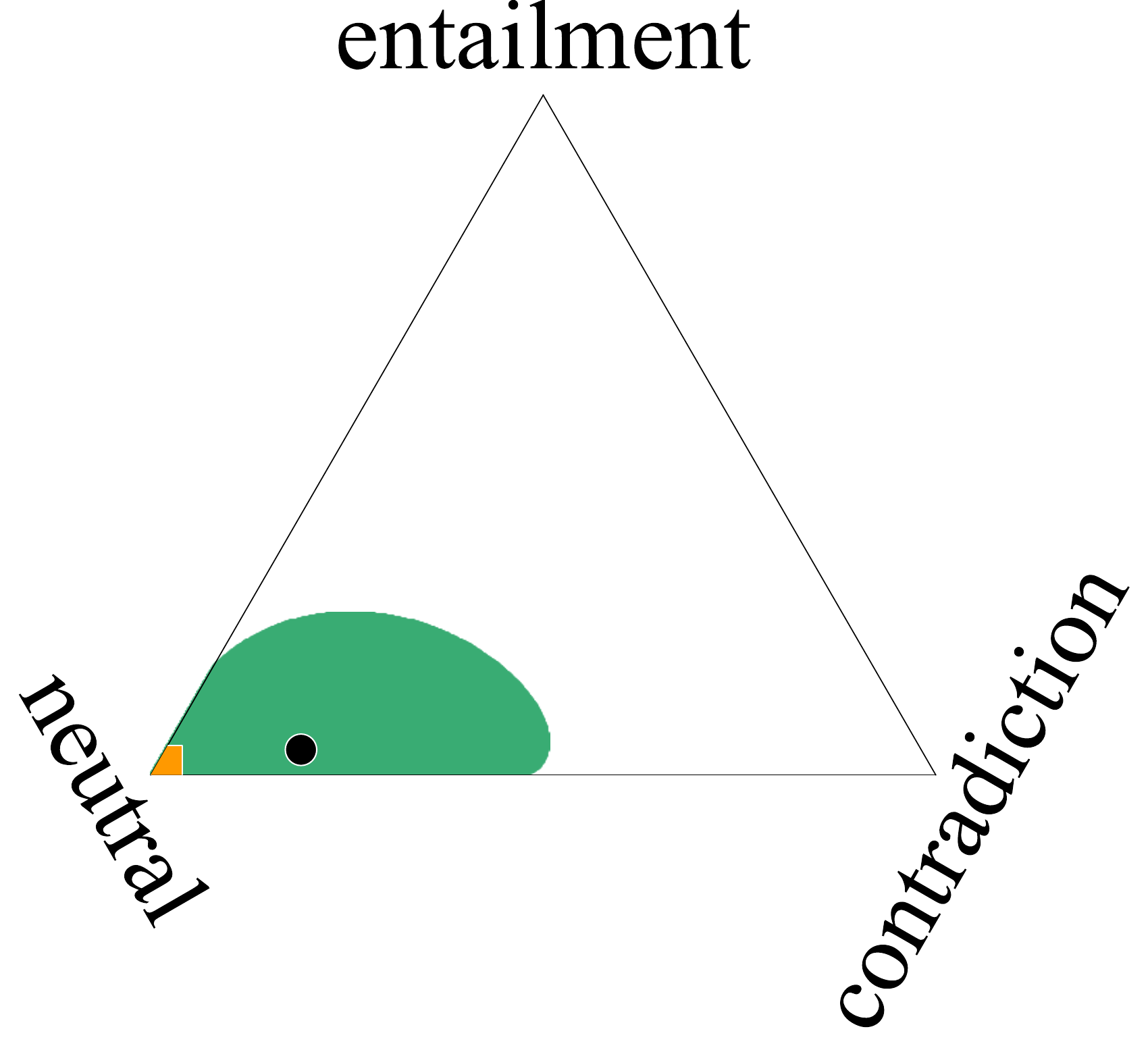} & 
        \includegraphics[width=0.12\textwidth, valign=c]{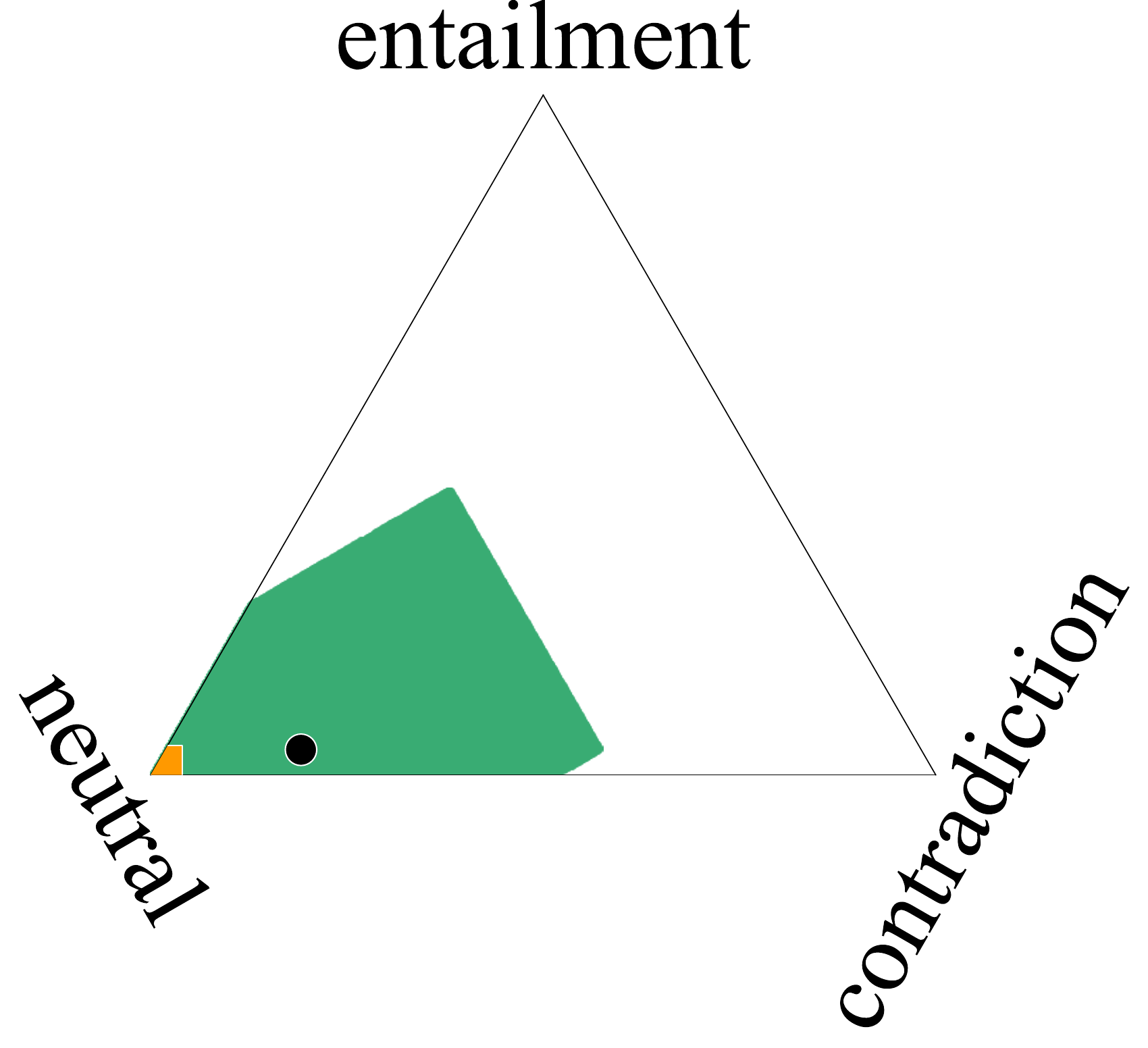} & 
        \includegraphics[width=0.12\textwidth, valign=c]{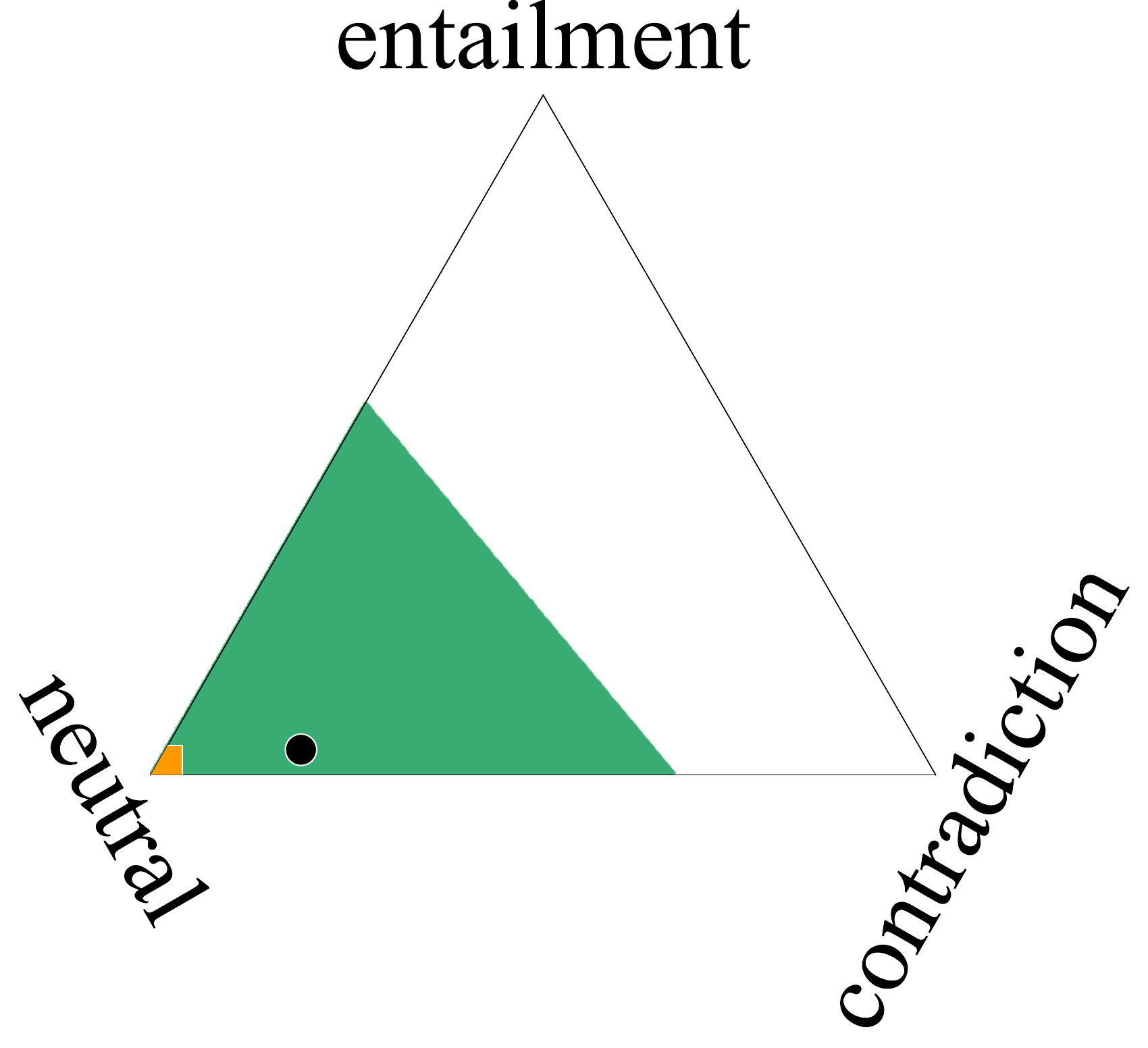} & 
        \includegraphics[width=0.12\textwidth, valign=c]{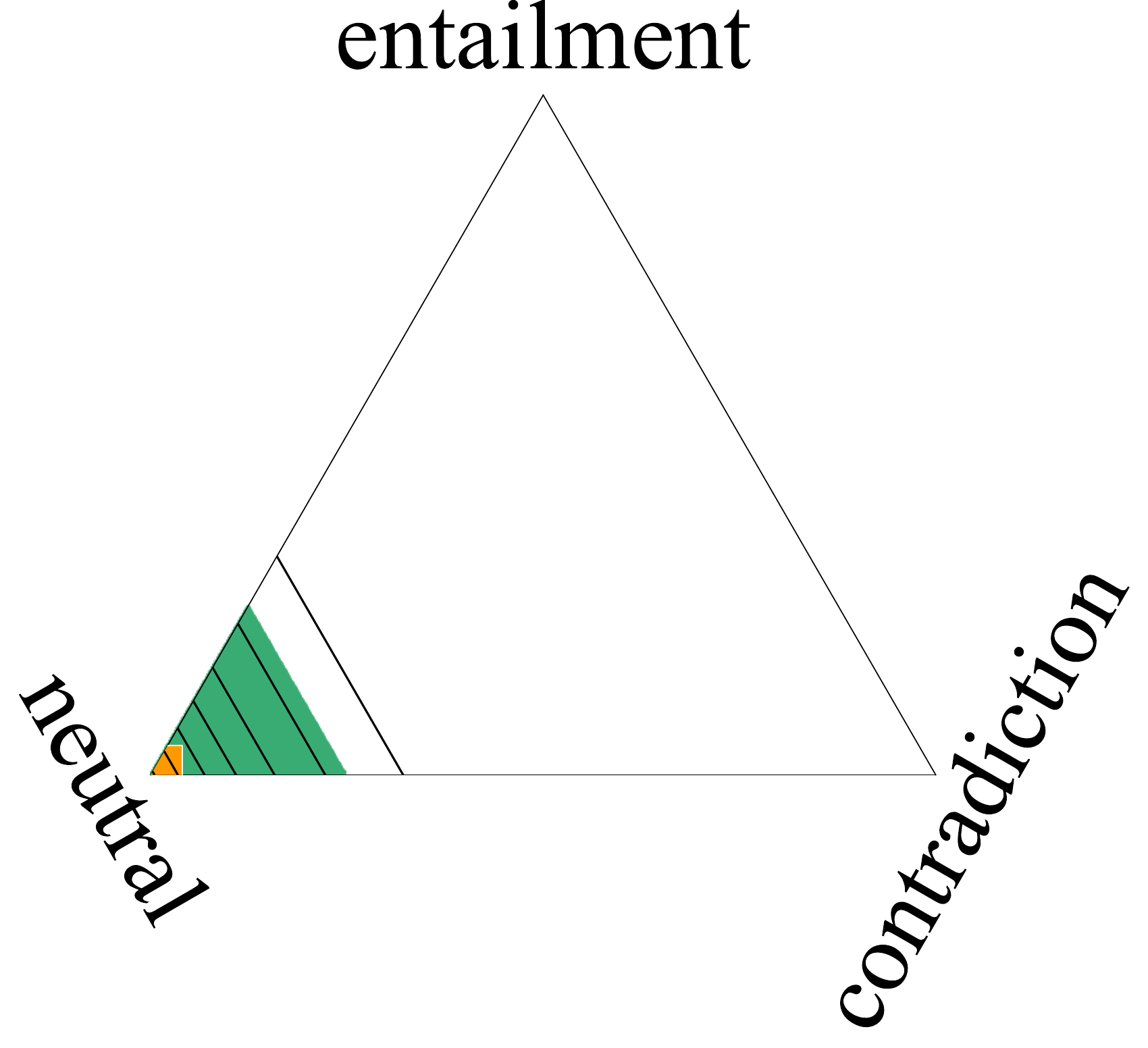} \\
        & $0.24$ & $0.18$ & $0.28$ & $0.36$ & $0.06$ \\
        % & \includegraphics[width=0.12\textwidth, valign=c]{figures/chaosNLI/seed_0_k_2/alpha_0.1/tv.pdf} & 
        % \includegraphics[width=0.12\textwidth, valign=c]{figures/chaosNLI/seed_0_k_2/alpha_0.1/kl.pdf} & 
        % \includegraphics[width=0.12\textwidth, valign=c]{figures/chaosNLI/seed_0_k_2/alpha_0.1/ws.pdf} & 
        % \includegraphics[width=0.12\textwidth, valign=c]{figures/chaosNLI/seed_0_k_2/alpha_0.1/inner.pdf} & 
        % \includegraphics[width=0.12\textwidth, valign=c]{figures/chaosNLI/seed_0_k_2/alpha_0.1/so.pdf} \\
        % & $0.24$ & $0.18$ & $0.28$ & $0.36$ & $0.06$ \\
        %  & \includegraphics[width=0.12\textwidth, valign=c]{figures/chaosNLI/seed_0_k_2/alpha_0.05/tv.pdf} & 
        % \includegraphics[width=0.12\textwidth, valign=c]{figures/chaosNLI/seed_0_k_2/alpha_0.05/kl.pdf} & 
        % \includegraphics[width=0.12\textwidth, valign=c]{figures/chaosNLI/seed_0_k_2/alpha_0.05/ws.pdf} & 
        % \includegraphics[width=0.12\textwidth, valign=c]{figures/chaosNLI/seed_0_k_2/alpha_0.05/inner.pdf} & 
        % \includegraphics[width=0.12\textwidth, valign=c]{figures/chaosNLI/seed_0_k_2/alpha_0.05/so.pdf} \\
        % & $0.24$ & $0.18$ & $0.28$ & $0.36$ & $0.06$ \\
        \midrule
        {\color{blue}Premise:} \textit{The streets are busy and people contemplate their futures.}

        {\color{blue}Hypothesis:} \textit{People are screaming}  
        & \includegraphics[width=0.12\textwidth, valign=c]{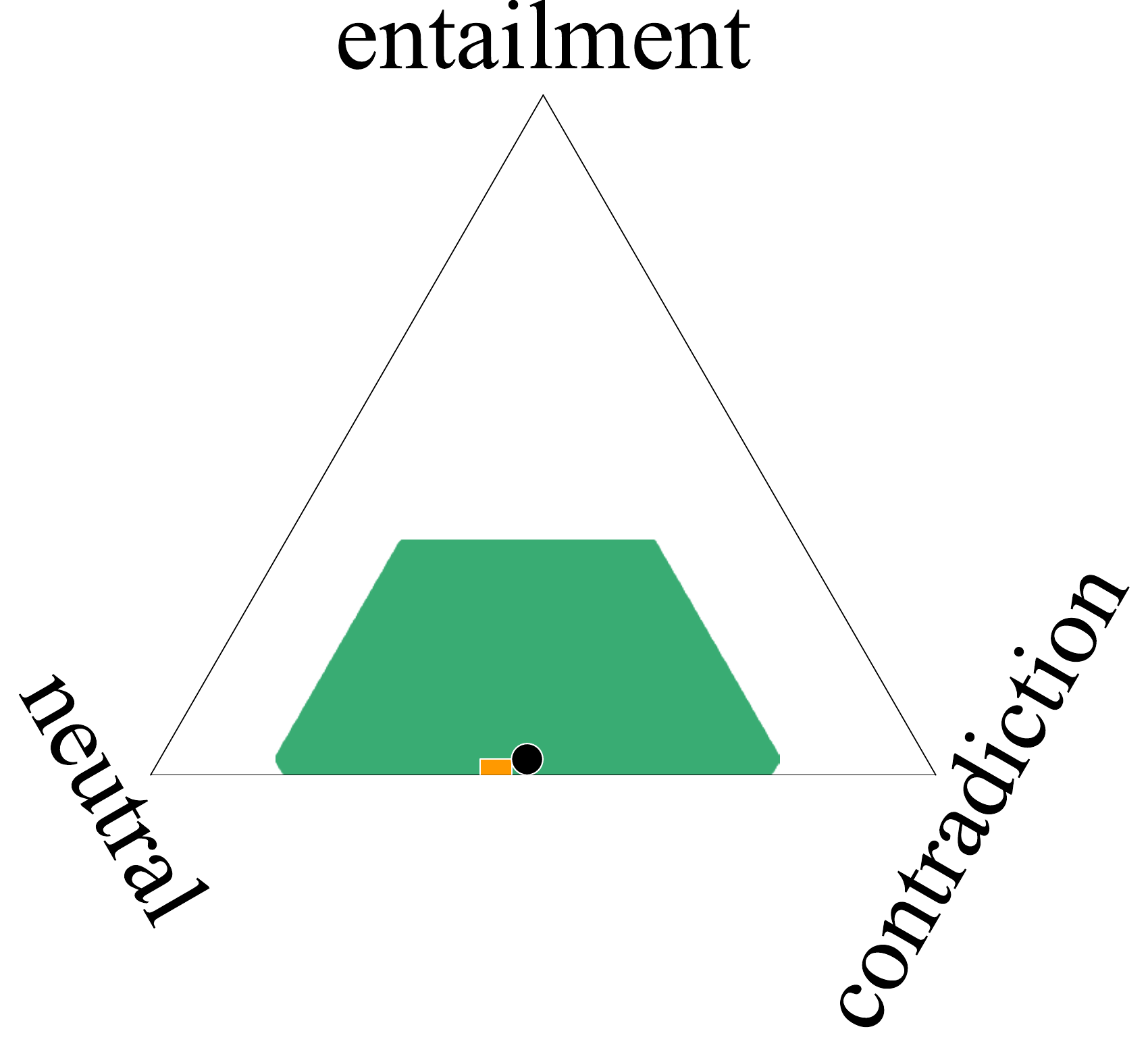} & 
        \includegraphics[width=0.12\textwidth, valign=c]{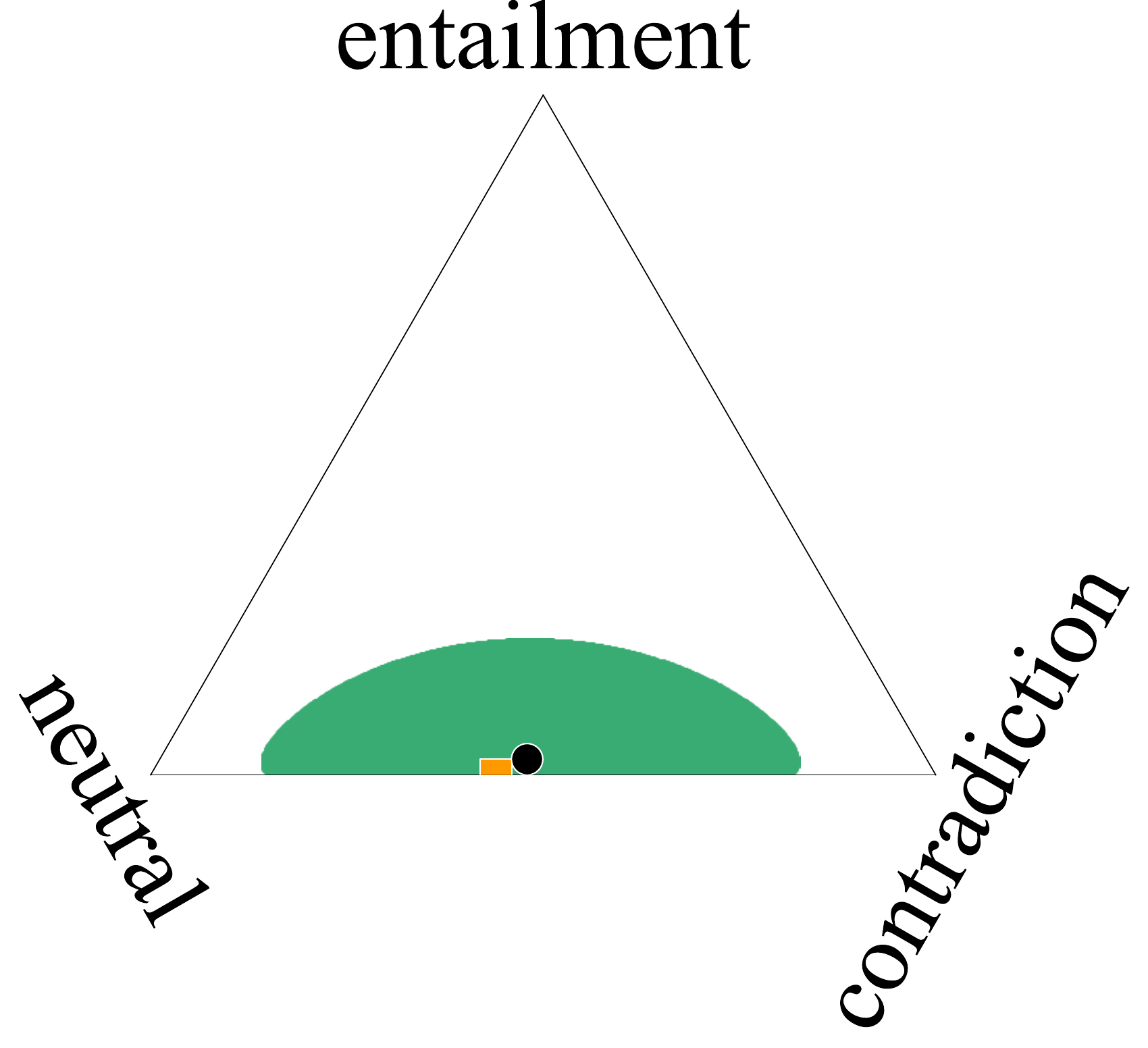} & 
        \includegraphics[width=0.12\textwidth, valign=c]{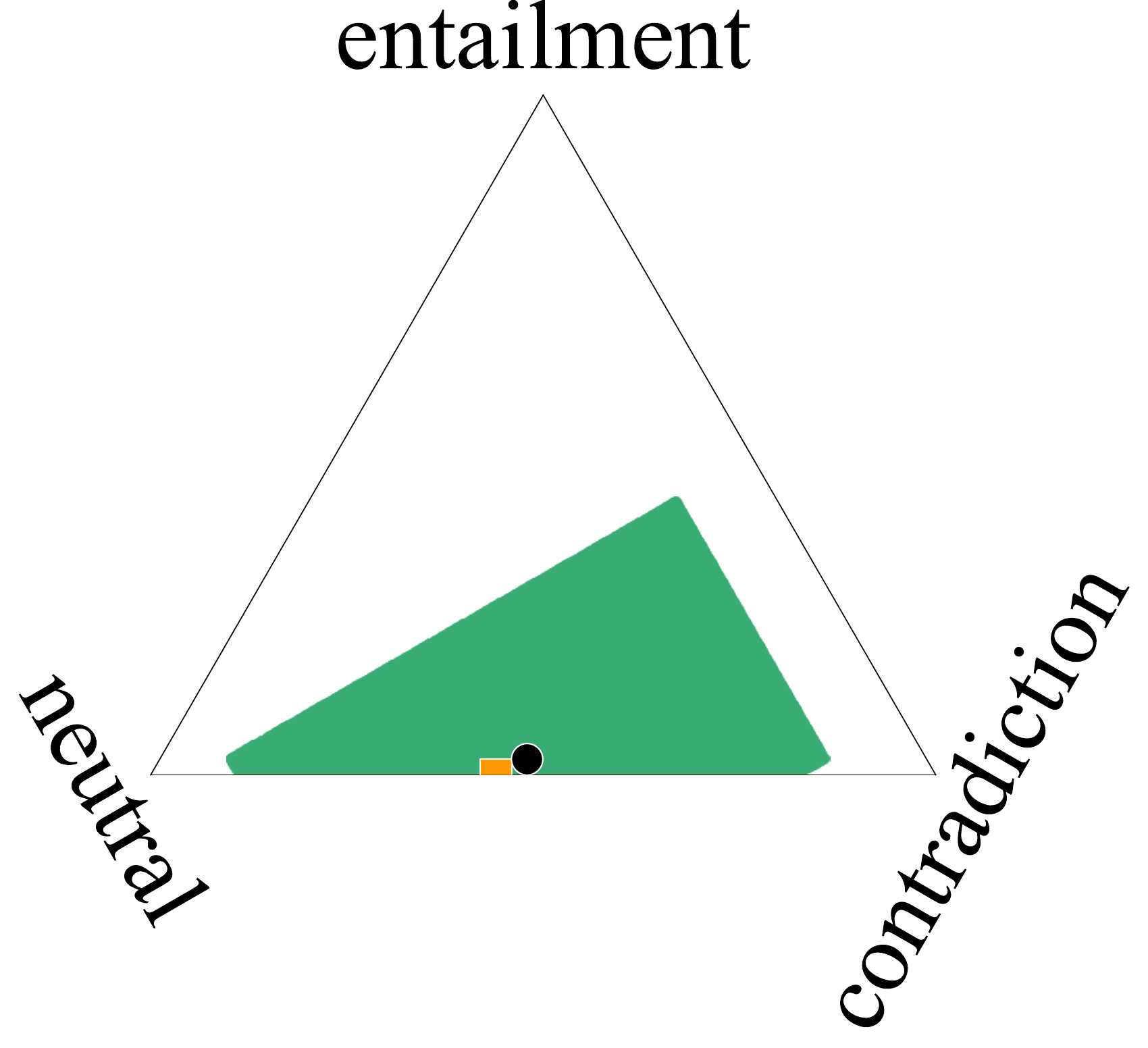} & 
        \includegraphics[width=0.12\textwidth, valign=c]{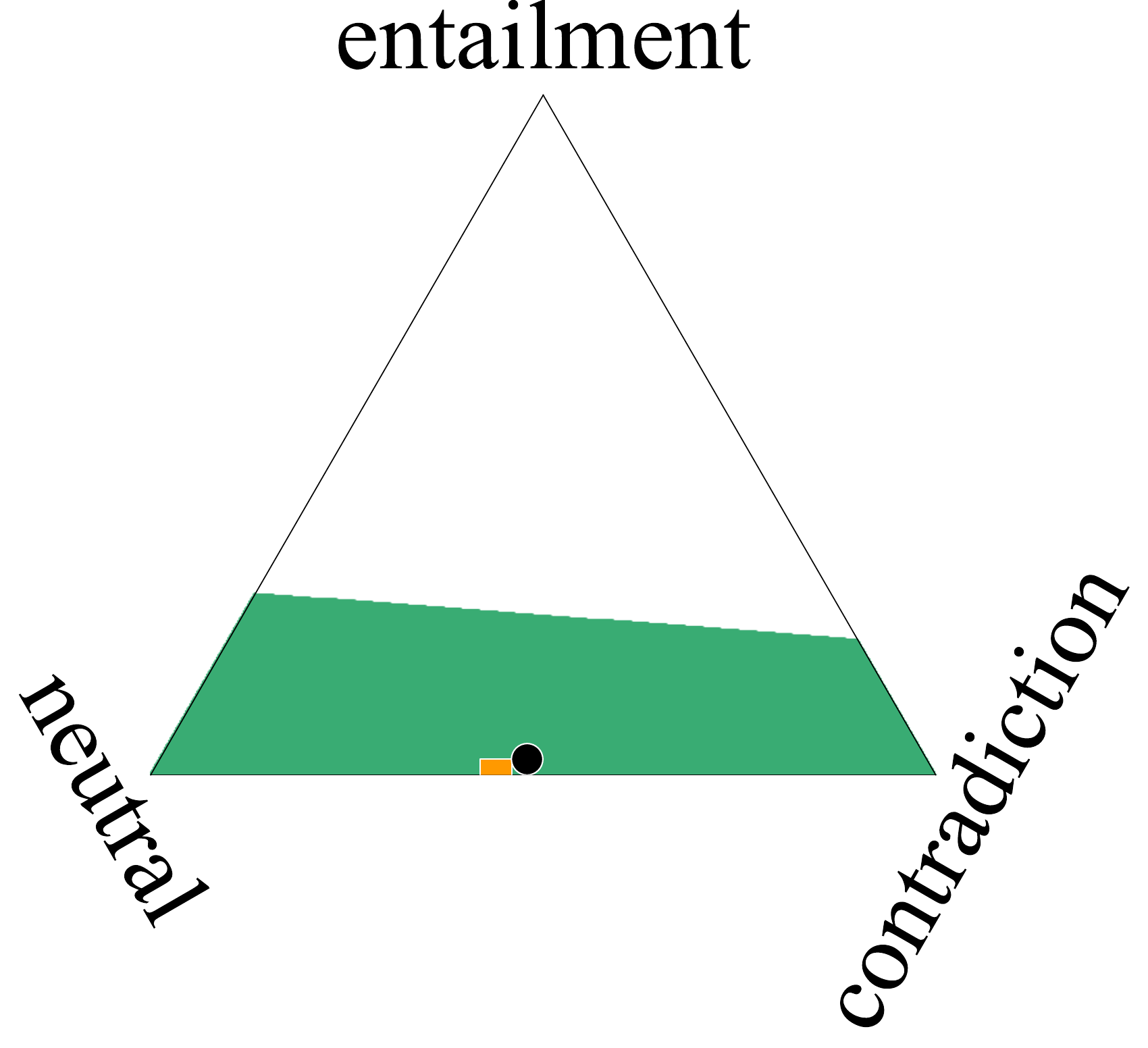} & 
        \includegraphics[width=0.12\textwidth, valign=c]{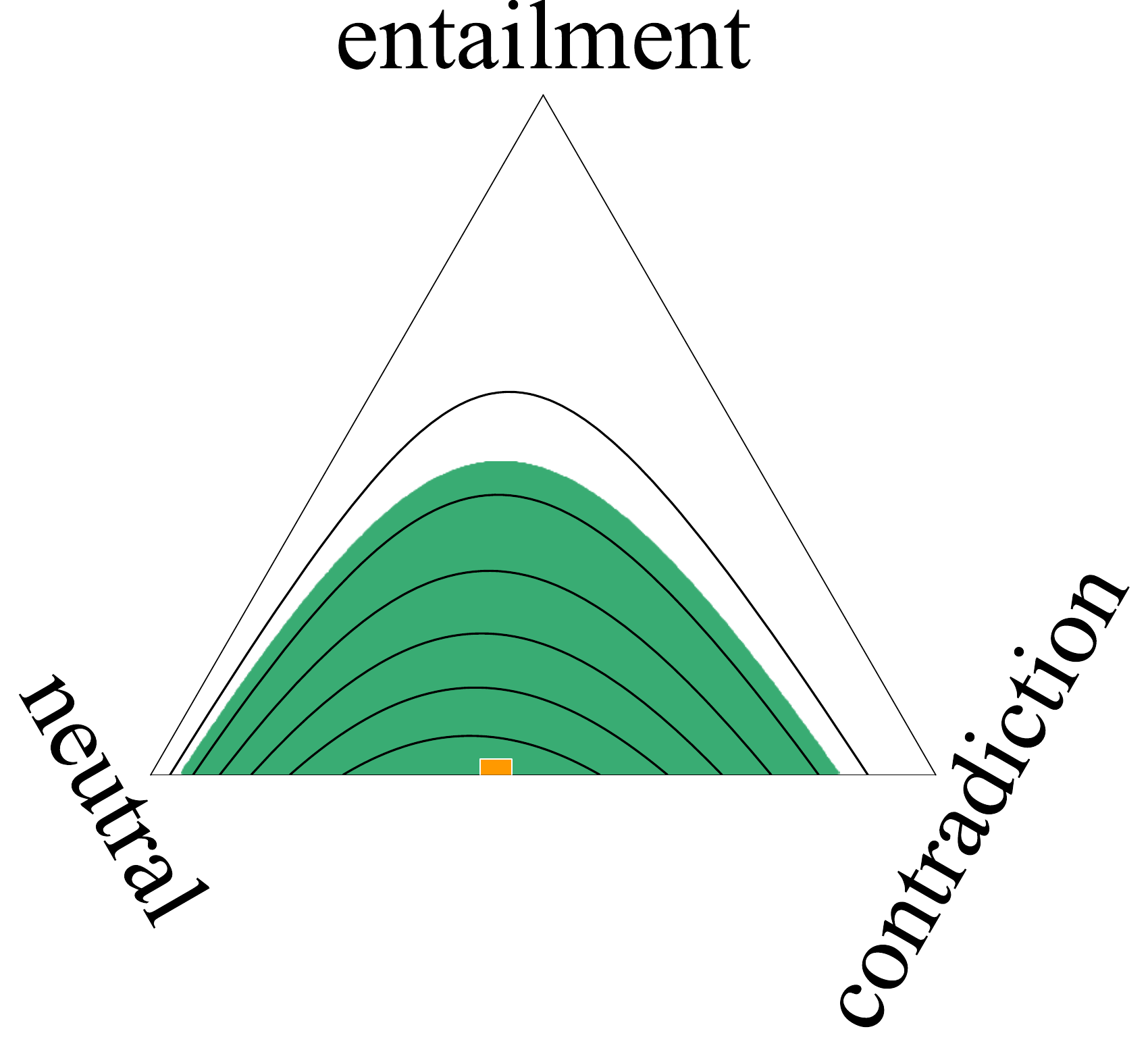} \\
        & $0.33$ & $0.20$ & $0.33$ & $0.41$ & $0.47$ \\
        \midrule
        {\color{blue}Premise:} \textit{A photographer snaps a midair action shot of a snowboarder. }
        {\color{blue}Hypothesis:} \textit{The midair shot snaps at a boarder of snow.}
& \includegraphics[width=0.12\textwidth, valign=c]{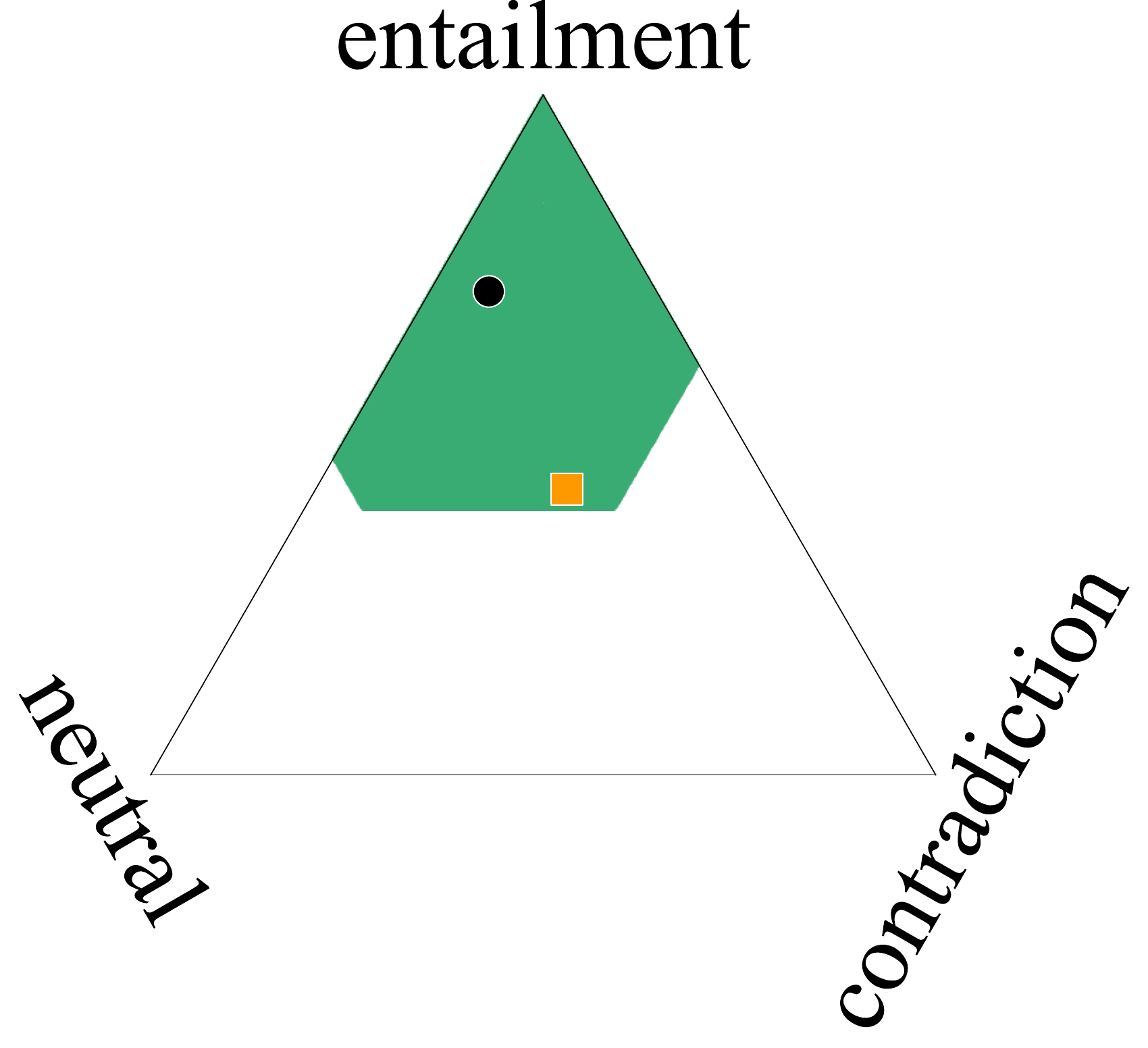} & 
        \includegraphics[width=0.12\textwidth, valign=c]{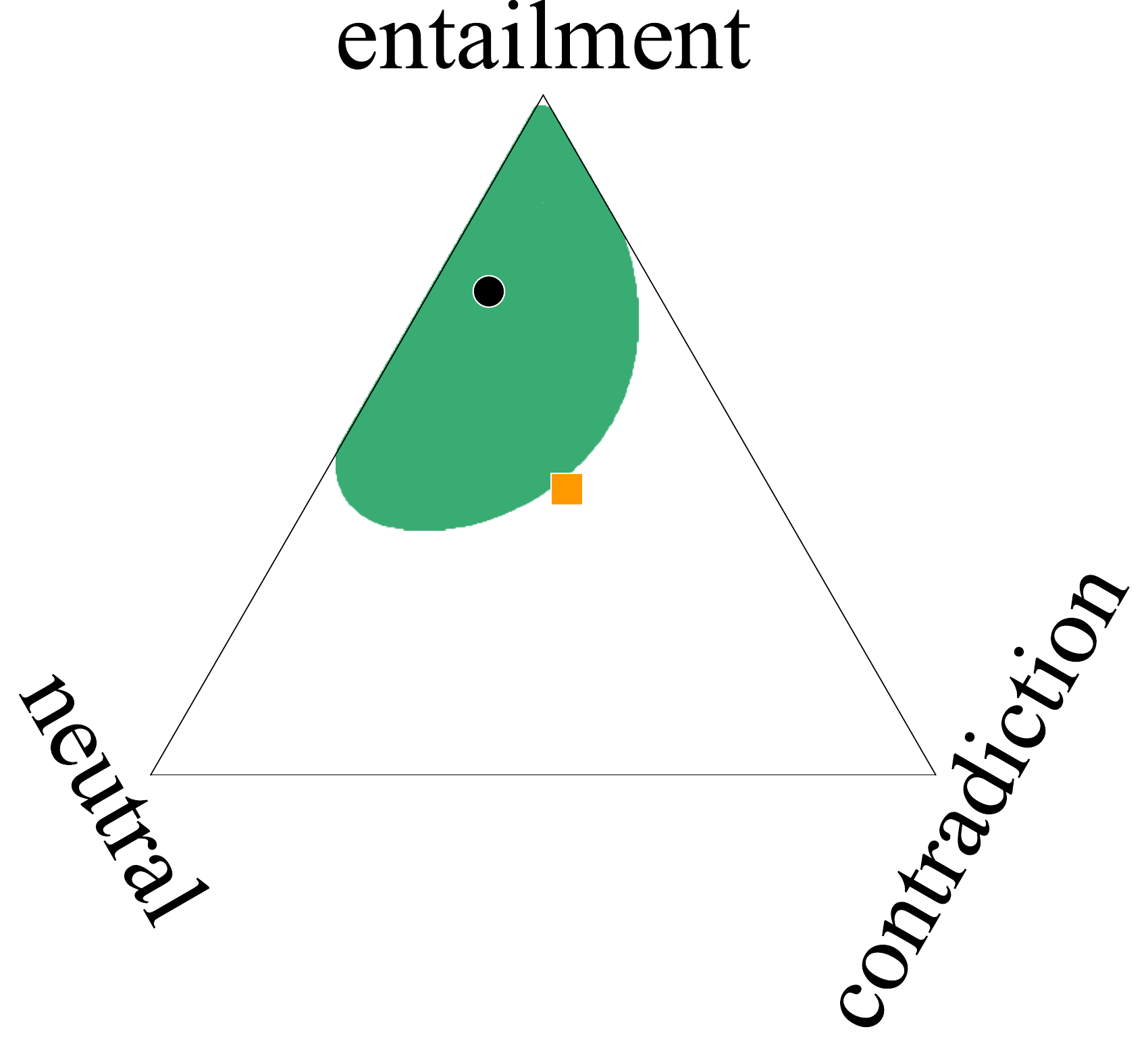} & 
        \includegraphics[width=0.12\textwidth, valign=c]{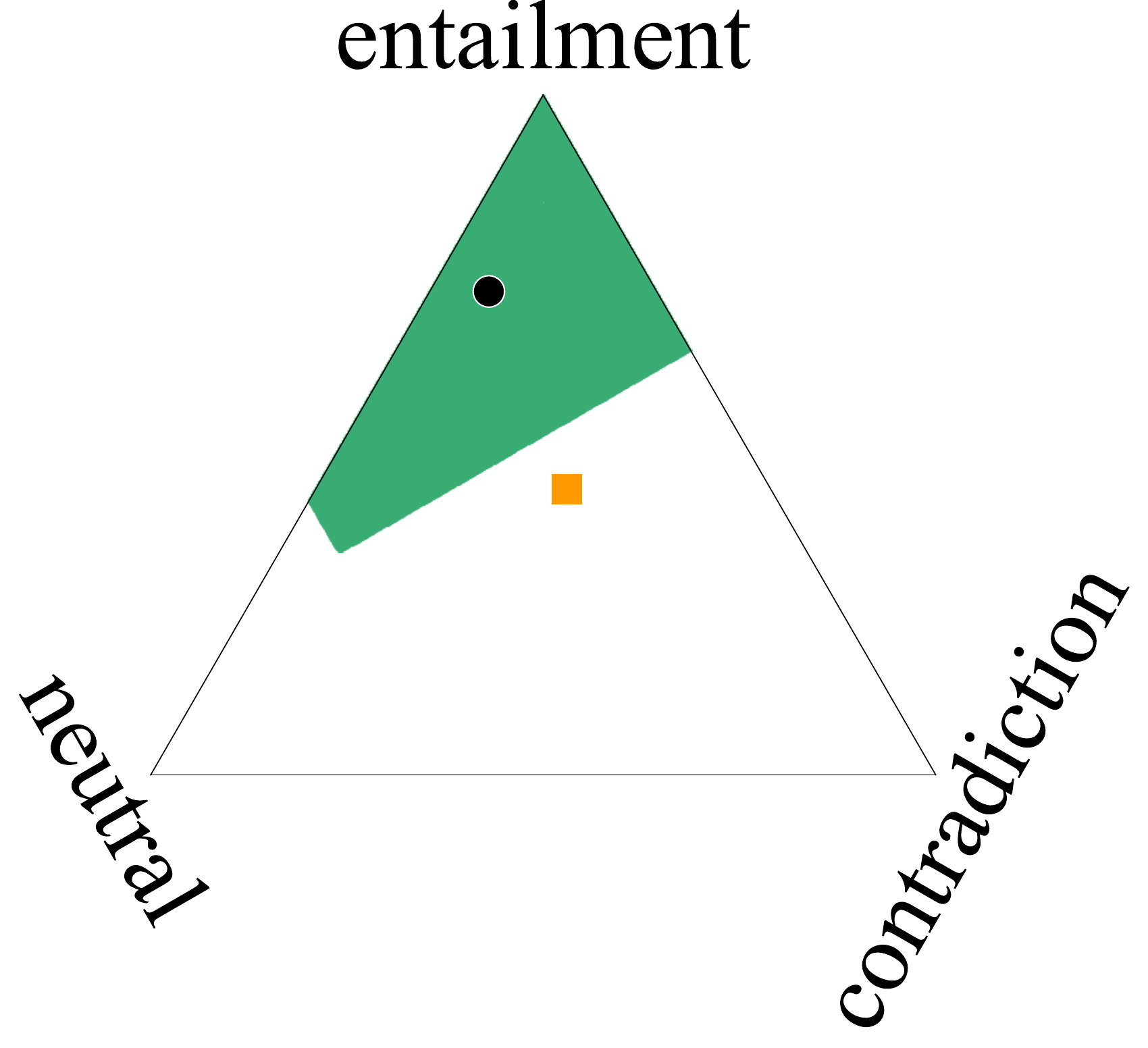} & 
        \includegraphics[width=0.12\textwidth, valign=c]{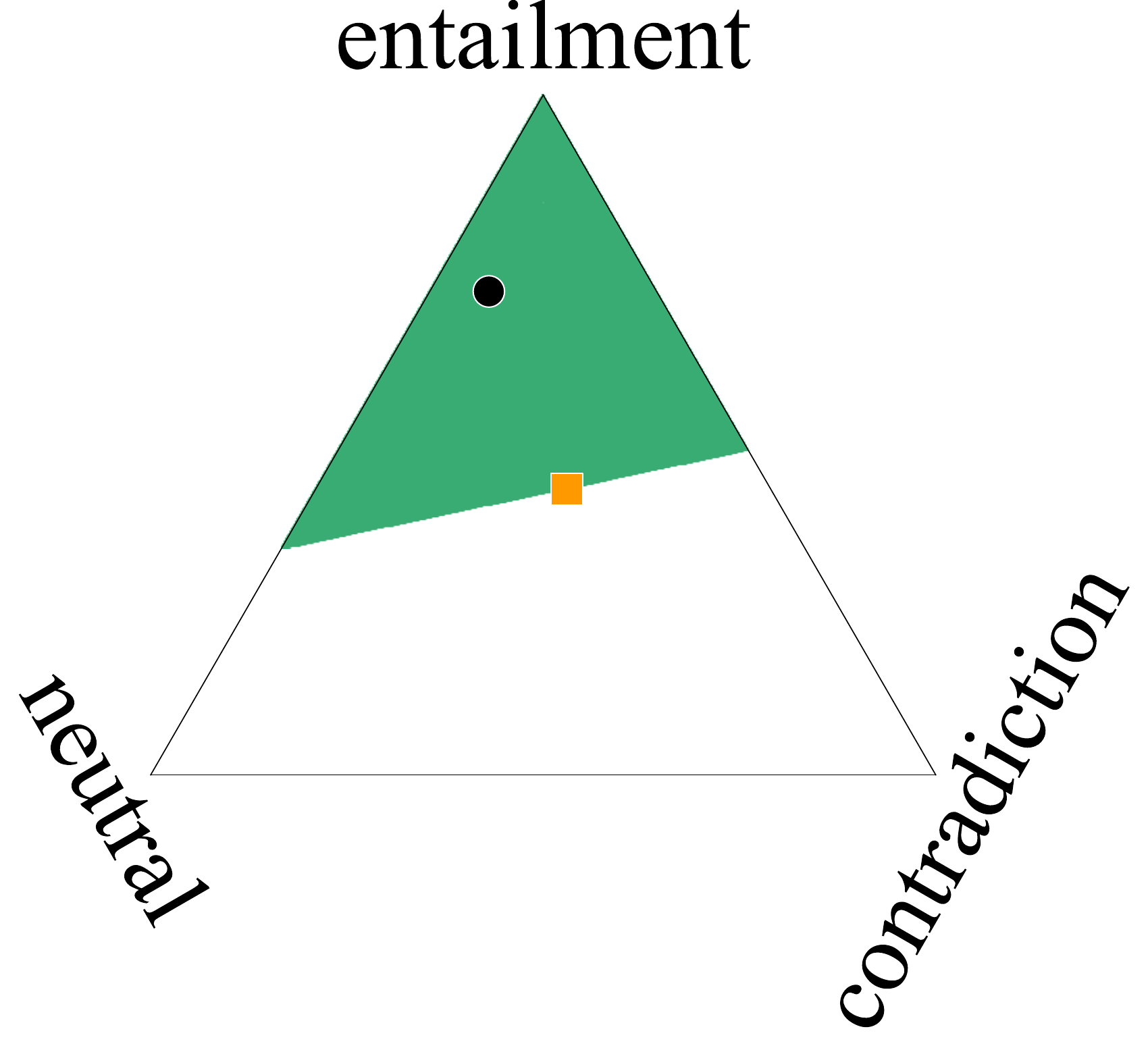} & 
        \includegraphics[width=0.12\textwidth, valign=c]{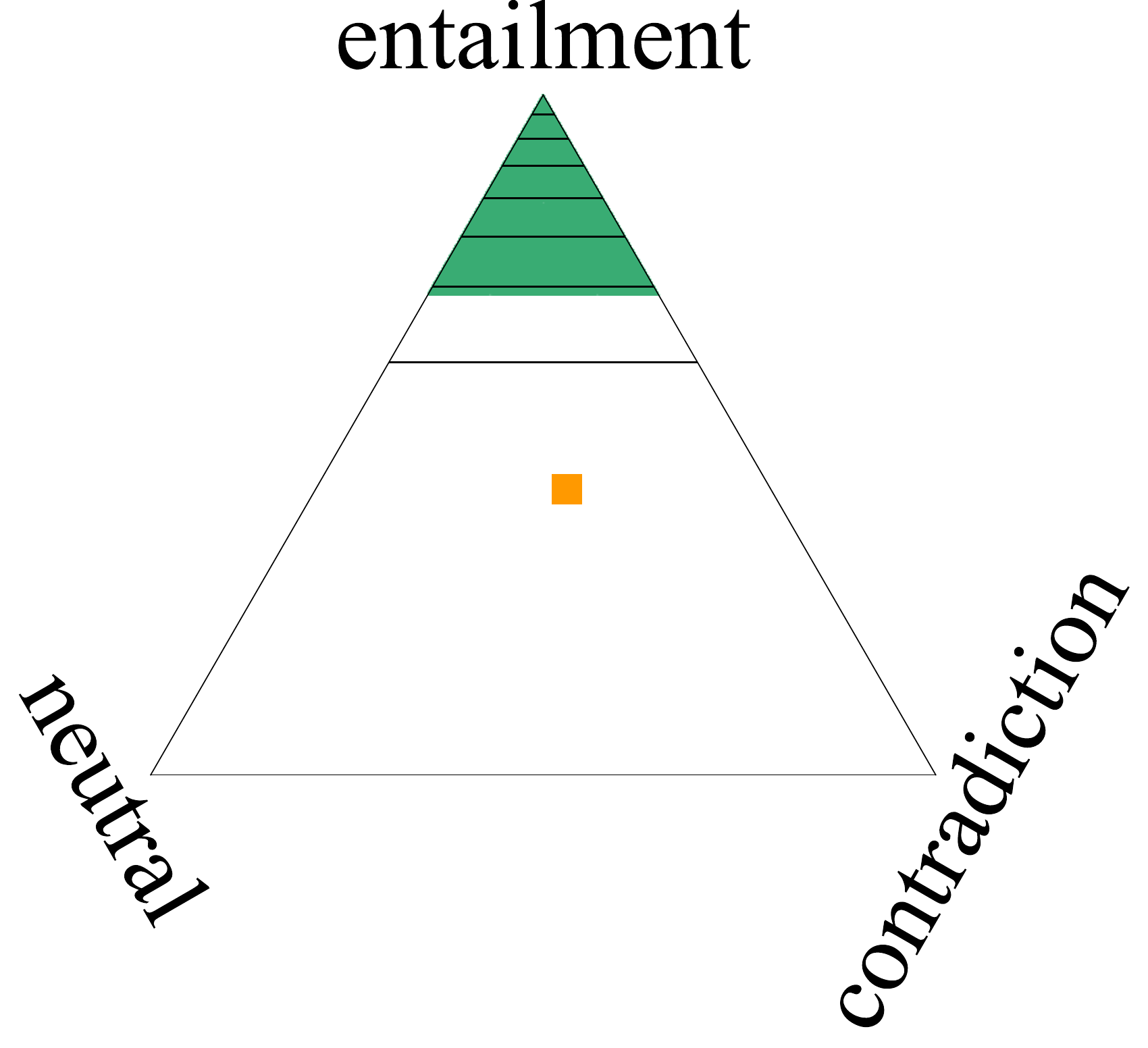} \\
        & $0.31$ & $0.27$ & $0.26$ & $0.34$ & $0.08$ \\
        \bottomrule
    \end{tabularx}
    \caption{Various credal sets obtained for three instances from ChaosNLI dataset \citep{nie2020what}. The ground truth distributions are denoted by {\color{gt_color} orange} squares. Black circles indicate model predictions in cases employing a first-order learner (first four columns). For the last column, utilizing a second-order learner, the predicted second-order distributions are represented through contour plots. The miscoverage rate is $\alpha=0.2$, and the efficiency of each credal set is written below it.} 
    \label{fig:chaos:3instances}
\end{figure*}

\begin{figure*}[t!]
\centering
\includegraphics[width=\textwidth]{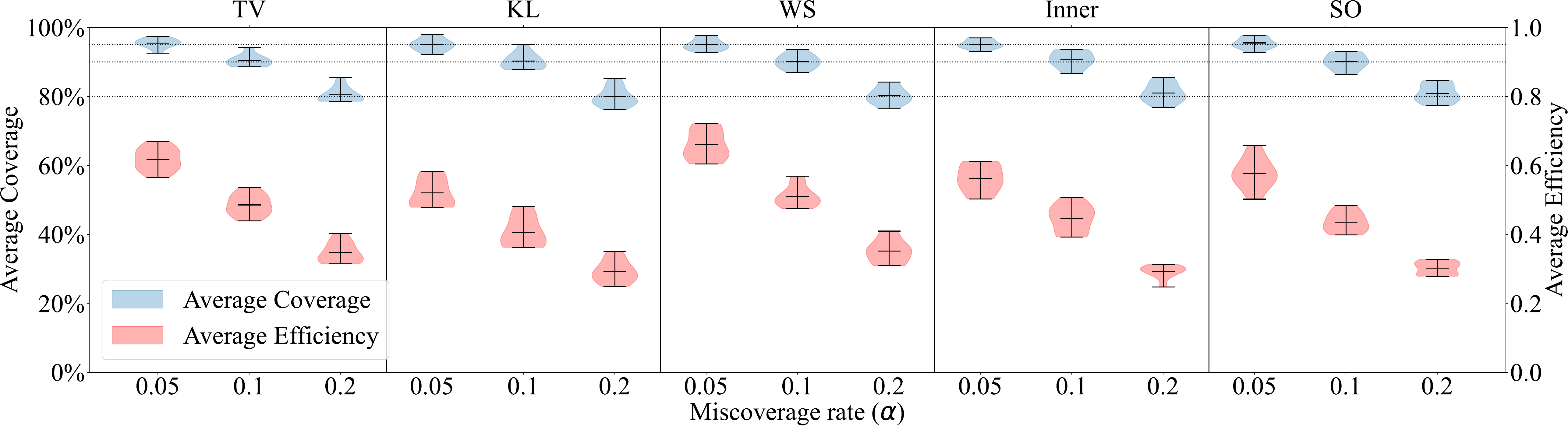}
\caption{Coverage and efficiency results of different nonconformity functions applied on the ChaosNLI dataset \citep{nie2020what}. The horizontal dashed lines indicate the nominal coverage levels.}
\label{fig:chaos:violin}
\end{figure*}

We combine the chaos-SNLI and chaos-MNLI subsets, resulting in a dataset of $3113$ datapoints. For model training, we leverage a language model from the Hugging Face \texttt{transformers} library \citep{wolf2019huggingface}, initially trained on SNLI and MultiNLI datasets for classification tasks\footnote{The model can be found at \url{https://huggingface.co/cross-encoder/nli-deberta-base}}. We utilize the last hidden layer output of this model to embed the premise-hypothesis pairs from our $3113$ instances, serving as inputs for our deep neural network.
To split the data, we randomly select $500$ instances for calibration, $500$ for testing, and the remaining for training. This process is repeated ten times with different random seeds. In Figure \ref{fig:chaos:3instances}, we compare the resulting credal sets of different nonconformity functions for three specific instances. Figure \ref{fig:chaos:violin} summarizes the overall performance of the proposed methods on this data under different miscoverage rates ($\alpha$). Notably, the mean of the average coverage over the test data across various random seeds aligns with or exceeds the nominal value, consistent with the conformal prediction guarantee.

% \begin{table}[t!]
%     \centering
% 		\caption{Coverage and efficiency results on ChaosNLI dataset with $\alpha = 0.25$.}
%   \resizebox{\columnwidth}{!}{
%     \begin{tabular}{ lc c c c}
% \toprule 
% Nonconformity & \textbf{TV} & \textbf{KL} & \textbf{WS} & \textbf{Inner}\\
% \midrule
% Avg. Coverage & $0.75\pm 0.017$ & $0.75\pm 0.011$ & $0.76\pm 0.006$ & $0.75\pm 0.013$\\
% Avg. Efficiency & $0.30\pm 0.004$ & $0.27\pm 0.004$ & $0.27\pm 0.005$ & $0.23\pm 0.005$\\
% \bottomrule
%     \end{tabular}
%     }
% 		\label{tab:res:real}
% \end{table}

\subsection{Synthetic Data}
The primary objective of conducting experiments with synthetic data is to illustrate the impact of noisy observations, particularly to showcase the behavior of the proposed credal sets when we only have access to an approximation of the ground truth distributions. Our experiment revolves around a $K$-class classification task with $K \in \{3, 4, 6, 8, 10\}$. For each $K$, we consider $10$-dimensional features $X \in \mathbb{R}^{10}$, where each $X_1,\ldots, X_{10}$ are independent standard normal random variables. Subsequently, we generate a random matrix $\beta \in \mathbb{R}^{10 \times K}$, with its elements drawn independently from the standard normal distribution.
To define the ground truth probability over the classes for object $X$, we use the following formulation:
\begin{align*}
    \lambda_k^{\vec{x}} := \mathbb{P}(Y=k|\vec{x}) = \frac{Z_j(\vec{x})}{\sum_{j}Z_j(\vec{x})},
\end{align*}
where $Z(\vec{x}) := \exp(\vec{x} ^\top \beta)$. Employing this data-generating process, we generate $N=1500$ samples to construct the dataset $\mathcal{D}^K = \{ (\vec{x}_i, \vec{\lambda}^{\vec{x}_i})\}_{i=1}^{N}$.
\begin{figure*}[t!]
\centering
\includegraphics[width=\textwidth]{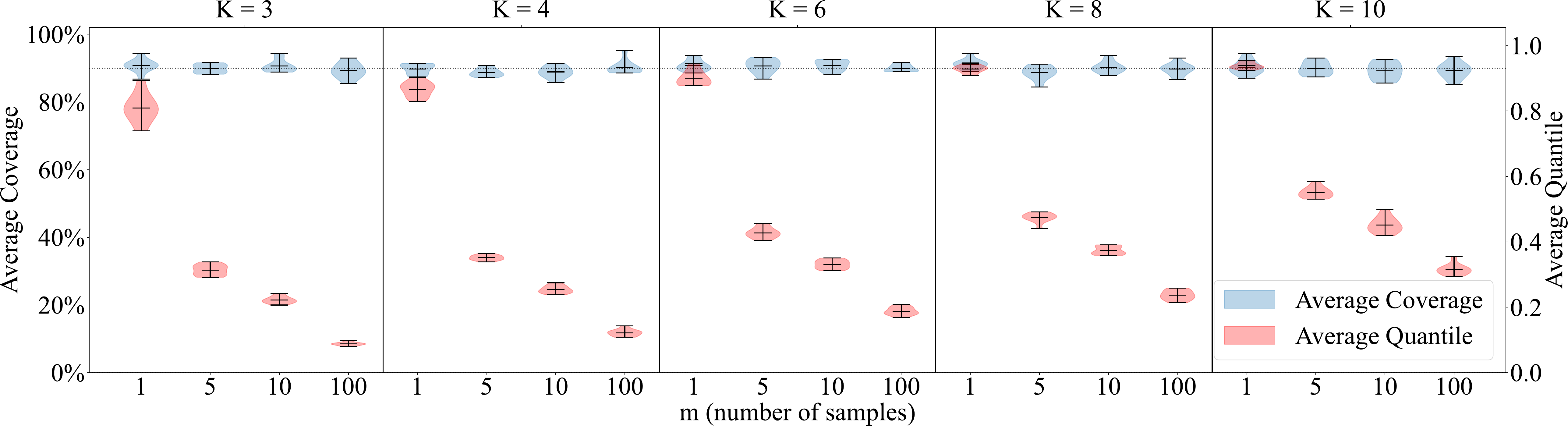}
\caption{Coverage and quantile results for synthetic data, where the ground truth distributions are approximated by observing $m$ samples from them. The horizontal dashed line indicates the nominal coverage levels $1-\alpha = 0.9$.}
\label{fig:synthetic:violin:tv}
\end{figure*}

To obtain noisy versions of $\mathcal{D}^K$, we employ a sampling approach. Specifically, we independently sample each distribution $\vec{\lambda}^{\vec{x}_i}$ $m$ times and utilize relative frequencies to create its noisy counterpart $\tilde{\vec{\lambda}}_m^{\vec{x}_i}$. We represent the resulting dataset as $\mathcal{D}^K_m = \{ (\vec{x}_i, \tilde{\vec{\lambda}}_m^{\vec{x}_i})\}_{i=1}^{N}$. We repeat this process four times with $m \in \{ 1, 5, 10, 100\}$. 

Given each dataset $\mathcal{D}^K_m$, we randomly partition data points equally into training, calibration, and test sets and perform the proposed methodologies accordingly. Again, we repeat this process ten times with different random seeds for each dataset $\mathcal{D}^K_m$. Due to the computational complexity in calculating efficiency for cases with $K>3$, we utilize the quantile of the calibration nonconformity scores as an efficiency metric. In Figure \ref{fig:synthetic:violin:tv}, we represent the overall result for \textbf{TV} under different $K$ and $m$ values. It can be observed that the coverage is fulfilled across almost all scenarios, including $m=1$ with degenerate distributions. This observed behavior is somewhat intuitive. The model endeavors to learn the underlying probabilistic relationship between $X$ and $Y$, even given the noisy data \footnote{Of course, this holds under some reasonable assumptions on noise.}. Consequently, during calibration with noisy instances, the nonconformity scores of noise-free instances are mostly upper-bounded by the scores of their noisy counterparts, resulting in more conservative sets that effectively cover the ground-truth distributions. It can also be seen that the quantile of the nonconformity scores shrinks as $m$ increases. Results for other nonconformity functions, along with some visualizations for the specific case of $K=3$, can be found in \ref{sec:appendix:synthetic}.

\section{Limitations}
%{\color{red} Let's discuss briefly our limitations, Including the fact that creating the sets is challenging analytically. Maybe something about data as well.}

The methods we proposed are promising but still subject to certain limitations. One challenge, for example, lies in the representation of credal sets as subsets of the probability simplex. For the nonconformity functions we used in this work, there are no closed-form equations for the resulting credal sets. Instead, the sets are only represented implicitly (through the nonconformity threshold). Numerical approximation is feasible but essentially limited to scenarios with a small number of classes. 

The issue of representation is also connected to the computation of uncertainty measures, i.e., numerical measures quantifying the total, aleatoric, and epistemic uncertainty associated with a credal set \citep{klir1999uncertainty}. Computation of these measures involves the computation of specific characteristics of the set, such as its distance from the center of the simplex or its volume \citep{sale2023volume}.

%Furthermore, the method's generalization capability may be hindered when confronted with datasets containing (any arbitrary) noisy versions of true distributions. 

Our generalization to the case of noisy training data, i.e., labelings $\tilde{\vec{\lambda}}$ that only approximate the ground-truth probabilities $\vec{\lambda}$, provides guarantees under the bounded noise assumption. While this assumption is plausible, and in a sense always achievable with a sufficiently large $\epsilon$, its practical use requires a meaningful choice of $\epsilon$ and $\delta$, which leads to inference that is both valid and efficient. This will be difficult in cases of limited knowledge about the labeling noise. If labels are constructed from relative frequencies (like in the case of multiple annotators), classical statistical methods might be applicable. In general, however, an appropriate choice of $\epsilon$ and $\delta$ for practical problems is still an open problem.

\section{Conclusion and Future Work}
Conformal credal set prediction connects machine learning with imprecise probability theory and offers a novel data-driven approach to constructing predictions that effectively capture both aleatoric and epistemic uncertainty. Thereby, it provides the basis of a new approach to reliable, uncertainty-aware machine learning. Leveraging the inherent validity of the conformal prediction framework, our conformalized credal sets are assured to cover the ground truth distributions with high probability. We have explored different nonconformity functions within this novel setting and evaluated their performance through numerical experiments.

A natural next step is to explore alternative approaches for defining nonconformity functions, with the goal of devising formulations amenable to closed-form solutions for credal sets. Besides, the nonconformity has a strong influence on the efficiency and hence the uncertainty of credal set predictions. Obviously, there is a preference for nonconformity functions leading to higher efficiency and lower uncertainty.

Another interesting direction is to extend the learning of credal set predictors to standard (zero-order) training data. As already mentioned, learning a second-order predictor from data of that kind turns out to be difficult in the case of second-order probability distributions. Broadly speaking, this is due to the inherent ambiguity of the missing first-order information, which cannot be resolved due to certain averaging effects  \citep{bengs22pitfals,Bengs2023}. For the case of credal predictors, the situation is still less clear.

% \clearpage
\section*{Acknowledgment}
Alireza Javanmardi was supported by the Deutsche Forschungsgemeinschaft (DFG, German Research Foundation): Project number 451737409. 
\bibliographystyle{plainnat}
\bibliography{references}

%%%%%%%%%%%%%%%%%%%%%%%%%%%%%%%%%%%%%%%%%%%%%%%%%%%%%%%%%%%%%%%%%%%%%%%%%%%%%%%
%%%%%%%%%%%%%%%%%%%%%%%%%%%%%%%%%%%%%%%%%%%%%%%%%%%%%%%%%%%%%%%%%%%%%%%%%%%%%%%
% APPENDIX
%%%%%%%%%%%%%%%%%%%%%%%%%%%%%%%%%%%%%%%%%%%%%%%%%%%%%%%%%%%%%%%%%%%%%%%%%%%%%%%
%%%%%%%%%%%%%%%%%%%%%%%%%%%%%%%%%%%%%%%%%%%%%%%%%%%%%%%%%%%%%%%%%%%%%%%%%%%%%%%
% \newpage
\appendix
\renewcommand\thesection{Appendix \Alph{section}.}
\onecolumn
\section{Proof of Theorem \ref{theorem:noisy}} \label{sec:appendix:proof}
\begin{proof}
Let $A$ denote the event $f( \vec{x}_\text{new}, {\vec{\lambda}}^{\vec{x}_\text{new}} ) < q + \epsilon$ and $\tilde{A}$ the event $f( \vec{x}_\text{new}, {\tilde{\vec{\lambda}}}^{\vec{x}_\text{new}} ) < q$. We have 
\begin{align*}
P(A) & \geq P(A \wedge \tilde{A}) \\
& = P(\tilde{A}) \cdot P(A \, | \, \tilde{A}) \\
& = P(\tilde{A}) \cdot ( 1 - P( \neg A \, | \, \tilde{A}) )
\end{align*}
Since $\neg A$ means that $f( \vec{x}_\text{new}, {\vec{\lambda}}^{\vec{x}_\text{new}} ) \geq q + \epsilon$, the conditional event $\neg A \, | \, B$ implies a violation of the closeness condition in (\ref{eq:close}), wherefore the probability $P(\neg A \, | \, B)$ is upper-bounded by $\delta$ according to (\ref{eq:close}). Therefore, noting that $P(\tilde{A}) \geq 1- \tilde{\alpha}$ is the standard guarantee by CP,  
\begin{align*}
P(A) & \geq P(\tilde{A}) \cdot ( 1 - P( \neg A \, | \, \tilde{A}) ) \\
& \geq (1 - \tilde{\alpha}) \cdot (1- \delta) \\
& = 1 - \alpha \, .
\end{align*}
\end{proof}
\section{More Results for the Synthetic Data}\label{sec:appendix:synthetic}
\begin{figure}
    \begin{subfigure}{\textwidth}
    \centering
        \includegraphics[width=0.5\textwidth]{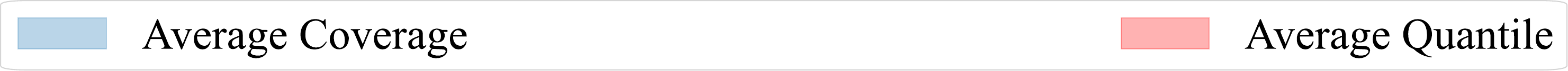}
        \vspace{3mm}
    \end{subfigure}
 
    \begin{subfigure}{\textwidth}
    \centering
        \includegraphics[width=\textwidth]{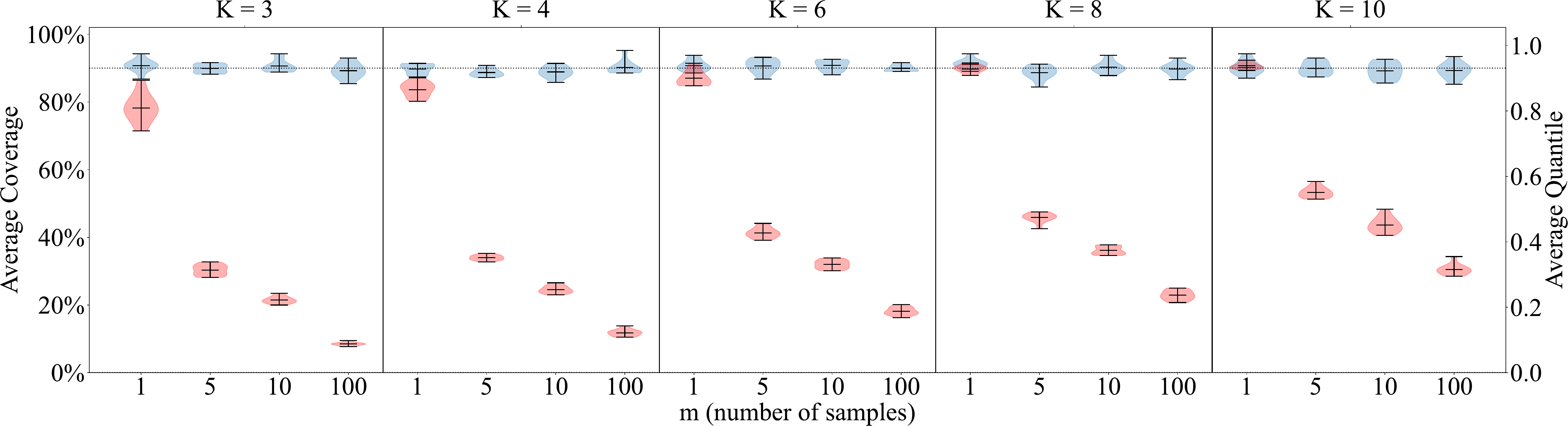}
        \caption{\textbf{TV}}
    \end{subfigure}

    \begin{subfigure}{\textwidth}
    \centering
        \includegraphics[width=\textwidth]{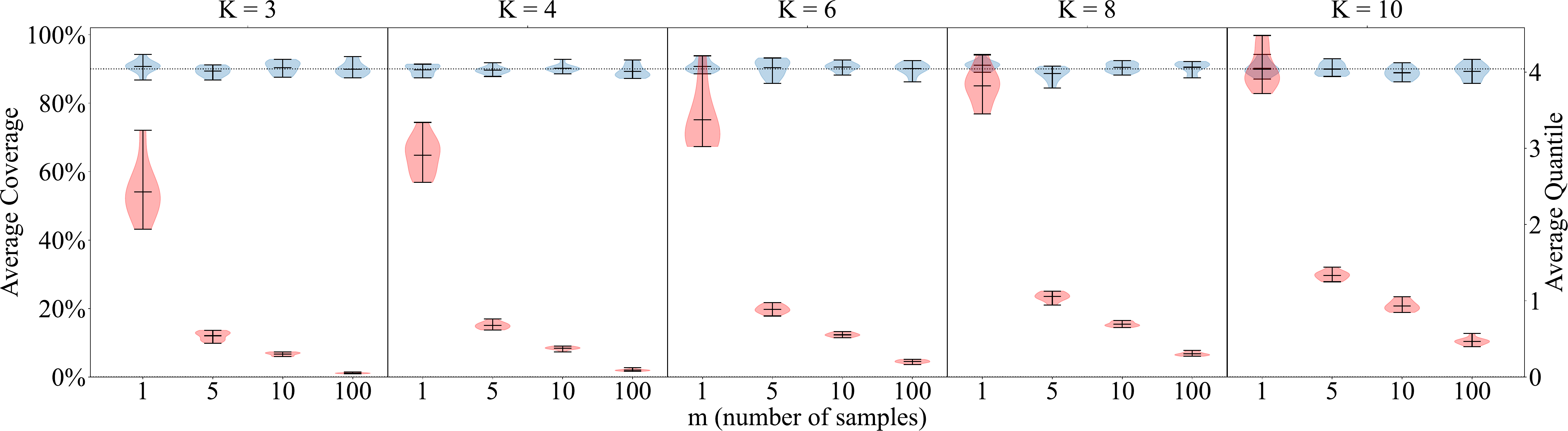}
        \caption{\textbf{KL}}
    \end{subfigure}
    \begin{subfigure}{\textwidth}
    \centering
        \includegraphics[width=\textwidth]{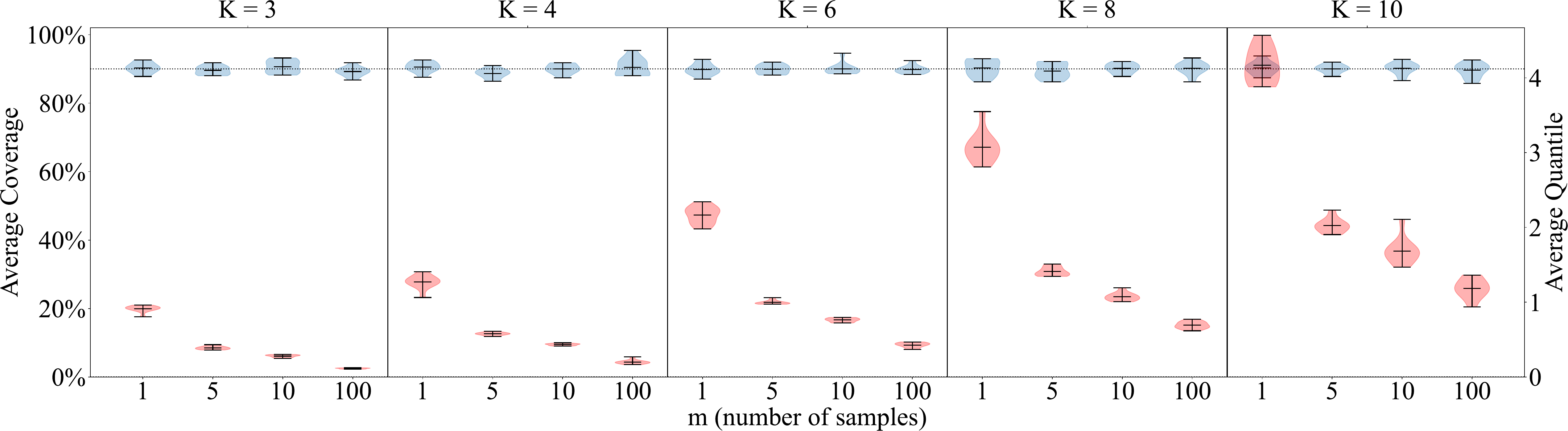}
        \caption{\textbf{WS}}
    \end{subfigure}
    \begin{subfigure}{\textwidth}
    \centering
        \includegraphics[width=\textwidth]{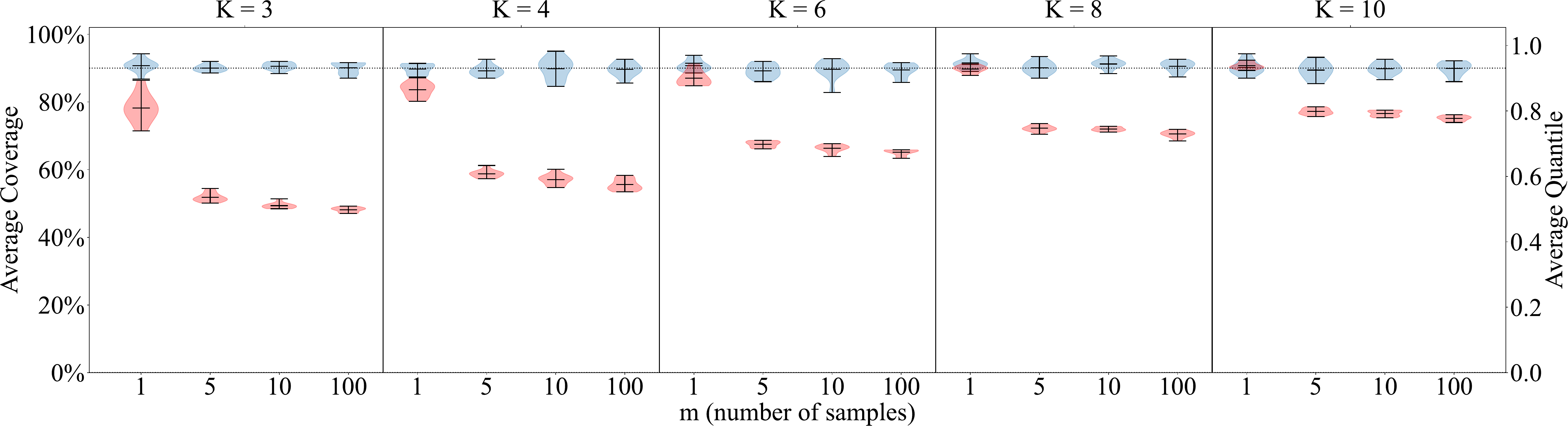}
        \caption{\textbf{Inner}}
    \end{subfigure}
    \caption{Coverage and quantile results for synthetic data, where the ground truth distributions are approximated by observing $m$ samples from them. The horizontal dashed lines indicate the nominal coverage levels $1-\alpha = 0.9$.}
    \label{fig:synthetic:violin:all}
\end{figure}

\begin{figure*}
    \centering
    \begin{tabularx}{\textwidth}{m{0.1\textwidth}YYYYY}
       \toprule
        $m$ & TV & KL & WS & Inner & SO\\
        \midrule
        1&\includegraphics[width=0.12\textwidth, valign=c]{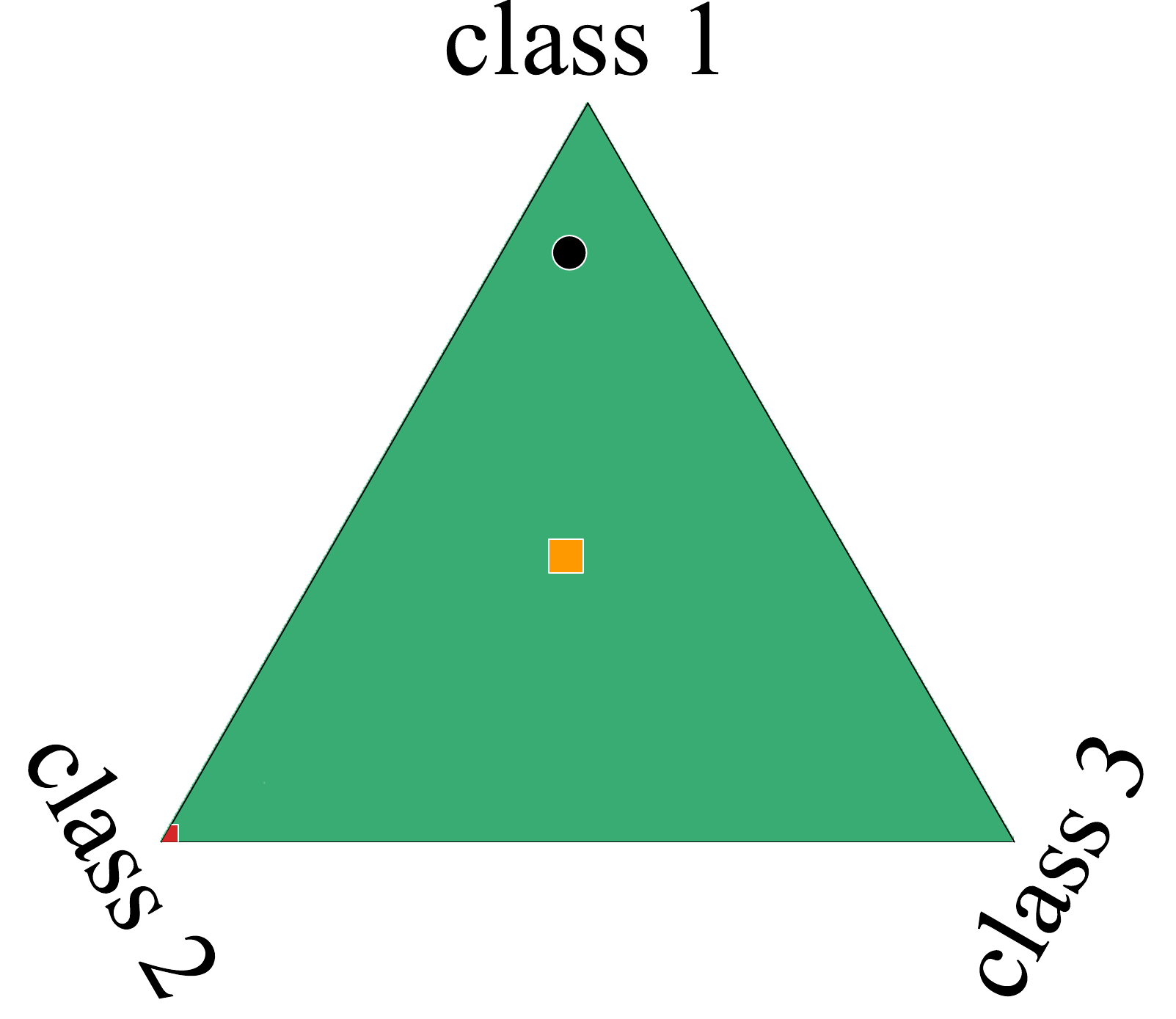} &
        \includegraphics[width=0.12\textwidth, valign=c]{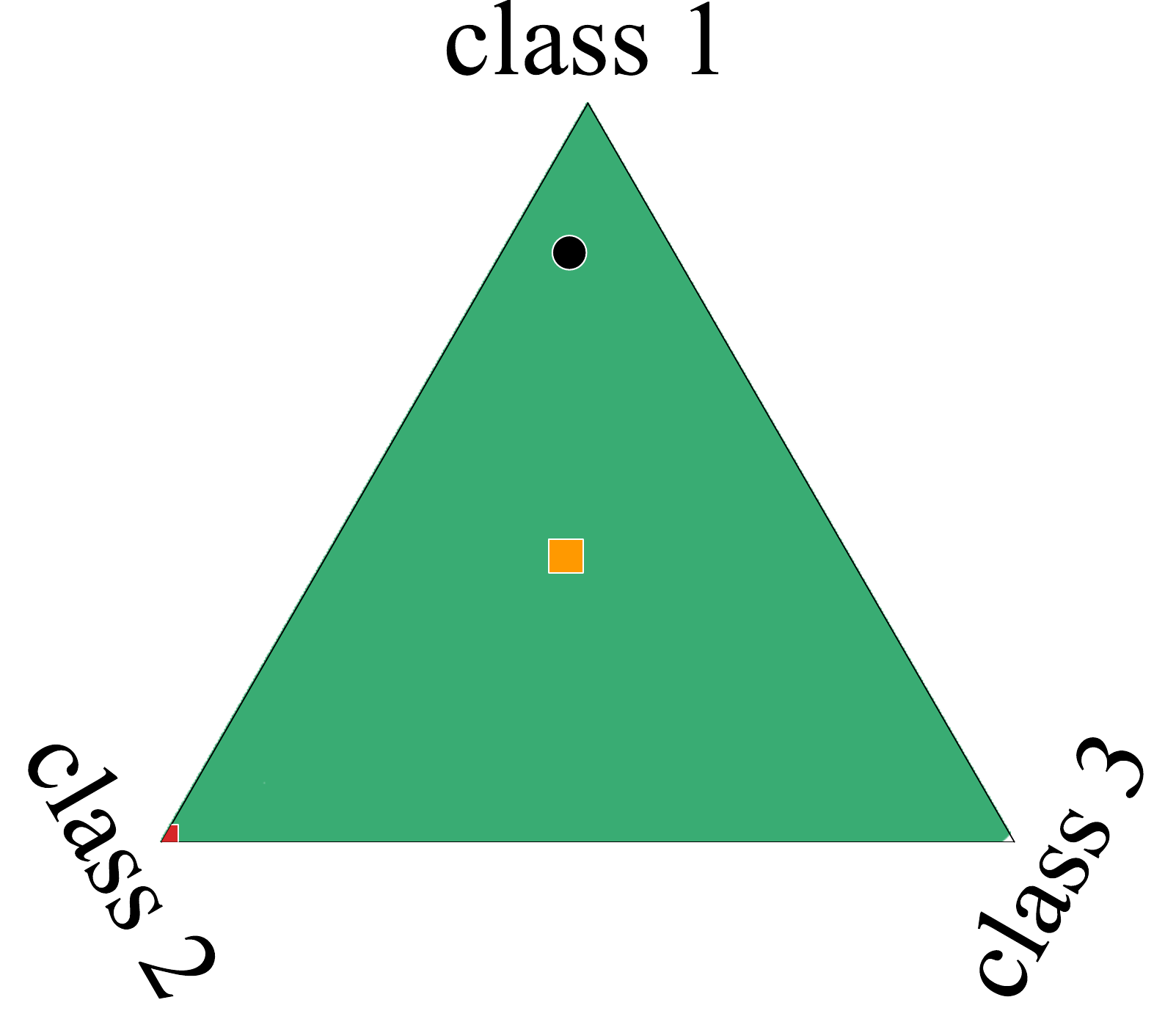} &
        \includegraphics[width=0.12\textwidth, valign=c]{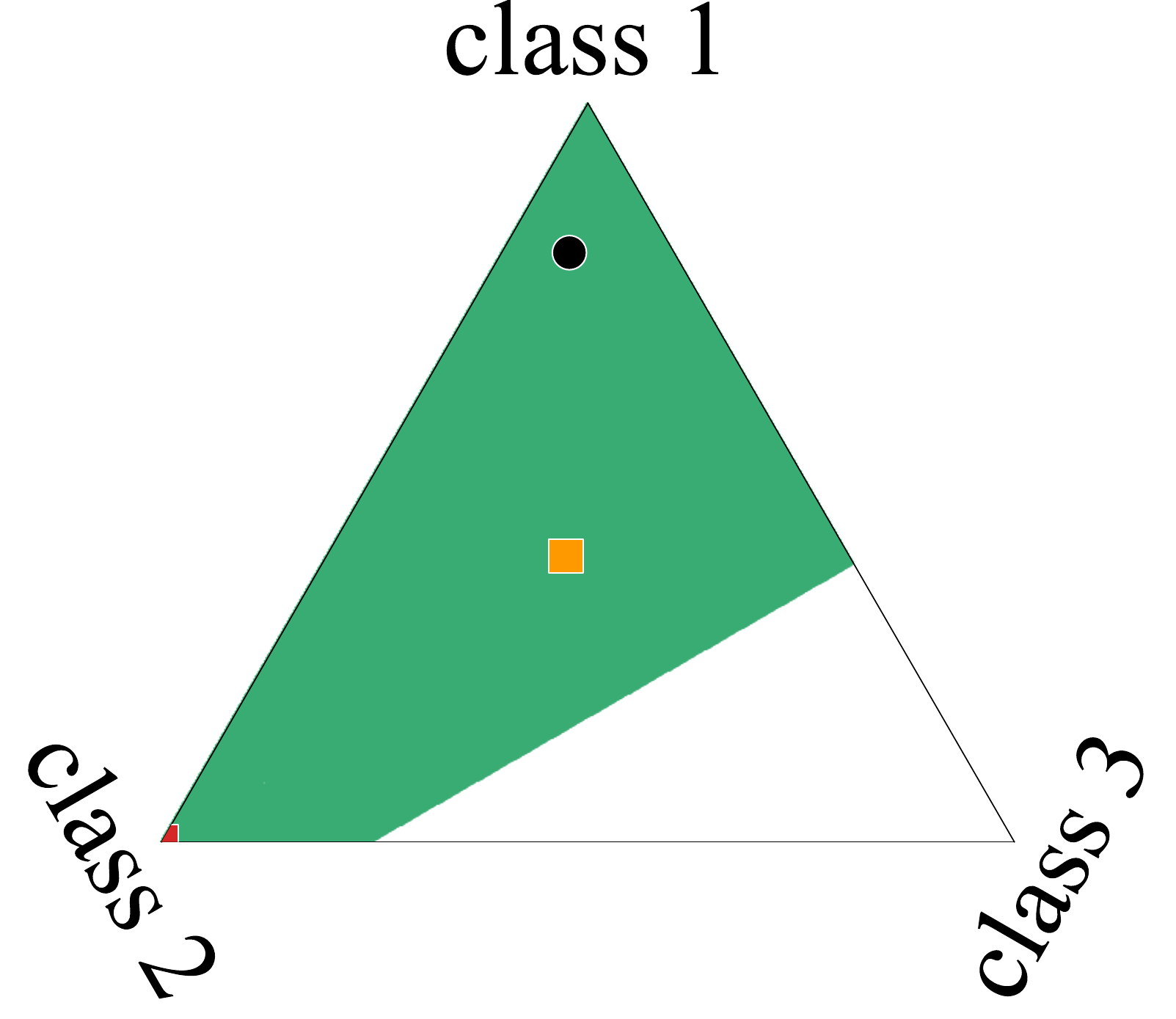} &
        \includegraphics[width=0.12\textwidth, valign=c]{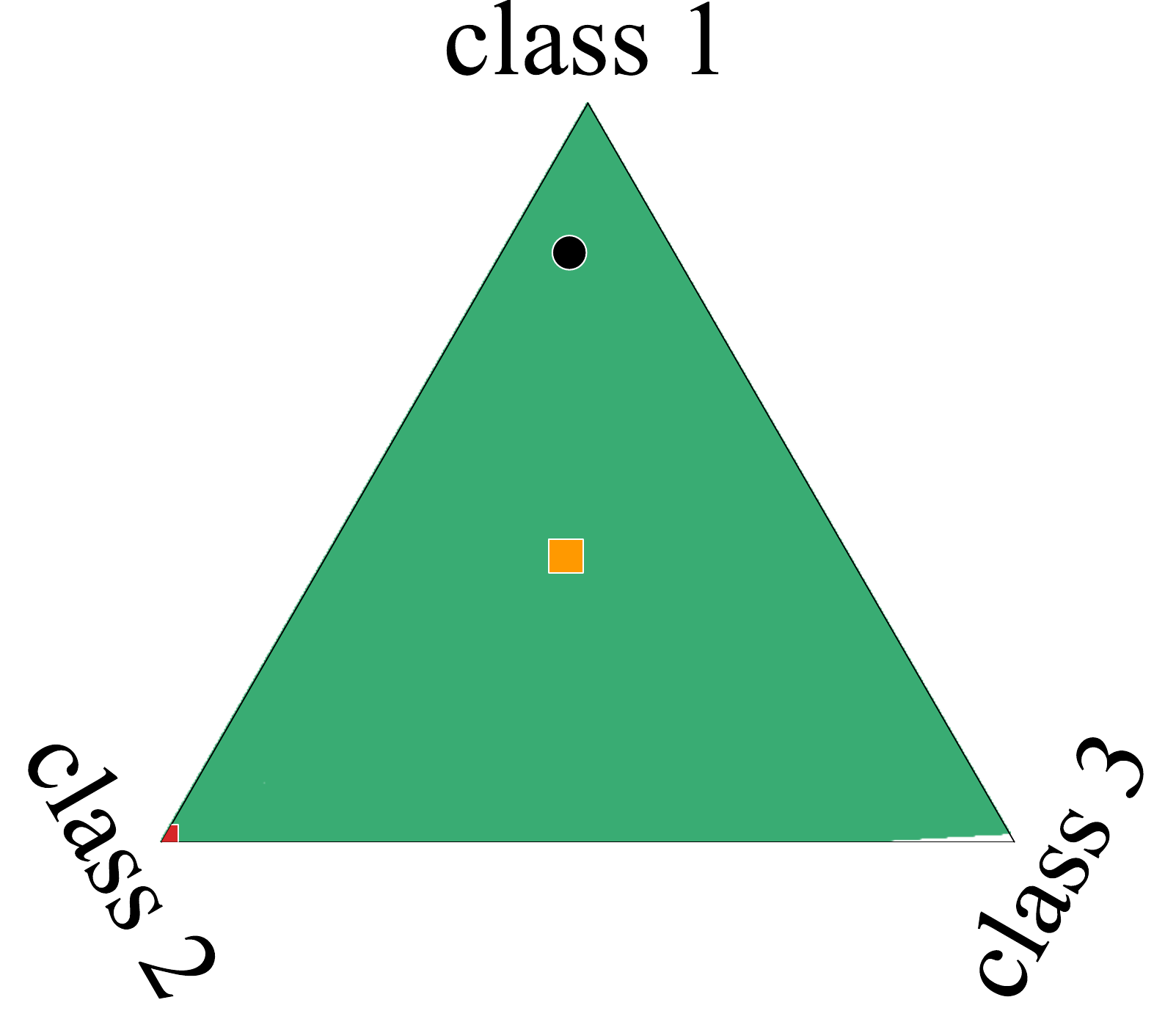}& 
        \includegraphics[width=0.12\textwidth, valign=c]{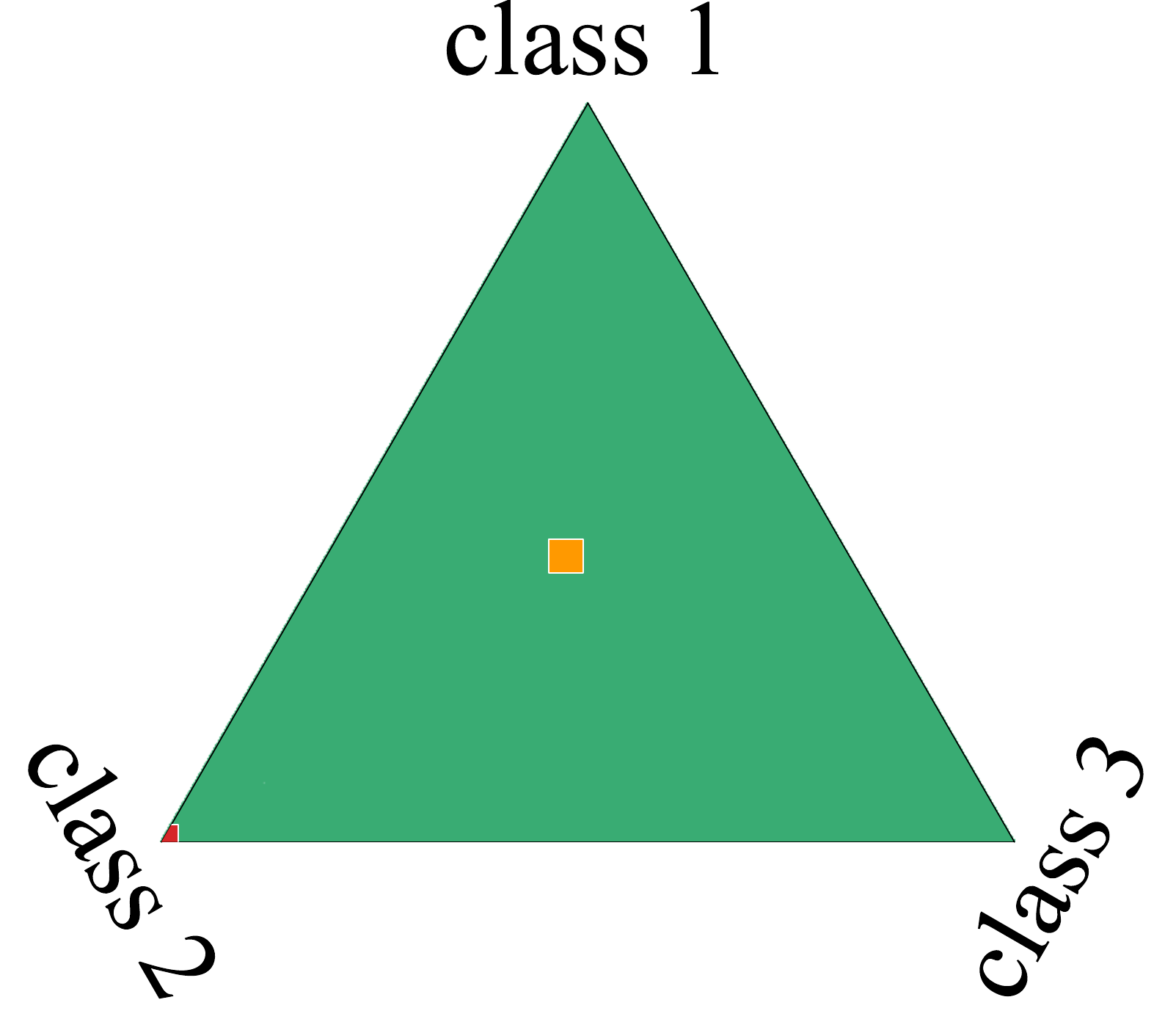} 
        \\ 
        & $1.00$ & $1.00$ & $0.70$ & $1.00$ & $1.00$   \\
        \midrule
        5&\includegraphics[width=0.12\textwidth, valign=c]{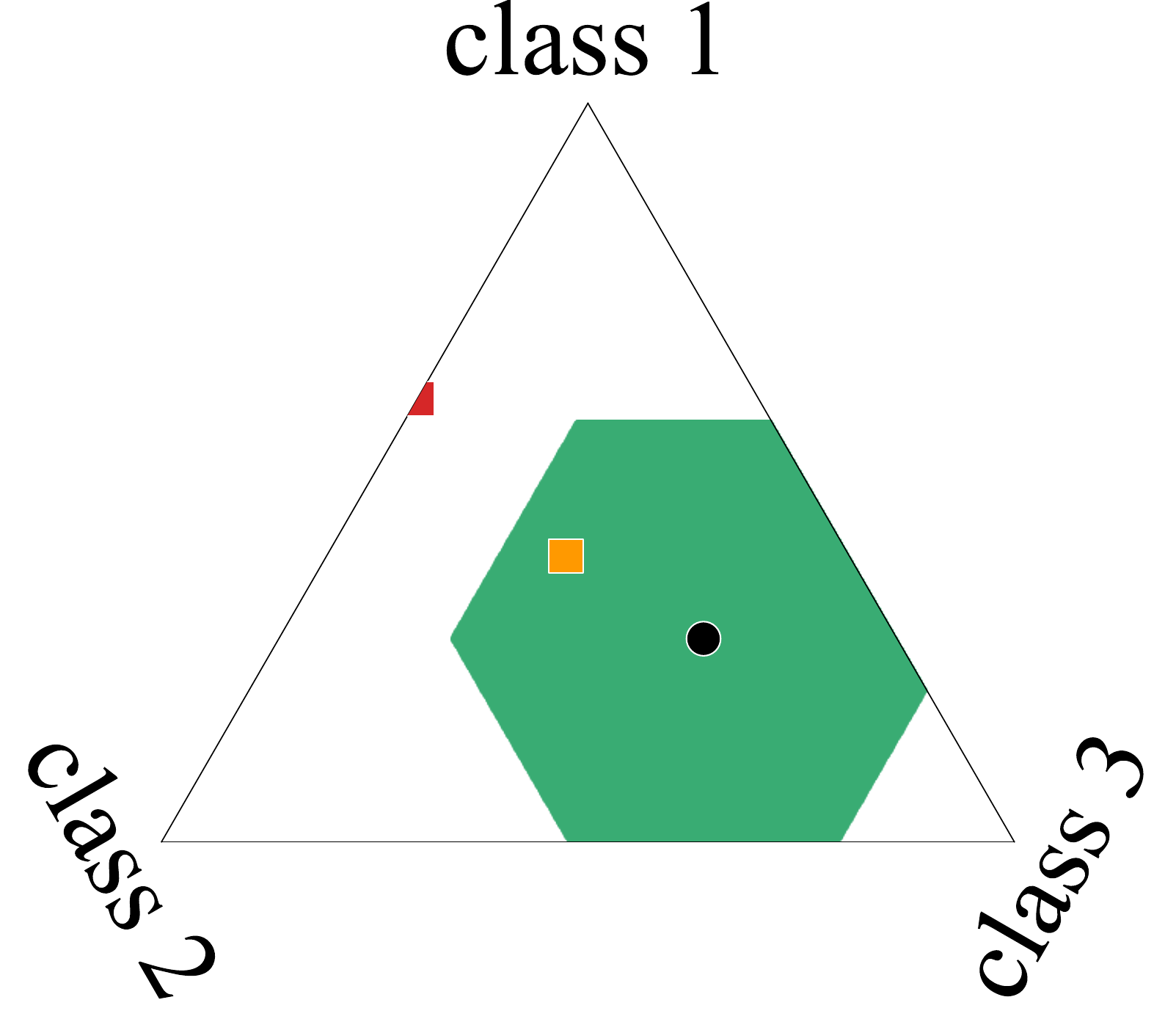} &
        \includegraphics[width=0.12\textwidth, valign=c]{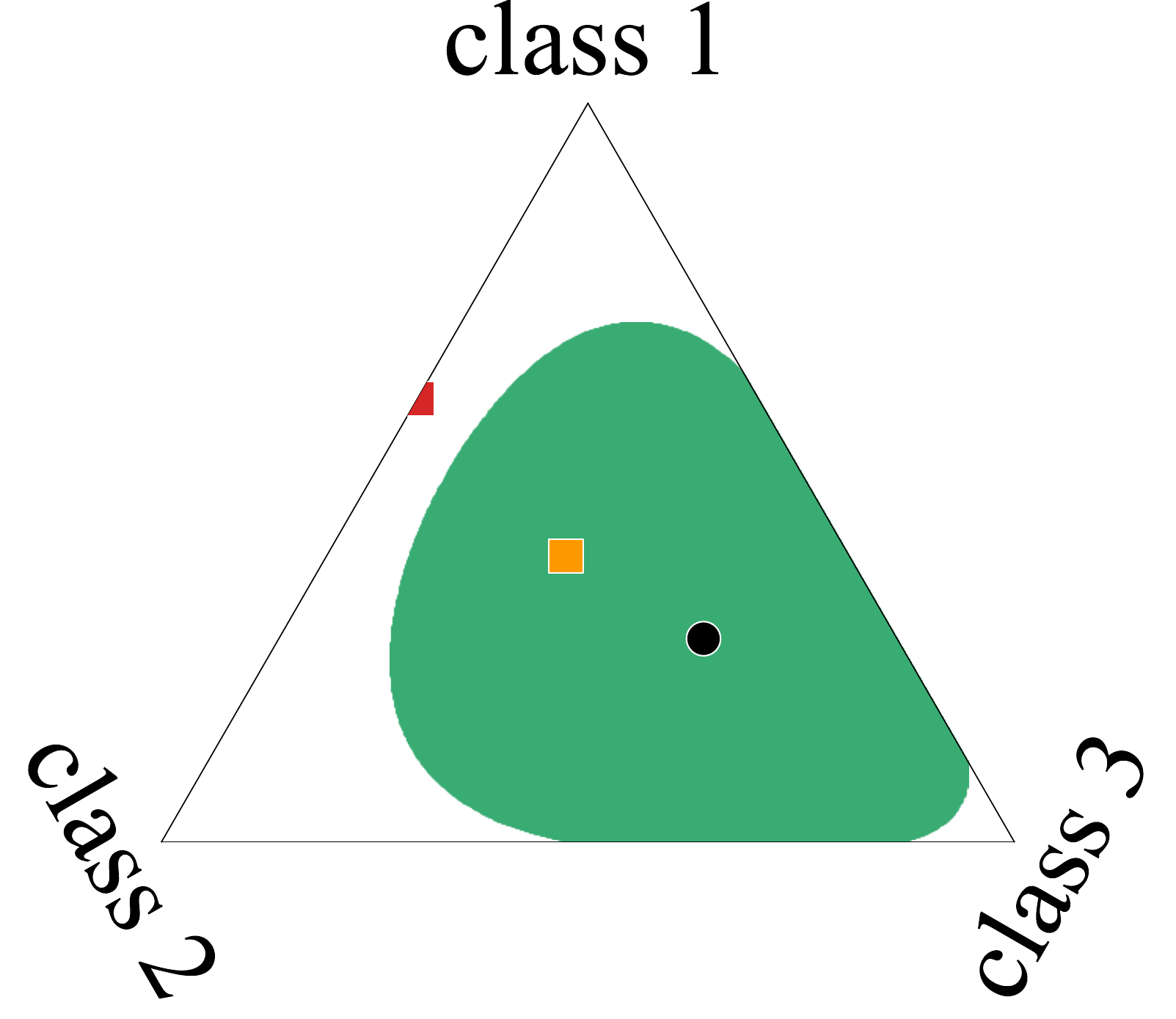} &
        \includegraphics[width=0.12\textwidth, valign=c]{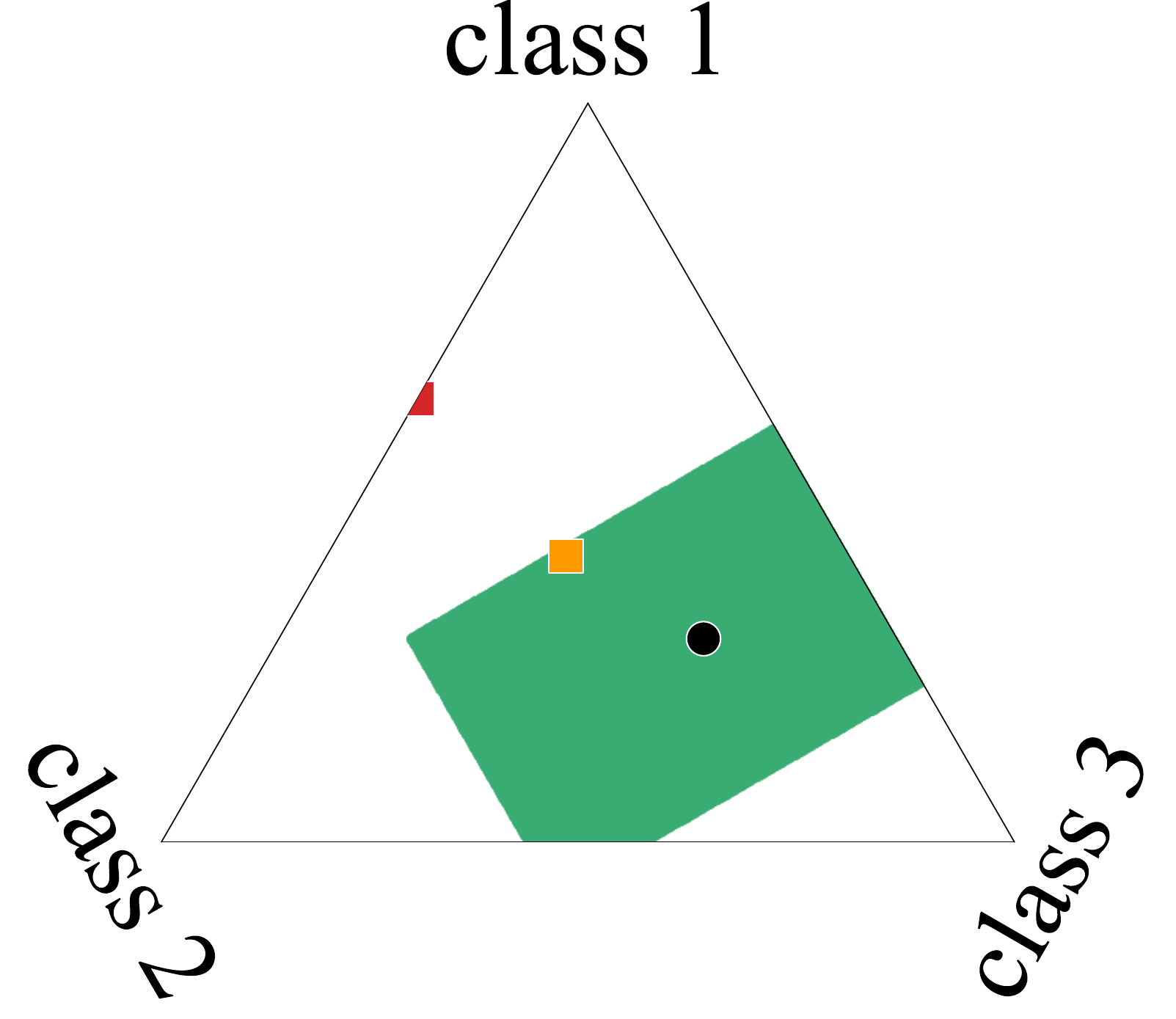} &
        \includegraphics[width=0.12\textwidth, valign=c]{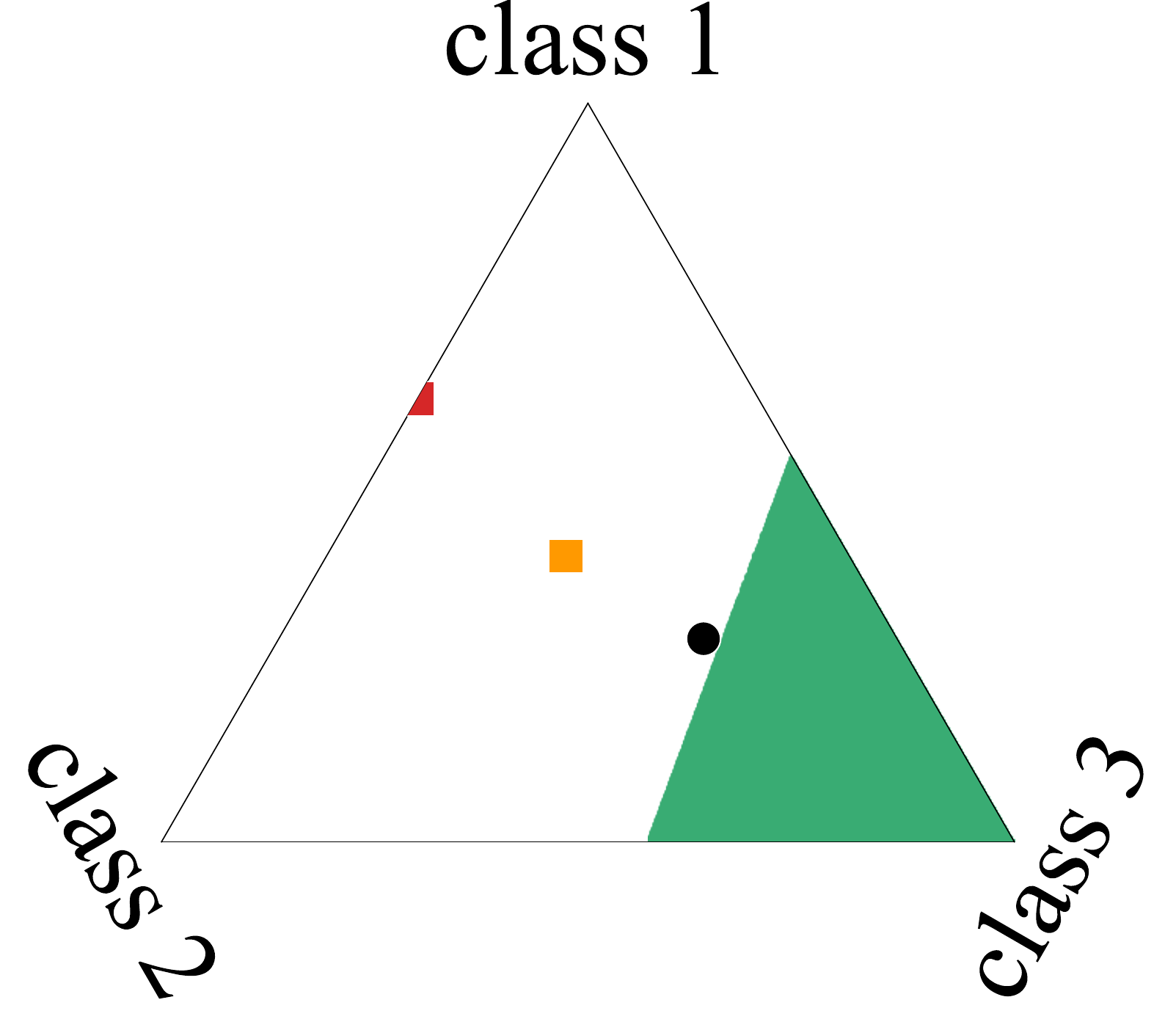}& 
        \includegraphics[width=0.12\textwidth, valign=c]{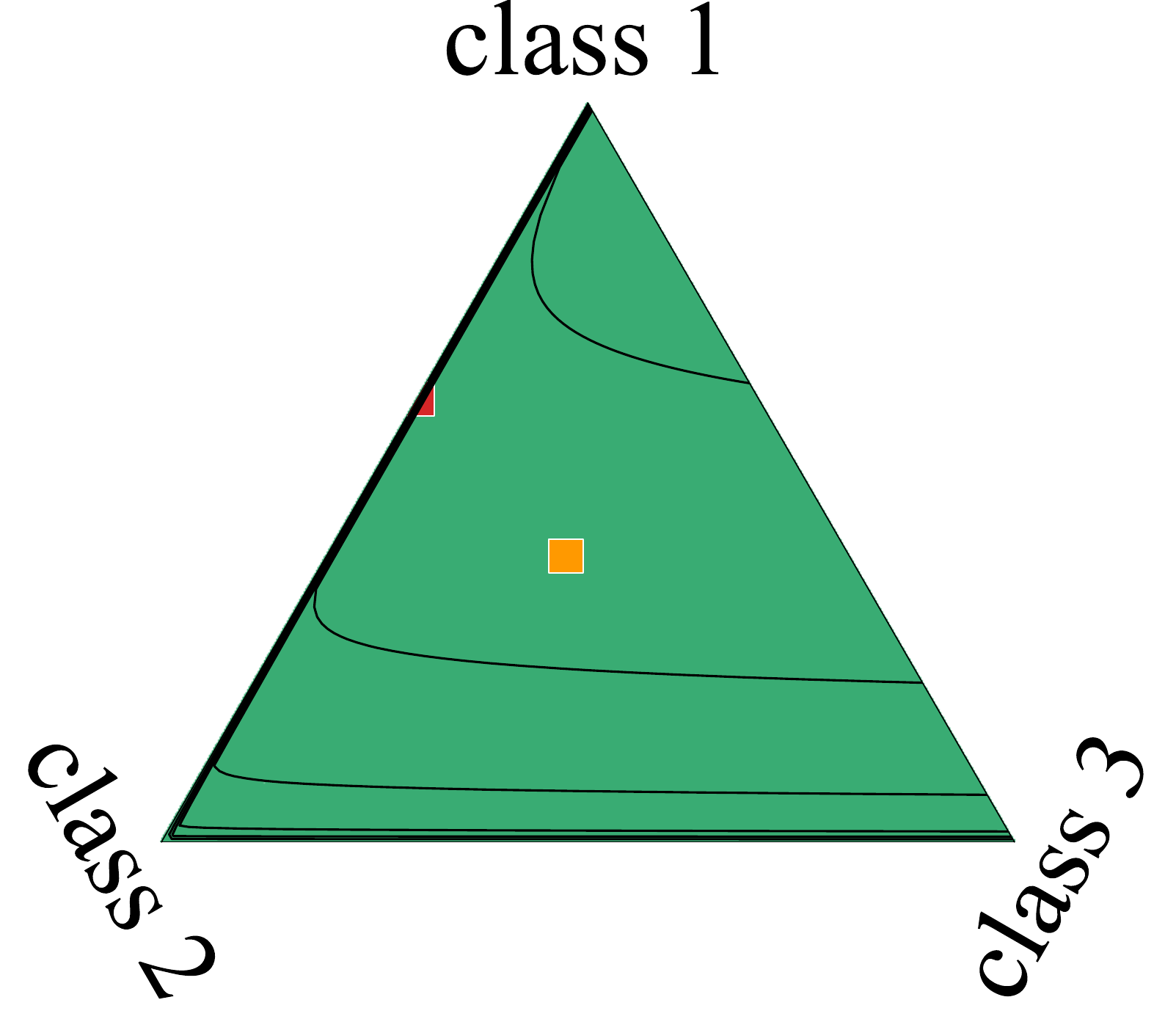} 
        \\ 
        & $0.46$ & $0.67$ & $0.39$ & $0.22$ & $1.00$    \\
        \midrule
        10&\includegraphics[width=0.12\textwidth, valign=c]{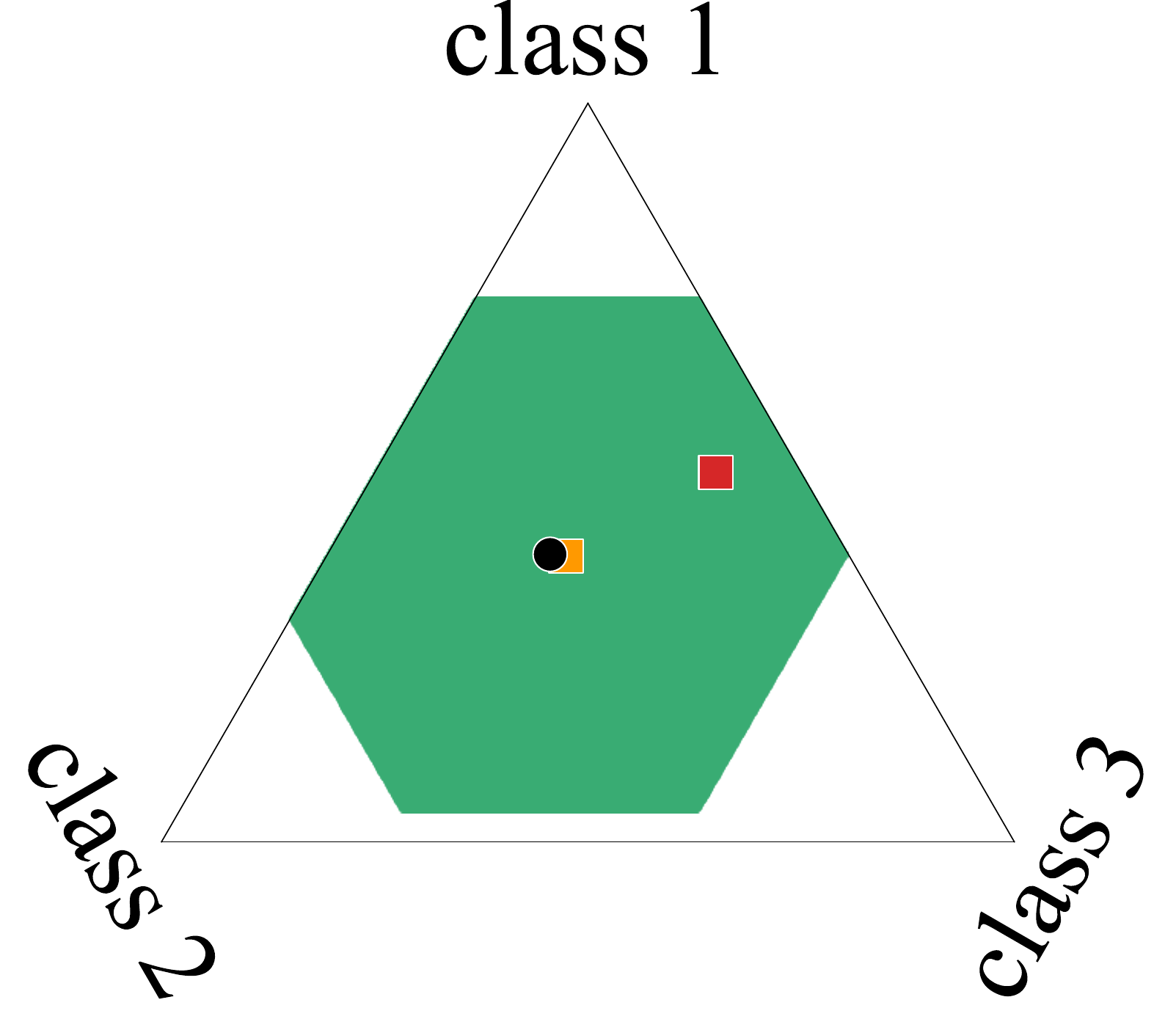} &
        \includegraphics[width=0.12\textwidth, valign=c]{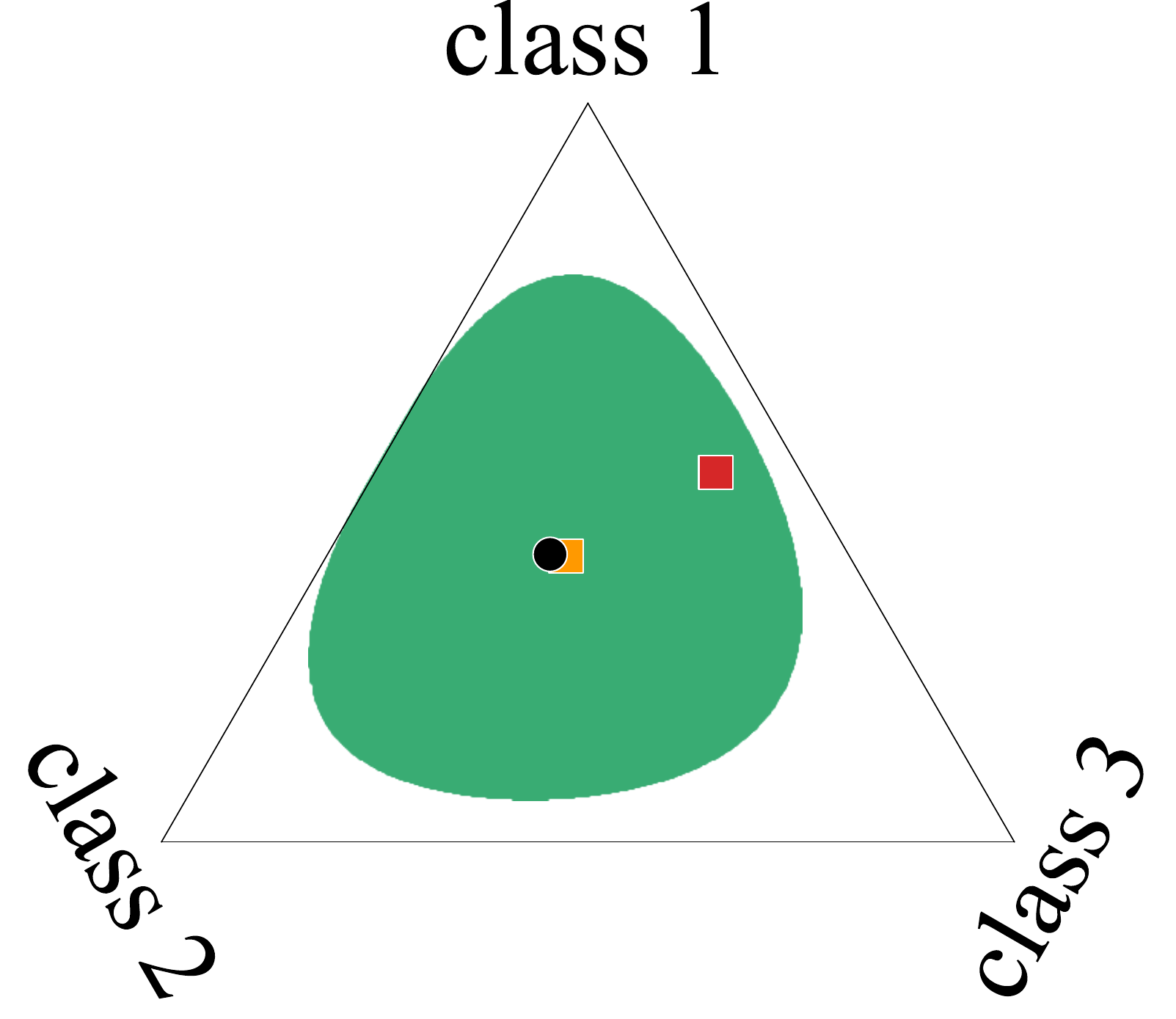} &
        \includegraphics[width=0.12\textwidth, valign=c]{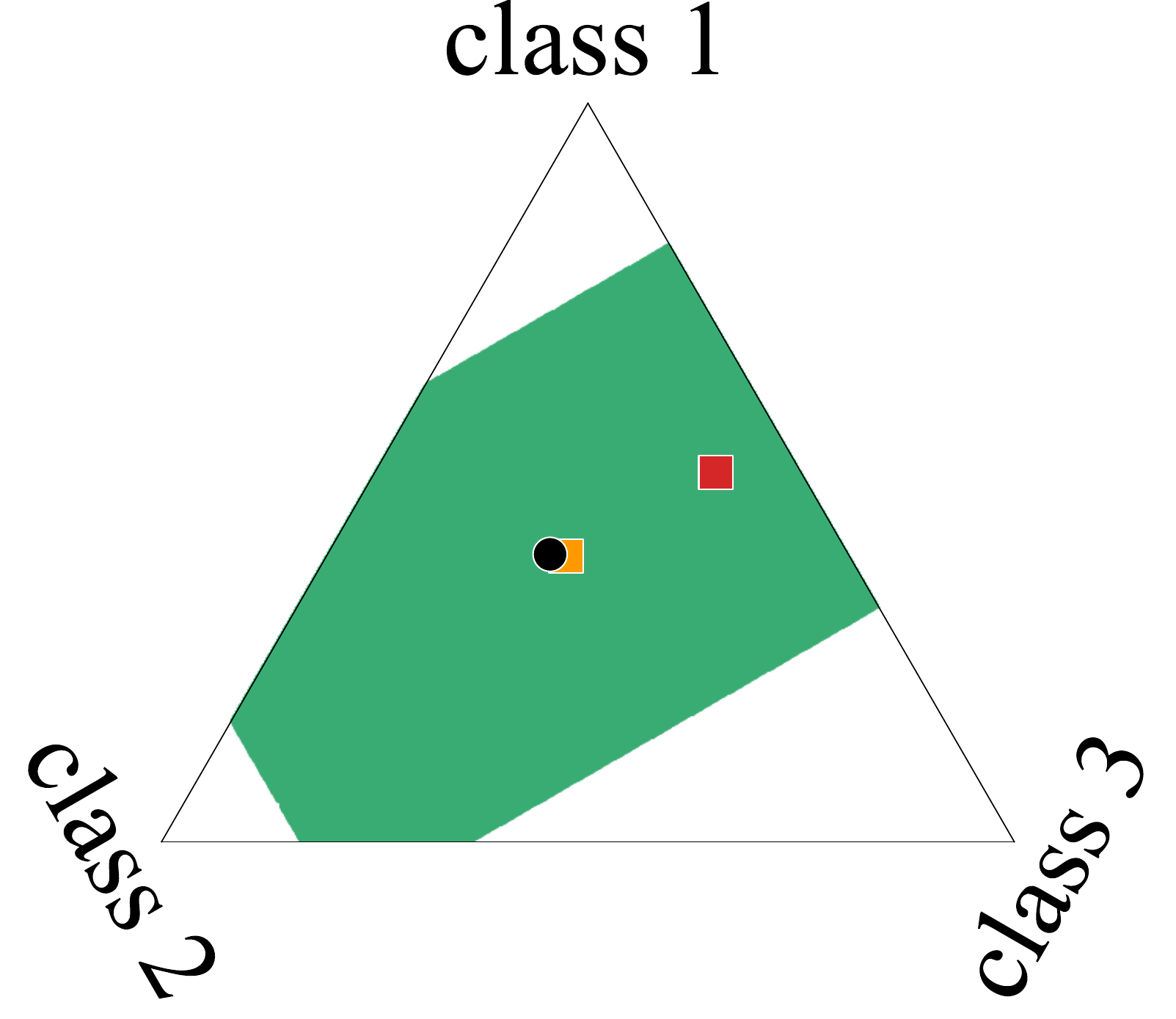} &
        \includegraphics[width=0.12\textwidth, valign=c]{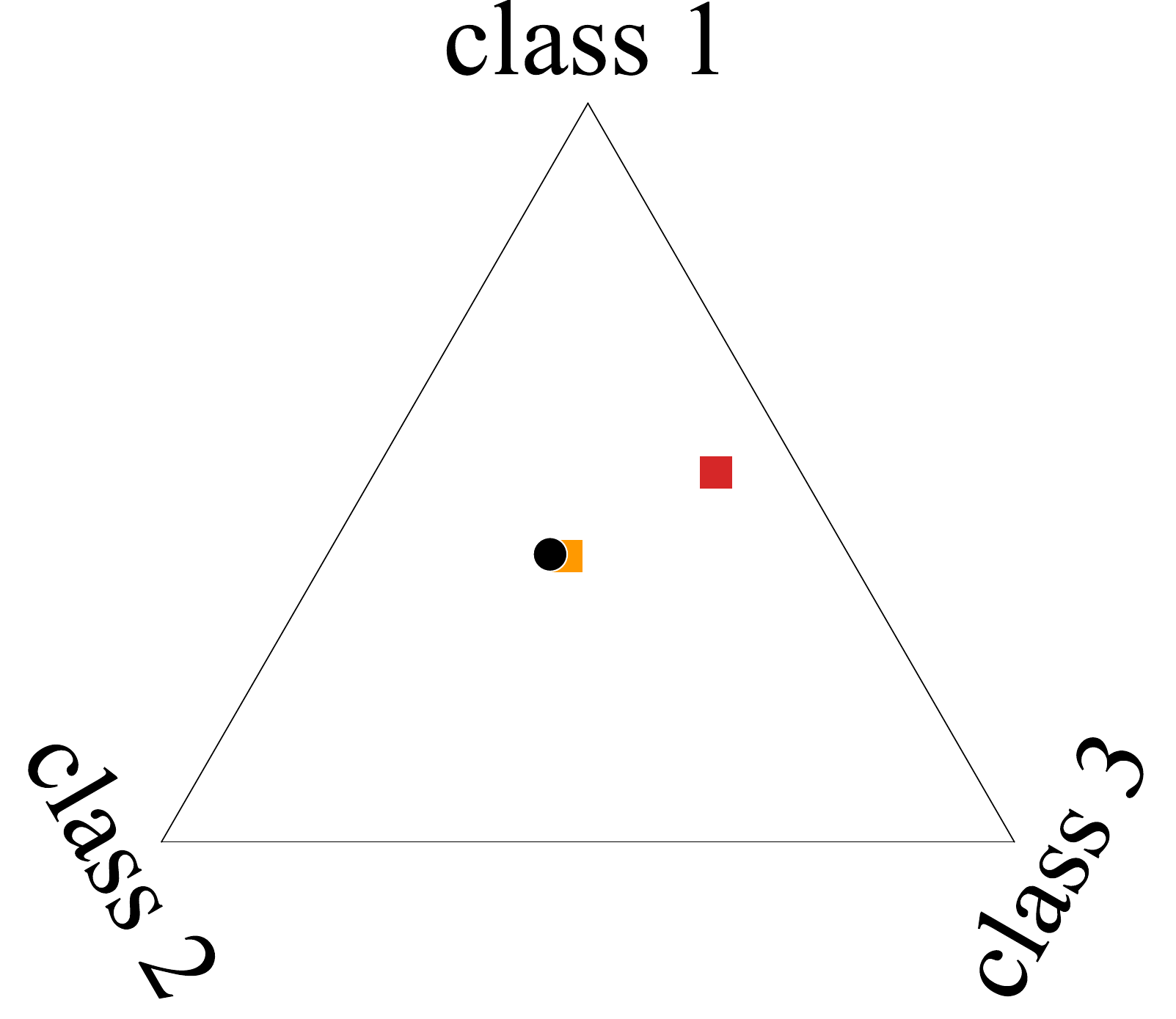}& 
        \includegraphics[width=0.12\textwidth, valign=c]{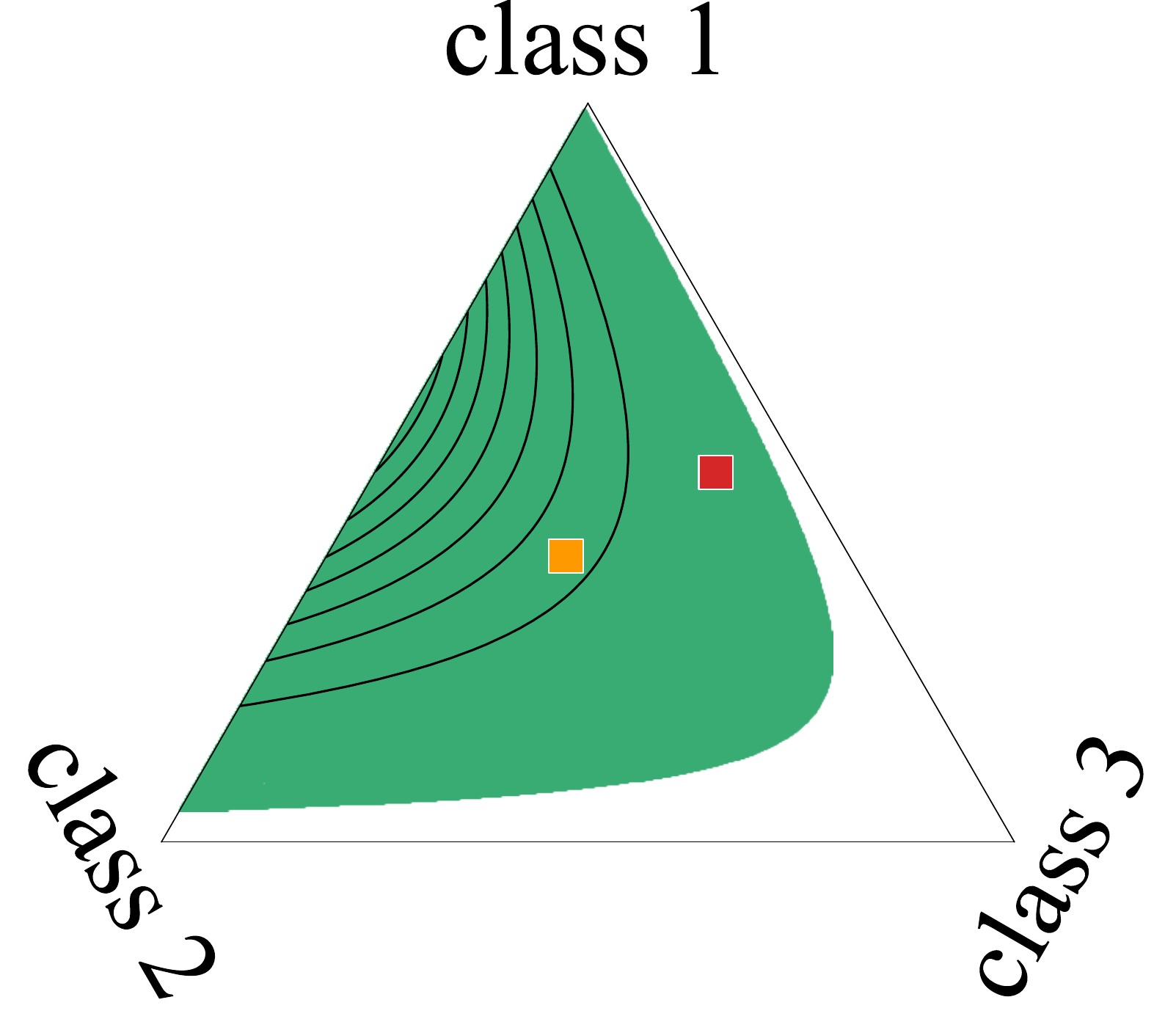} 
        \\ 
        & $0.64$ & $0.58$ & $0.68$ & $0.00$ & $0.74$   \\
        \midrule
        100&\includegraphics[width=0.12\textwidth, valign=c]{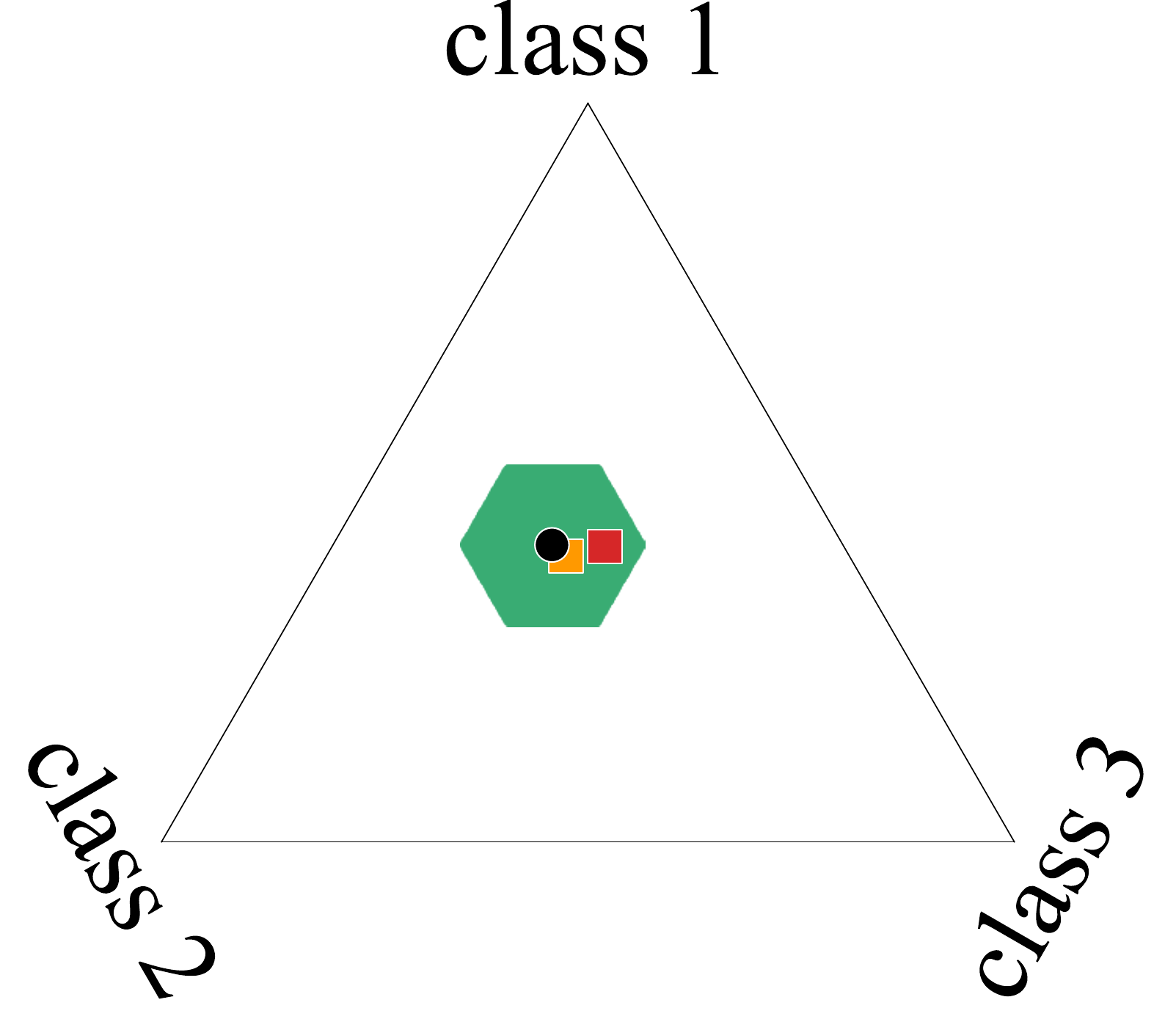} &
        \includegraphics[width=0.12\textwidth, valign=c]{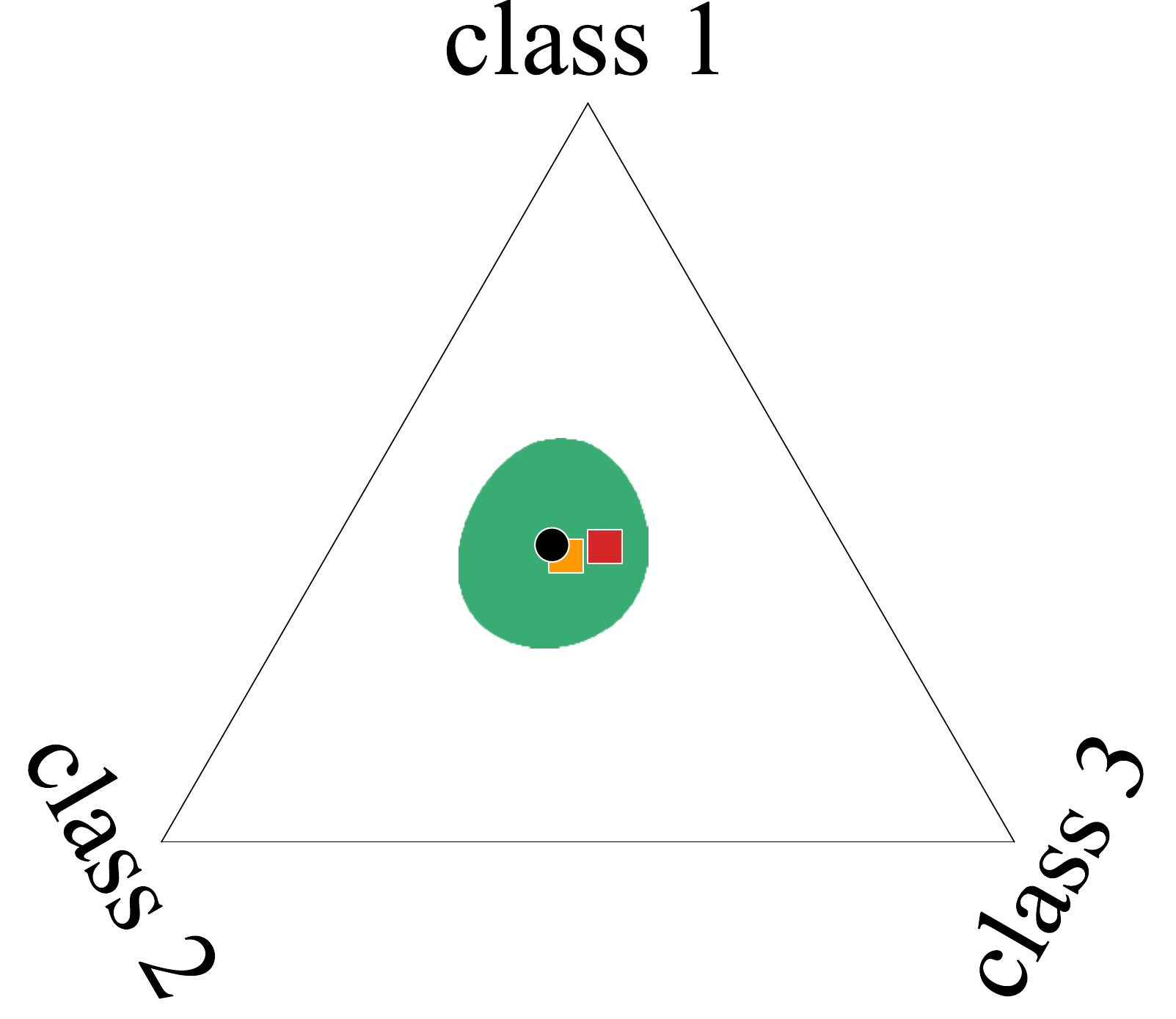} &
        \includegraphics[width=0.12\textwidth, valign=c]{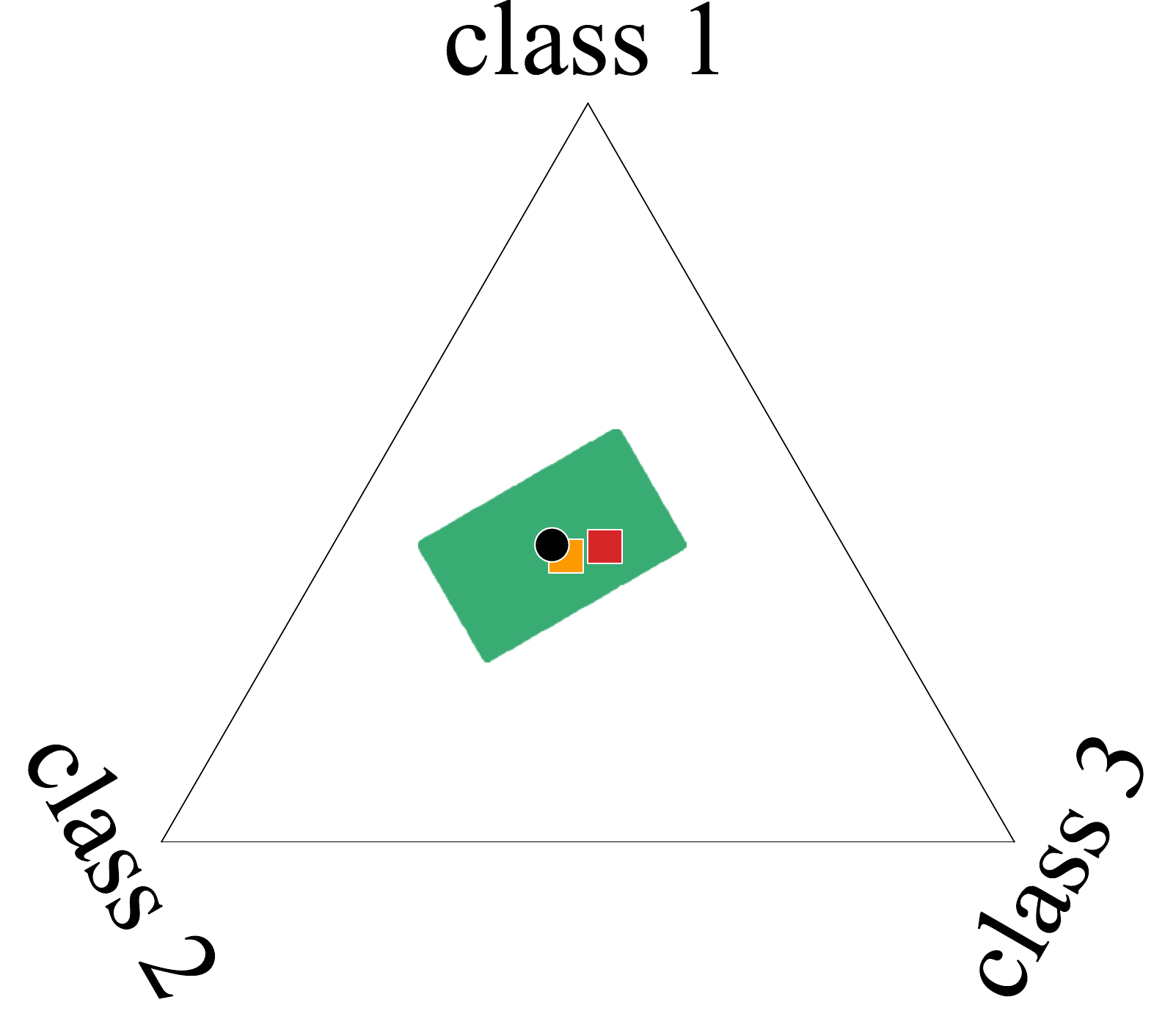} &
        \includegraphics[width=0.12\textwidth, valign=c]{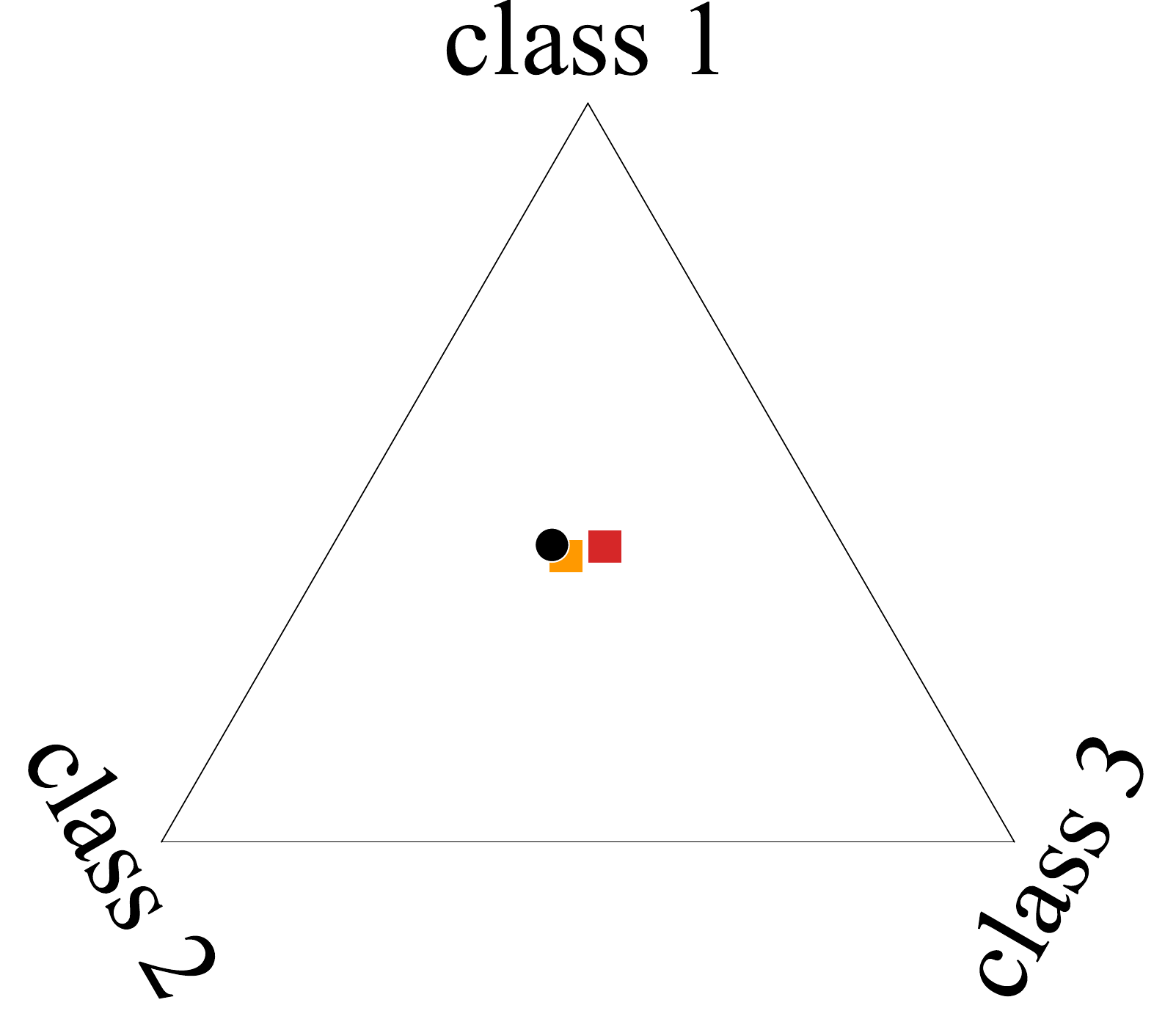}& 
        \includegraphics[width=0.12\textwidth, valign=c]{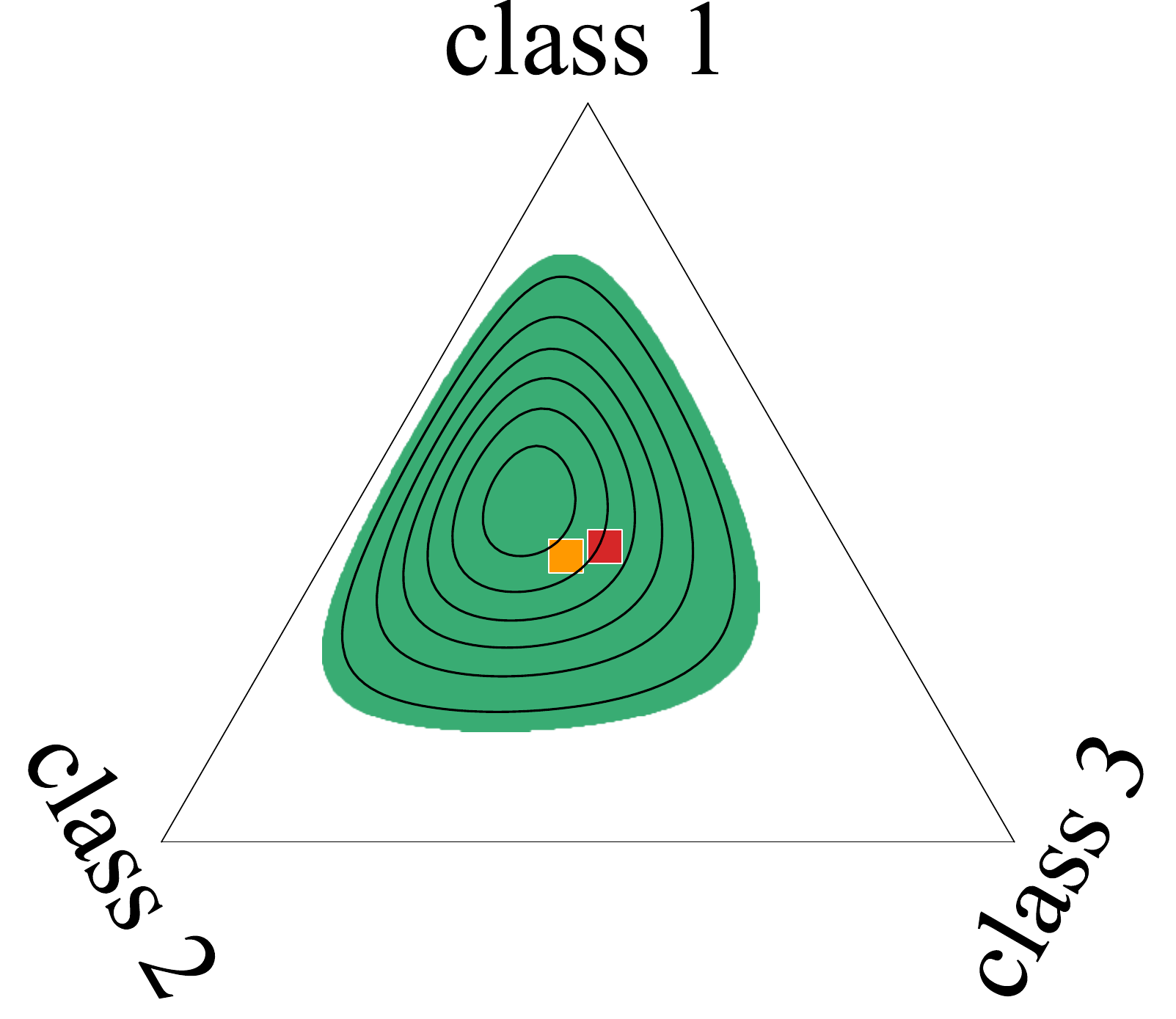} 
        \\ 
        & $0.06$ & $0.09$ & $0.09$ & $0.00$ & $0.43$    \\
        \bottomrule
    \end{tabularx}
    \caption{Credal sets derived for a synthetic data instance using various credal set predictors. Rows correspond to the number of samples utilized for distribution estimation. The ground truth distribution is marked by an {\color{gt_color} orange} square, and its noisy versions are denoted by {\color{red} red} squares. In cases employing a first-order learner (first four columns), model predictions are denoted by black circles. The predicted second-order distributions are illustrated via contour plots in the last column, where a second-order learner is employed. The miscoverage rate is $\alpha=0.05$, and the efficiency of each credal set is indicated below it.} 
    \label{fig:synthetic:m_effect}
\end{figure*}

\begin{figure*}
 \begin{subfigure}{\textwidth}
    \centering
        \includegraphics[width=0.5\textwidth]{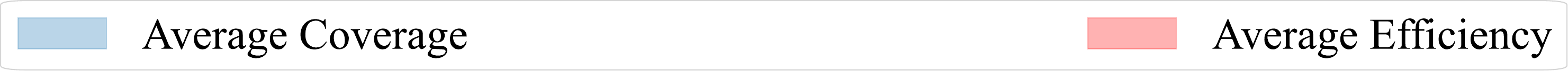}
        \vspace{3mm}
    \end{subfigure}
    \begin{subfigure}{\textwidth}
    \centering
        \includegraphics[width=\textwidth]{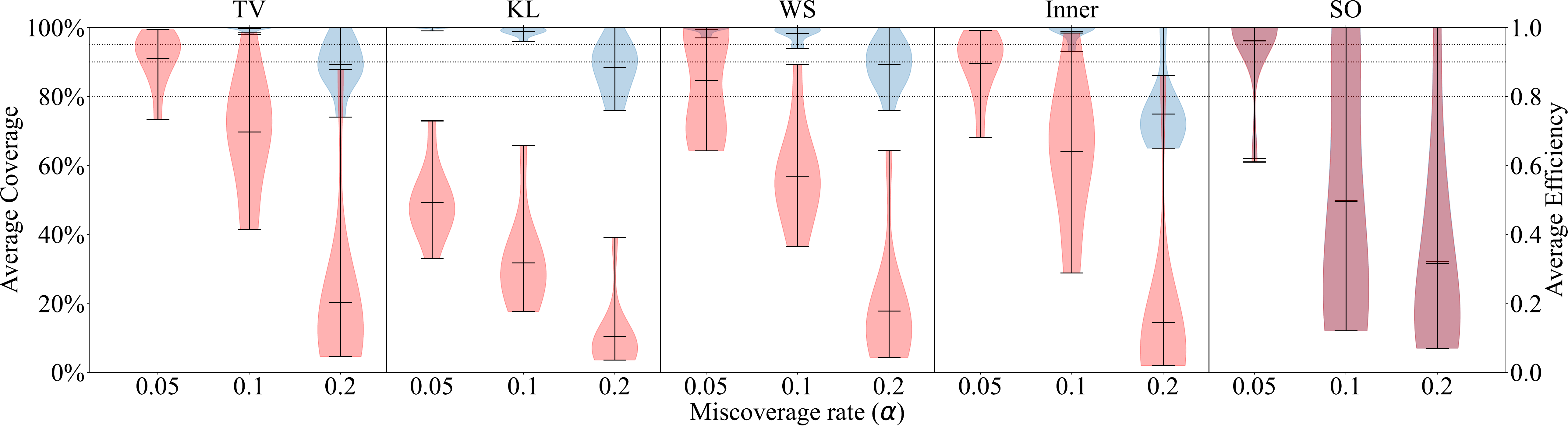}
        \caption{$m=1$.}
    \end{subfigure}

       \begin{subfigure}{\textwidth}
       \centering
        \includegraphics[width=\textwidth]{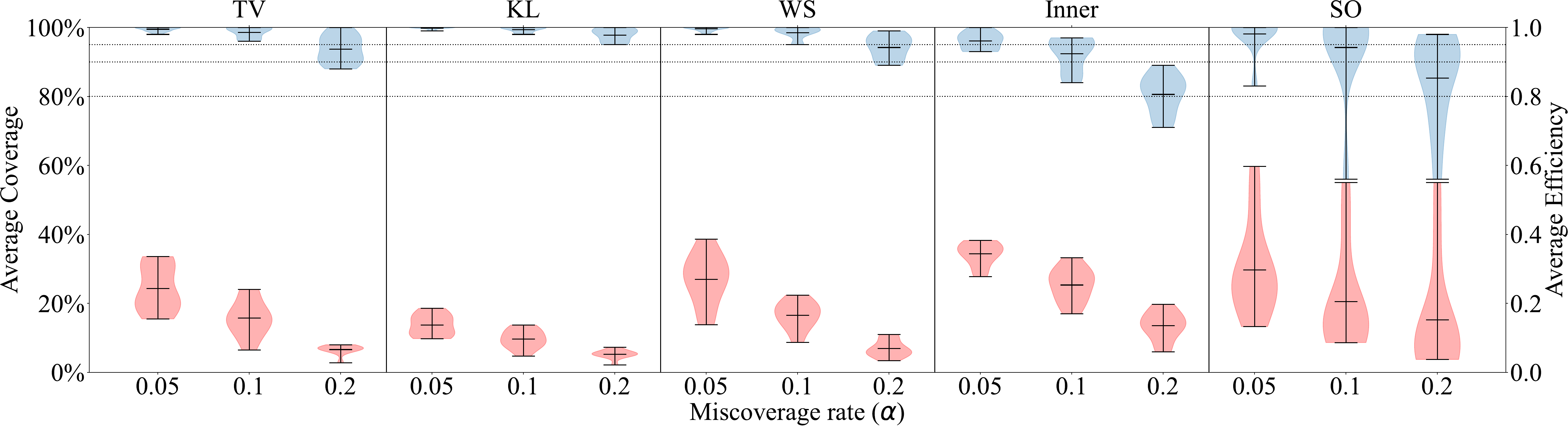}
        \caption{$m=5$.}
    \end{subfigure}
       \begin{subfigure}{\textwidth}
       \centering
        \includegraphics[width=\textwidth]{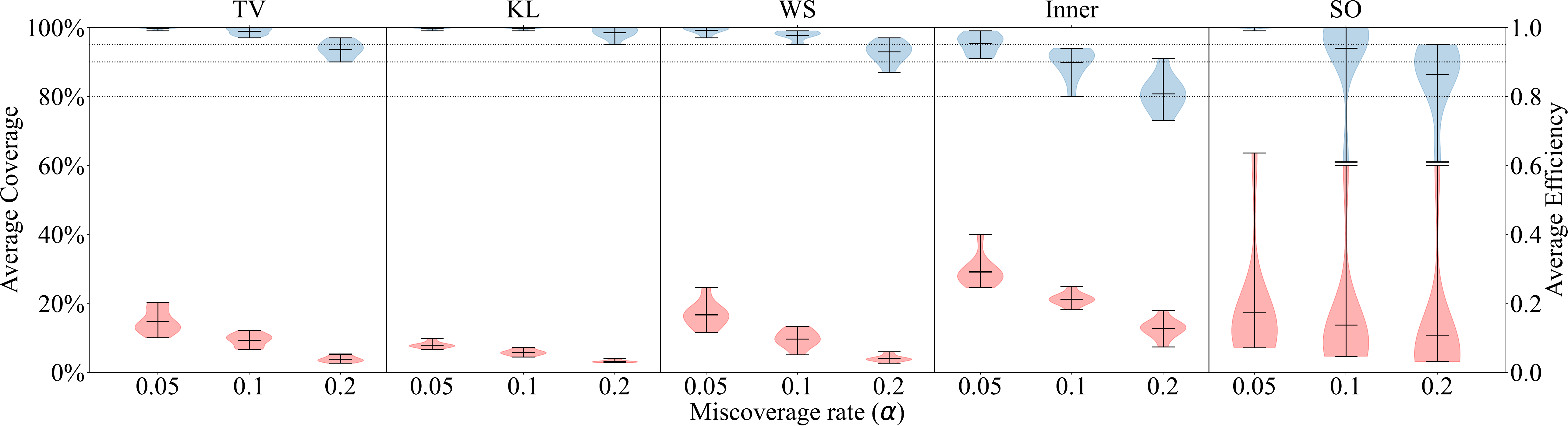}
        \caption{$m=10$.}
    \end{subfigure}
       \begin{subfigure}{\textwidth}
       \centering
        \includegraphics[width=\textwidth]{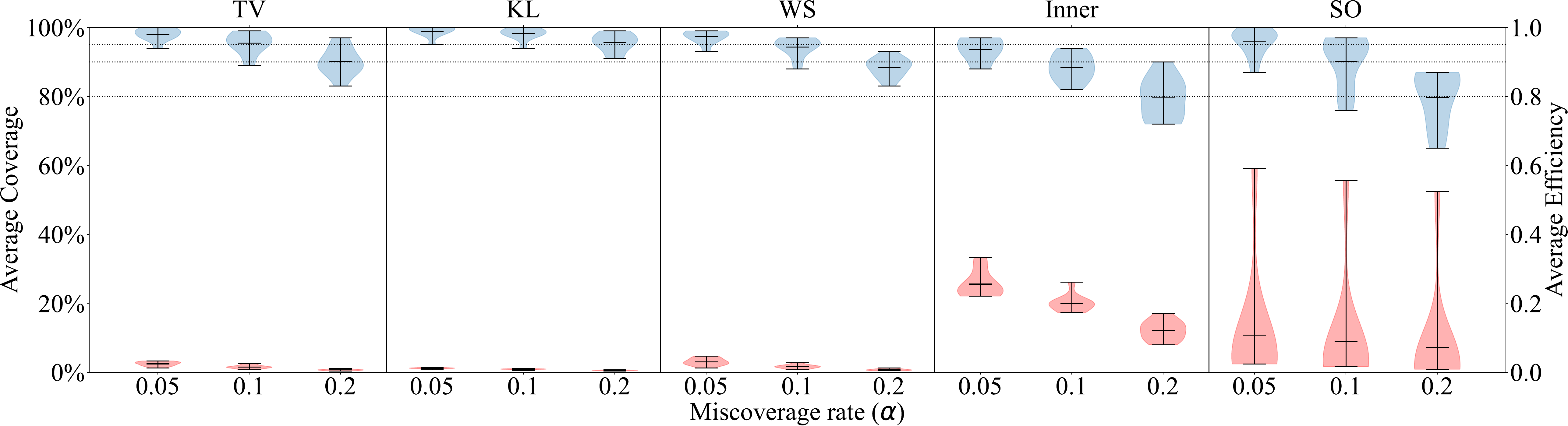}
        \caption{$m=100$.}
    \end{subfigure}
    \caption{Coverage and efficiency results of different nonconformity functions applied on the synthetic data with $K=3$. The horizontal dashed lines indicate the nominal coverage levels.}
\label{fig:synthetic:violinK=3}
\end{figure*}

Figure \ref{fig:synthetic:violin:all} depicts the overall coverage and quantile comparison between four different nonconformity functions \textbf{TV}, \textbf{KL}, \textbf{WS}, and \textbf{Inner}. In Figure \ref{fig:synthetic:m_effect}, we illustrate the evolution of the credal sets as $m$ changes from $1$ to $100$ for different nonconformity functions when $K=3$. For this case, the full comparison of efficiency and coverage across various nonconformity functions is provided in Figure \ref{fig:synthetic:violinK=3}.

% You can have as much text here as you want. The main body must be at most $8$ pages long.
% For the final version, one more page can be added.
% If you want, you can use an appendix like this one, even using the one-column format.
% %%%%%%%%%%%%%%%%%%%%%%%%%%%%%%%%%%%%%%%%%%%%%%%%%%%%%%%%%%%%%%%%%%%%%%%%%%%%%%%
% %%%%%%%%%%%%%%%%%%%%%%%%%%%%%%%%%%%%%%%%%%%%%%%%%%%%%%%%%%%%%%%%%%%%%%%%%%%%%%%

\end{document}